%% file: main.tex
\documentclass[lettersize,journal]{IEEEtran}

\usepackage{amsmath,amsfonts}
\usepackage{algorithmic}
\usepackage{algorithm}
\usepackage{array}
\usepackage[caption=false,font=normalsize,labelfont=sf,textfont=sf]{subfig}
\usepackage{textcomp}
\usepackage{stfloats}
\usepackage{url}
\usepackage{verbatim}
\usepackage{graphicx}
\usepackage{cite}
\usepackage{xcolor}
\newcommand{\new}[1]{\textcolor{black}{#1}} 
\newcommand{\newer}[1]{\textcolor{black}{#1}} 

\usepackage[colorlinks=true, linkcolor=blue]{hyperref}

\usepackage{booktabs}
\usepackage{multirow}
\usepackage{amssymb}
\usepackage{pifont}
\usepackage{makecell}
\usepackage{tabularx}

\newcommand{\eg}{\textit{e.g.,}\ }

\usepackage[edges]{forest}
\usepackage{tikz}
\usepackage{graphicx}
\definecolor{hidden-red}{RGB}{205, 44, 36}
\definecolor{hidden-blue}{RGB}{194,232,247}
\definecolor{hidden-orange}{RGB}{243,202,120}
\definecolor{hidden-green}{RGB}{34,139,34}
\definecolor{hidden-pink}{RGB}{255,245,247}
\definecolor{hidden-black}{RGB}{20,68,106}
\definecolor{purple}{RGB}{144,153,196}
\definecolor{yellow}{RGB}{255,228,123}
\definecolor{hidden-yellow}{RGB}{255,248,203}
\definecolor{tkcolor}{RGB}{224,223,255}
\definecolor{darkblue}{rgb}{0, 0.40, 0.75}

\begin{document}

\title{Embodied Robot Manipulation in the Era of Foundation Models: Planning and Learning Perspectives}

\input{sections/authers}

\maketitle



\input{sections/0-abstract}
\input{sections/1-introduction}
\input{sections/5_6-methodologies}

\input{sections/9-future_research}
\input{sections/10-conclusion}

\bibliographystyle{IEEEtran}
\bibliography{
refs/1-intro,
refs/2-hardware, refs/2-control_and_learning, refs/2-robotics_models,
refs/3-benchmarks,
refs/4-manipulation_tasks,
refs/5-high-level_planner, refs/5-affordance, refs/5-3d_representation,
refs/6-rl, refs/6-il, refs/6-rl_il, refs/6-learn_with_auxilary_tasks, refs/6-vla, refs/6-va, refs/6-latent_learning, refs/6-policy_learning, refs/6-tactile_action,
refs/7-data, refs/7-generalization,
refs/8-application,
refs/9-future_research
}

\end{document}


\title{Appendix of Embodied Robot Manipulation in the Era of Foundation Models: Planning and Learning Perspectives}

\maketitle



\input{sections/0-abstract_appendix}
\input{sections/11-appendix}

\bibliographystyle{IEEEtran}
\bibliography{
refs/1-intro,
refs/2-hardware, refs/2-control_and_learning, refs/2-robotics_models,
refs/3-benchmarks,
refs/4-manipulation_tasks,
refs/5-high-level_planner, refs/5-affordance, refs/5-3d_representation,
refs/6-rl, refs/6-il, refs/6-rl_il, refs/6-learn_with_auxilary_tasks, refs/6-vla, refs/6-va, refs/6-latent_learning, refs/6-policy_learning, refs/6-tactile_action,
refs/7-data, refs/7-generalization,
refs/8-application,
refs/9-future_research,
refs/11-appendix
}

%% file: sections/authers.tex


\author{
  Shuanghao Bai$^{1*}$ \quad Wenxuan Song$^{2*}$ \quad Jiayi Chen$^{2}$ \quad  Yuheng Ji$^{3}$ \quad Zhide Zhong$^{2}$ \quad
  Jin Yang$^{1}$ \quad Han Zhao$^{4,5}$ \quad Wanqi Zhou$^{1}$ \quad Zhe Li$^{6}$ \quad
  Pengxiang Ding$^{4,5}$ \quad Cheng Chi$^{7}$ \quad Chang Xu$^{6}$ \quad Xiaolong Zheng$^{3}$ \\ Donglin Wang$^{4}$ \quad Haoang Li$^{2\dagger}$ \quad Shanghang Zhang$^{8\dagger}$ \quad Badong Chen$^{1\dagger}$ \\
    $^{1}$ Xi’an Jiaotong University,
    $^{2}$ Hong Kong University of Science and Technology (Guangzhou), \\
    $^3$ Chinese Academy of Sciences, 
    $^4$ Westlake University, 
    $^5$ Zhejiang University, \\
    $^6$ University of Sydney, 
    $^7$ BAAI, 
    $^8$ Peking University
    \vspace{1mm} \\
    {\small 
    $^*$ \textbf{Equal Contributors}~~  \textbf{$^\dagger$~ Corresponding Authors}} \\
  GitHub: \href{https://github.com/BaiShuanghao/Awesome-Robotics-Manipulation}{\texttt{Awesome-Robotics-Manipulation}} 
}

%% file: sections/0-abstract.tex
\begin{abstract}
Recent advances in vision, language, and multimodal learning have significantly accelerated progress in robotic foundation models, with robotic manipulation remaining one of the most challenging embodied tasks. Its difficulty lies in integrating perception, semantic understanding, task reasoning, physically grounded action generation, and reliable execution. This survey examines robotic manipulation from an algorithmic perspective and organizes recent learning-based approaches through a unified abstraction of high-level planning and \new{low-level action modeling}. At the high level, we extend the classical notion of task planning to include reasoning over language, code, affordances, geometric constraints, and 3D representations. At the low level, we present a learning-paradigm-oriented taxonomy of learning-based action models, covering input modeling, latent learning, and policy learning. \new{Within this abstraction, foundation models contribute either by generating structured planning artifacts that are instantiated as constraints or latent inputs for downstream action generation, or by directly modeling executable actions and trajectories.} Finally, we summarize open challenges and future directions related to scalability, generalization, data efficiency, multimodal physical interaction, and safety. Together, this survey provides a structured view of the design space and 
emerging trends in foundation models for robotic manipulation.
\end{abstract}


%% file: sections/1-introduction.tex
\section{Introduction}
\label{sec: intro}

In recent years, embodied intelligence has emerged as a major frontier in artificial intelligence, driven by advances in computer vision, natural language processing, and large-scale foundation models~\cite{touvron2023llama, liu2023visual}. Among embodied tasks, robotic manipulation is particularly challenging because it requires closing the loop from perception and semantic understanding to task reasoning, physically grounded action generation, and reliable execution.  \new{This difficulty is further compounded by the diversity of manipulation settings, which vary across operation mode, object property, and robot embodiment. Representative examples include basic manipulation with single-arm or dual-arm robots for object-centric tasks, dexterous manipulation with multi-fingered hands for fine-grained control, deformable object manipulation over non-rigid objects, and embodied settings such as mobile, quadrupedal, and humanoid manipulation~\cite{bai2025towards}, as illustrated in Figure~\ref{fig: manipulation_tasks}. Unlike navigation or locomotion, which primarily regulate the robot's own motion, manipulation involves contact-rich interactions that intentionally alter object states, making it inherently a multi-level problem spanning reasoning, action generation, and control.}

\input{figures/1_intro/manipulation_tasks}
\input{figures/1_intro/summary_new}

Recent progress in robotic manipulation has been largely driven by data-driven approaches, including deep visuomotor policies~\cite{levine2016end, pinto2016supersizing, yang2024attribute}, imitation learning~\cite{duan2017one}, reinforcement learning~\cite{rajeswaran2018learning}, and, more recently, robotic foundation models~\cite{zitkovich2023rt, kim2025openvla, ji2025robobrain}. 
\new{These foundation models enhance perceptual and semantic representations and enable flexible task conditioning through multimodal inputs such as language and vision. Importantly, their roles in manipulation systems naturally align with the multi-level structure of the problem: they can operate either as high-level modules that generate structured guidance (e.g., task plans~\cite{brohan2023can, driess2023palm}, code~\cite{liang2023code}, affordances~\cite{shridhar2022cliport}, or geometric constraints~\cite{huang2023voxposer}) or as low-level models that directly parameterize actions or trajectories~\cite{black2025pi_0}.}

\new{
To systematically capture this multi-level structure, we adopt the taxonomy illustrated in Figure~\ref{fig: taxonomy}, which decomposes robotic manipulation into three layers: High-Level Planning, Low-Level Action Modeling, and Actuation-Level Control. High-level modules produce structured artifacts that guide downstream behavior, while low-level modules generate executable actions or trajectories that are ultimately realized through the actuation layer on the robot. Within this hierarchy, our primary focus is on High-Level Planning (Section~\ref{sec: high_level_planner}) and Learning-Based Low-Level Action Modeling (Section~\ref{sec: low_level_control}), where foundation models most directly reshape robotic manipulation. In this context, we define a robotic foundation model as a model pretrained on diverse robotic and/or multimodal data that enables transferable manipulation across tasks, objects, or environments through reusable representations or policies. Rather than imposing rigid thresholds on model size or dataset scale, we characterize such models by three practical properties: diverse pretraining, broad capability scope, and reusability for downstream adaptation.
}

Building on this taxonomy, we develop a structured understanding of learning-based robotic manipulation through the dual lens of High-Level Planning and Learning-based Low-Level Action Modeling, providing a unified view of how modern systems couple high-level reasoning with low-level action generation. Unlike prior surveys that organize the literature mainly by model families, such as vision-language-action models~\cite{ma2024survey, zhong2025survey, xiang2025parallels}, diffusion policies~\cite{wolf2025diffusion}, or generative approaches~\cite{zhang2025generative}, our perspective emphasizes the functional roles of different components and their interactions across levels.
Under this view, we introduce a systematic taxonomy that categorizes high-level methods into task planning, programmatic planning, geometric constraint generation, affordance modeling, and 3D representations, while organizing low-level action models by input modeling, latent learning, and policy learning. This unified framework enables more coherent comparisons across approaches and reveals common design patterns that are less visible when methods are grouped only by model family.
Based on this structured perspective, we also synthesize emerging trends and identify four future directions: more general-purpose robotic foundation models, improved data scaling and sim-to-real transfer, richer multimodal physical interaction, and safer real-world deployment.

%% file: figures/1_intro/manipulation_tasks.tex
\begin{figure}[t]
\centering
\includegraphics[width=0.95\linewidth]{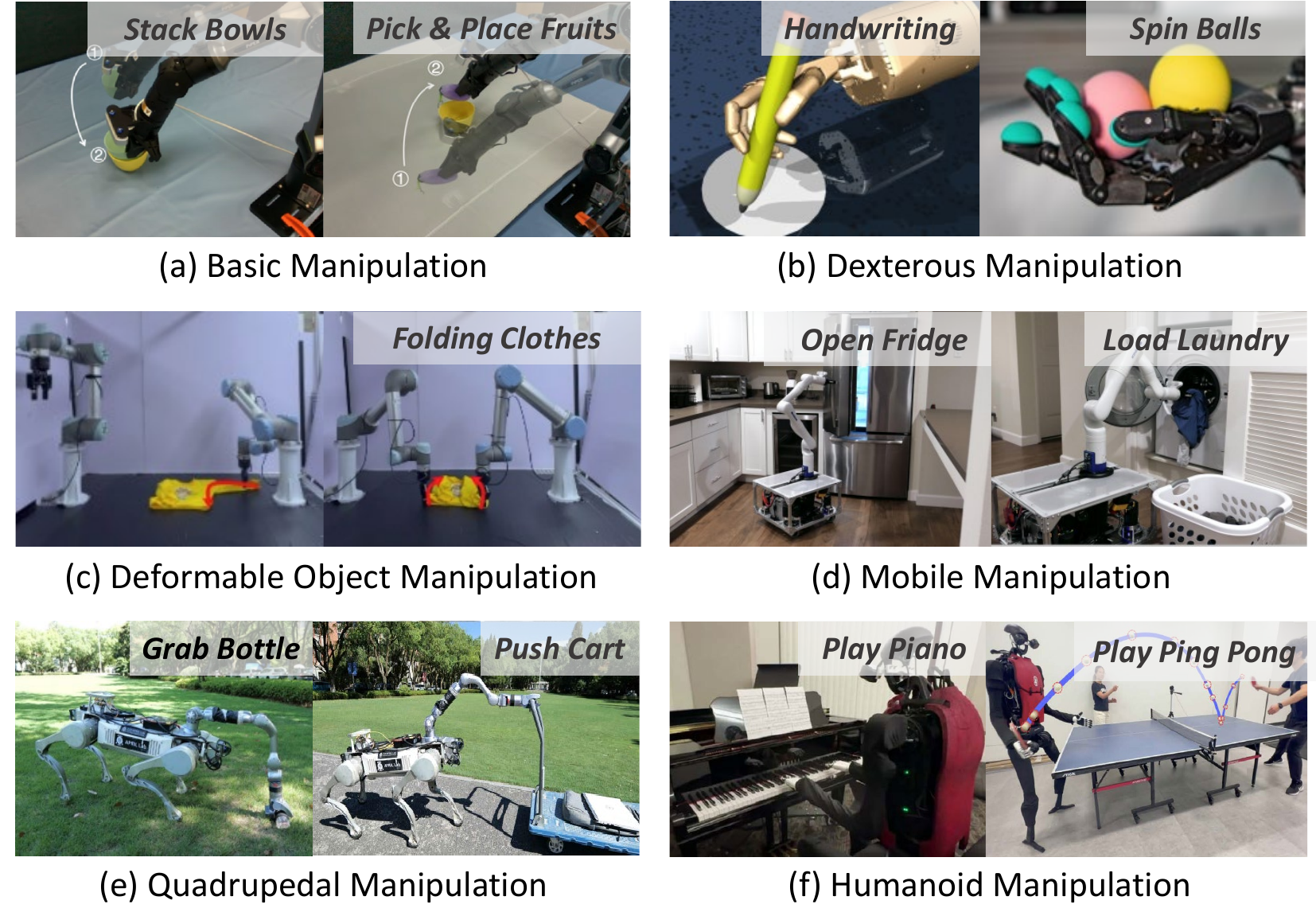}  
\vskip -0.1in
\caption{
\new{Representative robotic manipulation scenarios spanning diverse manipulation settings, including basic manipulation~\cite{song2026reconvla}, dexterous manipulation~\cite{nagabandi2020deep}, deformable object manipulation~\cite{luan2025learning}, and embodied manipulation such as mobile~\cite{wu2025tidybot++}, quadrupedal~\cite{hou2025efficient}, and humanoid manipulation~\cite{fu2025humanplus}.}
}
\label{fig: manipulation_tasks}
\end{figure}

%% file: figures/1_intro/summary_new.tex
\begin{figure}[t]
\centering
\includegraphics[width=0.9\linewidth]{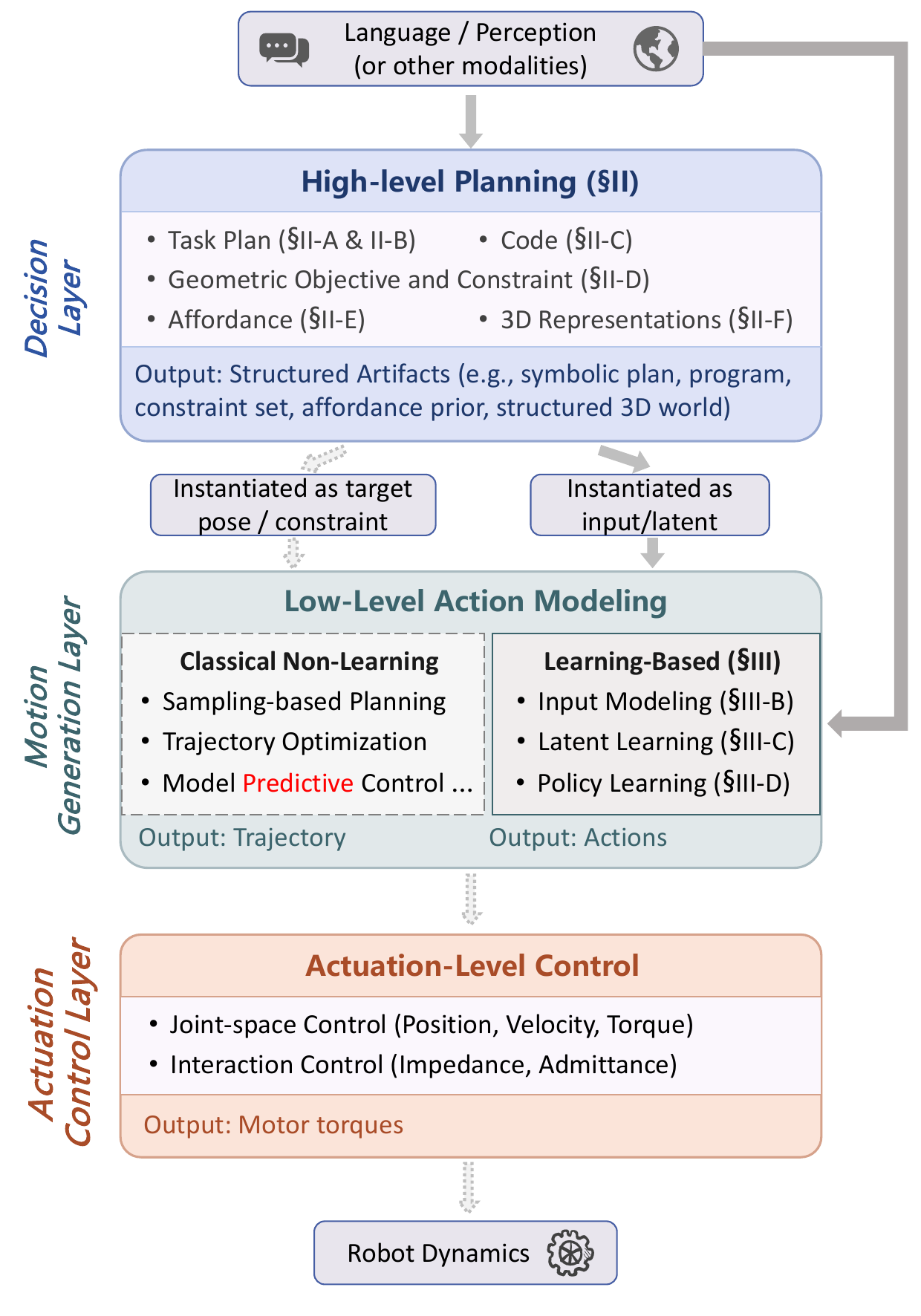} 
\vskip -0.12in
\caption{
\new{Overview of the proposed taxonomy for foundation-model-based robotic manipulation. The pipeline is decomposed into three layers: High-Level Planning, Low-Level Action Modeling, and Actuation-Level Control. High-level modules produce structured planning artifacts, either as explicit inputs or as constraints for downstream action generation. Low-level action modeling includes both classical non-learning methods and learning-based approaches, and operates either on such artifacts or directly on raw multimodal observations. Solid arrows denote the primary scope of this survey, namely High-Level Planning and Learning-Based Low-Level Action Modeling.}
}
\label{fig: taxonomy}
\end{figure}

%% file: sections/5_6-methodologies.tex


\input{sections/5_high-level_planner/high-level_planner}

\input{sections/6-low_level_controller/low-level_learning-based_control}

%% file: sections/5_high-level_planner/high-level_planner.tex

\section{High-level Planning}
\label{sec: high_level_planner}

\new{High-level planning plays a central role in robotic manipulation by providing structured guidance for downstream execution. It determines action intent, temporal organization, and attentional focus over the environment, thereby shaping how perception and control are coordinated. Large language models and multimodal large language models have become increasingly influential at this level, supporting task decomposition, skill sequencing, and adaptive reasoning grounded in language and perception. Complementarily, affordance learning highlights action-relevant regions and object relations, while 3D scene representations provide spatial structure and geometric constraints for planning. Together, these approaches show that high-level planning can guide manipulation through different forms of structured intermediate representations.
Formally, this layer can be written as
\begin{equation}
A_{\text{plan}} = f_{\text{plan}}(o,l),
\end{equation}
where $o$ denotes observations and $l$ denotes language instructions. Depending on the method, the planning artifact $A_{\text{plan}}$ may take the form of task plans, programs, geometric constraints, affordance priors, or 3D representations. 
These artifacts are not executable actions themselves; instead, they provide structured guidance for downstream action generation, for example by being transformed into target poses or spatial constraints for conventional motion planning and control, or by being encoded as conditions for learning-based action models.}

This view provides a natural basis for organizing the literature in this section. We summarize this taxonomy in Figure~\ref{fig: high-level_planner_taxonomy_colored} and provide an illustrative overview in Figure~\ref{fig: high-level_planner}.

\input{tables/5_high-level_planner/high-level_planner}
\input{figures/5_high-level_planner/high-level_planner}

\subsection{LLM-based Task Planning}
\label{subsec: llm-based_planners}

Early high-level planning methods followed a symbolic grounding paradigm, where neural networks mapped demonstrations and observations into symbolic states and goals~\cite{huang2019continuous}. With the emergence of large language models, this paradigm shifted toward language-centric planning. SayCan~\cite{brohan2023can} was an early milestone, using an LLM as a global planner to select executable skills based on language-level relevance and affordance-based feasibility. Grounded Decoding~\cite{huang2023grounded} further relaxed the reliance on fixed skill libraries by enabling joint decoding between language and grounding models for open-vocabulary planning.
A key limitation of early LLM-based planners is the lack of execution-time feedback. Inner Monologue~\cite{huang2022inner} addresses this through a closed-loop framework that incorporates real-time feedback, scene descriptions, and human input for dynamic plan revision. Subsequent work further improves robustness and scope through feedback-driven planning~\cite{song2023llm}, explicit modeling of constraints and failures for long-horizon reasoning~\cite{liu2023llm+, liu2023reflect}, multi-agent collaboration~\cite{singh2024malmm}, more open-ended interactive settings~\cite{wang2024polaris, zhao2023chat}, and multi-robot coordination~\cite{mandi2024roco}.

\new{
Overall, LLM-based task planning provides strong semantic reasoning and task decomposition, but purely language-based plans often lack numerically consistent physical understanding and may fail to satisfy geometric, kinematic, or contact constraints required for reliable execution. As a result, semantically plausible plans may still be physically infeasible, making downstream grounding, affordance estimation, and execution feedback essential for robust robotic manipulation.
}

\subsection{MLLM-based Task Planning}
\label{subsec: mllm-based_planners}

Early large language models are inherently unimodal, operating only on text, with perception handled by separate vision modules whose outputs are serialized into language. Recent multimodal large language models (MLLMs)~\cite{liu2023visual, li2023blip} relax this separation by jointly reasoning over vision and language, and have increasingly been adopted in robotic manipulation to improve high-level planning while simplifying system design.

Existing work mainly follows three directions. A first line adapts general-purpose MLLMs for embodied decision making. PaLM-E~\cite{driess2023palm} co-trains a vision--language model on robot embodiment data alongside standard VLM objectives, enabling end-to-end task reasoning at the cost of substantial data and computation. \new{VILA~\cite{hu2024look}} instead shows that off-the-shelf GPT-4V can support manipulation planning without robot-specific fine-tuning, while PG-InstructBLIP~\cite{gao2024physically} improves manipulation-relevant reasoning by injecting physical priors through lightweight object-centric fine-tuning. A second line strengthens MLLM-based planning through additional reasoning structure, such as chain-of-thought reasoning~\cite{mu2023embodiedgpt, zhang2025learning2}, scene-graph-based spatial structure~\cite{qi2025sofar}, multi-agent coordination~\cite{zeng2023socratic}, and failure-aware planning~\cite{duan2025aha}. More recently, a third line has begun to develop robotics-specific MLLMs trained on large-scale robot-centric data, including RoboBrain~\cite{ji2025robobrain}, \new{Gemini Robotics-ER~\cite{team2025gemini}}, and RynnEC~\cite{dang2025rynnec}, suggesting a promising step toward foundation models tailored for robotic manipulation.

\new{
A practical consideration of MLLM-based task planning is computational cost and latency. Unlike reactive visuomotor policies or VLA-style action models, whose inference is directly coupled to action generation, MLLM-based planners typically operate at the task level by producing intermediate plans or subgoals. Their cost therefore mainly affects high-level decision or replanning latency, rather than the inner control loop during trajectory execution.
}

\subsection{\new{Programmatic Planning}}
\label{subsec: code_generation}

While LLMs and MLLMs have shown strong capabilities in task decomposition and high-level reasoning, purely language-based plans often lack the precision and executability required for reliable manipulation in diverse environments. Programmatic planning addresses this limitation by using code generation as an intermediate abstraction between high-level reasoning and low-level control. By expressing plans as executable programs, these methods provide explicit structure, conditional logic, and compositionality, enabling finer-grained and more adaptive control.

Early efforts explored translating natural language instructions into programmatic representations without modern foundation models~\cite{venkatesh2021translating}. With the advent of LLMs, Code as Policies~\cite{liang2023code} established this paradigm by exposing perception and control APIs as prompts, allowing an LLM to generate executable code for robot behavior. Related work such as ProgPrompt~\cite{singh2023progprompt} further demonstrated the generality of code-driven control for manipulation. Subsequent studies expanded the scope and robustness of this paradigm: Instruct2Act~\cite{huang2023instruct2act} improves zero-shot generalization by combining code generation with visual foundation models, Demo2Code~\cite{wang2023demo2code} summarizes long-horizon demonstrations into compact programs, and SHOWTELL~\cite{murray2024teaching} directly translates visual demonstrations into policy code. To improve execution robustness, Statler~\cite{yoneda2024statler} maintains an explicit world state during program execution, while HyCodePolicy~\cite{liu2025hycodepolicy} integrates symbolic execution traces with perceptual feedback for more reliable closed-loop control.
Overall, programmatic planning provides a structured and interpretable interface that more tightly couples high-level reasoning with low-level execution. 
\new{However, the explicit executability of generated programs also raises challenges for safety and formal verification, since LLM-generated code may contain invalid operations, unsafe action sequences, or logic that fails to respect physical constraints. These issues make program correctness, constraint satisfaction, and runtime monitoring especially important in safety-critical robotic settings.}

\subsection{\new{Geometric Constraint-based Planning}}
\label{subsec: motion_planning}

\new{
Geometric objective and constraint generation serves as an intermediate abstraction between symbolic task reasoning and low-level motion execution. Rather than directly synthesizing trajectories, these approaches translate language and perception into structured spatial objectives or relational constraints that are subsequently optimized by classical motion planners.
VoxPoser~\cite{huang2023voxposer} constructs composable 3D value maps conditioned on language instructions, enabling motion generation through optimization over spatial cost fields. ReKep~\cite{huang2025rekep} formulates manipulation as relational keypoint constraint satisfaction, producing geometry-aware constraints over object-centric keypoints. CoPa~\cite{huang2024copa} infers part-level spatial relations and converts them into target end-effector poses for downstream planning, explicitly incorporating contact geometry and functional part relations. Similarly, GeoManip~\cite{tang2025geomanip} treats geometric constraints as a general-purpose interface between semantic reasoning and optimization-based motion generation.
Collectively, these methods use foundation models to infer structured geometric planning artifacts, while delegating final trajectory generation to downstream planners or optimization modules. This decomposition improves modularity, interpretability, and physical grounding, and positions geometric objective generation as a mid-level planning interface within robotic manipulation.
}

\subsection{\new{Affordance-based Planning}}
\label{subsec: affordance_as_planner}

Affordance learning shifts the focus of robotic manipulation from recognizing what objects are to inferring what actions they enable. Originating from Gibson’s theory of affordances~\cite{gibson2014theory}, this concept links perception to action by characterizing the interaction possibilities that objects or environments provide relative to an agent’s capabilities. In robotics, affordances serve as a useful abstraction for grounding manipulation decisions in physical structure and functional intent. Early work emphasized geometric and model-based reasoning, while recent advances increasingly learn affordances directly from sensory data, often with self-supervision or foundation-model priors. In this section, we review affordance learning from four perspectives: geometric, visual, semantic, and multimodal.

\noindent \textbf{Geometric Affordance.}
Geometric affordance relates interaction possibilities to an object's 3D geometry, part structure, and kinematic properties. This line of work focuses on inferring part-level geometry and articulation constraints directly from 3D observations~\cite{jiang2022ditto, geng2023gapartnet}. Representative methods such as Ditto~\cite{jiang2022ditto} and GAPartNet~\cite{geng2023gapartnet} ground affordances in object dynamics and reusable part taxonomies, while compositional approaches such as CPM~\cite{liu2023composable} model manipulation as structured combinations of geometric constraints.

\noindent \textbf{Visual Affordance.}
Visual affordance learning infers action possibilities directly from 2D observations, typically through dense pixel-wise affordance prediction~\cite{zeng2021transporter, borja2022affordance}. Transporter Networks~\cite{zeng2021transporter} established this paradigm by predicting pick-and-place heatmaps in pixel space without explicit object models, while later work such as VAPO~\cite{borja2022affordance} improved scalability through self-supervised learning from unstructured play data.

\noindent \textbf{Semantic Affordance.}
Semantic affordance leverages semantic priors, such as object categories or part names, to guide manipulation. Early work such as affordance-based imitation learning~\cite{lopes2007affordance} associated semantic parts (e.g., handles or lids) with manipulation trajectories, illustrating how abstract symbolic cues can support generalization across objects with similar functional structure.

\noindent \textbf{Multimodal Affordance.}
Recent work increasingly combines visual, linguistic, geometric, and spatial cues to support more holistic affordance reasoning. Representative examples include CLIPort~\cite{shridhar2022cliport}, which decouples semantic grounding from spatial localization, extensions to part-level affordances~\cite{yin2025partinstruct, geng2024sage}, and approaches that incorporate explicit 3D spatial reasoning for precise grounding~\cite{chen2024spatialvlm, yuan2025robopoint}. More recent efforts aim to learn unified and transferable affordance representations across objects and tasks~\cite{tang2025uad, kuang2025ram}.

\new{
Affordance-based planning offers an action-centric bridge from perception to manipulation, but predictions can be ambiguous, perception-sensitive, and unreliable in cluttered, novel, or long-horizon settings. Since affordances do not ensure geometric feasibility or executability, they typically require downstream planning and control.
}

\subsection{\new{3D Representation-based Planning}}
\label{subsec: 3d_as_planners}

Although 3D representations such as Gaussian Splatting and neural descriptor fields do not directly output control commands, they serve as mid-level planning interfaces by transforming perception into structured action-relevant representations, such as grasp candidates, spatial correspondences, or optimization objectives. In this sense, they bridge perception and action and naturally fall under high-level planning.

One line of work uses Gaussian Splatting as an editable and semantically enriched scene representation for manipulation. Splat-MOVER~\cite{shorinwa2025splat} distills open-vocabulary semantics into a 3DGS scene to propose grasp candidates for downstream planning, while object-aware and physically embodied variants extend Gaussian Splatting with object-centric reconstruction and physics coupling for dynamic interaction~\cite{li2024object, abou2025physically}. Building on this editability, RoboSplat~\cite{yang2025novel} synthesizes diverse demonstrations by directly manipulating reconstructed scenes, improving one-shot generalization of visuomotor policies. A complementary line represents scenes as implicit or descriptor fields in continuous 3D space. Neural Descriptor Fields learn SE(3)-equivariant representations for pose transfer and relational rearrangement~\cite{simeonov2022neural, simeonov2023se}, while F3RM and $D^3$Fields distill foundation-model features into 3D fields for language-conditioned grasping, placing, and zero-shot rearrangement~\cite{shen2023distilled, wang2025d}. More structured world models, such as Imagination Policy~\cite{huang2025imagination} and RoboEXP~\cite{jiang2025roboexp}, further use 3D fields and scene structure for downstream planning.

\new{
Overall, 3D representation-based planning enables rich spatial reasoning and language grounding, but depends heavily on accurate scene reconstruction and can degrade in dynamic, cluttered, or partially observed environments. Moreover, these representations provide structured guidance rather than executable actions, and therefore still require downstream planning or control modules.
}

%% file: tables/5_high-level_planner/high-level_planner.tex
\definecolor{catBlue}{HTML}{87CEFA}   
\definecolor{catGreen}{HTML}{8FBC8F}
\definecolor{catOrange}{HTML}{FF7F0E}
\definecolor{catPurple}{HTML}{9467BD}

\begin{figure*}[t]
\centering
\scriptsize

\forestset{
  taxo/.style={
    for tree={
      grow'=0,
      parent anchor=east,
      child anchor=west,
      anchor=west,
      edge path'={(!u.parent anchor) -- +(3pt,0) |- (.child anchor)},
      edge={draw, line width=0.6pt},
      rounded corners,
      draw,
      minimum height=4mm,
      inner xsep=4pt,
      inner ysep=2pt,
      s sep=3pt,
      l sep=10pt,
      align=left,
      font=\scriptsize
    },
    cat/.style={draw, rounded corners=2pt, minimum height=5mm, minimum width=15mm,
                 align=center, font=\bfseries},
    catBlue/.style={cat, fill=catBlue},
    catGreen/.style={cat, fill=catGreen},
    catOrange/.style={cat, fill=catOrange},
    catPurple/.style={cat, fill=catPurple},
    sub_1/.style={draw, rounded corners=2pt, fill=hidden-blue!32, text width=30mm, inner sep=4pt, align=center},
    sub/.style={draw, rounded corners=2pt, fill opacity=.5, text width=35mm, inner sep=4pt},
    item/.style={draw, rounded corners=2pt, fill=yellow!32, text width=72mm, 
                 align=left,inner sep=4pt}
  }
}

\begin{forest} taxo
[{\rotatebox{90}{\textbf{High-level Planner (\S\ref{sec: high_level_planner})}}}, draw, inner sep=2pt, fill opacity=.5,
  [{LLM-based Task Planning~\S\ref{subsec: llm-based_planners}},sub
    [{Planning \& Skill Selection}, sub_1
      [{SayCan~\cite{brohan2023can}, Grounded Decoding~\cite{huang2023grounded}, LLM-Planner~\cite{song2023llm}}, item]
    ]
    [{Enhanced Capabilities}, sub_1
      [{LLM+P~\cite{liu2023llm+}, REFLECT~\cite{liu2023reflect}, MALMM~\cite{singh2024malmm}, Polaris~\cite{wang2024polaris}}, item]
    ]
  ]
  [{MLLM-based Task Planning~\S\ref{subsec: mllm-based_planners}},sub
    [{End-to-End MLLMs}, sub_1
      [{PaLM-E~\cite{driess2023palm}, \new{VILA~\cite{hu2024look}}, PG-InstructBLIP~\cite{gao2024physically}}, item]
    ]
    [{Reasoning \& Cooperation}, sub_1
      [{EmbodiedGPT~\cite{mu2023embodiedgpt}, RoboBrain~\cite{ji2025robobrain}, Gemini Robotics~\cite{team2025gemini}}, item]
    ]
  ]
  [{\new{Programmatic Planning}~\S\ref{subsec: code_generation}},sub
    [{Language-to-Code}, sub_1
      [{Code as Policies~\cite{liang2023code}, ProgPrompt~\cite{singh2023progprompt}}, item]
    ]
    [{Demo-to-Code}, sub_1
      [{Demo2Code~\cite{wang2023demo2code}, SHOWTELL~\cite{murray2024teaching}, Statler~\cite{yoneda2024statler}}, item]
    ]
  ]
  [{\new{Geometric Constraint-based} \\ \new{Planning}~\S\ref{subsec: motion_planning}},sub
    [{Geometric Objective}, sub_1
      [{VoxPoser~\cite{huang2023voxposer}, CoPa~\cite{huang2024copa}}, item]
    ]
    [{Geometric Constraint}, sub_1
      [{ReKep~\cite{huang2025rekep}, GeoManip~\cite{tang2025geomanip}}, item]
    ]
  ]
  [{\new{Affordance-based Planning}~\S\ref{subsec: affordance_as_planner}},sub
    [{Geometric}, sub_1
      [{Ditto~\cite{jiang2022ditto}, GAPartNet~\cite{geng2023gapartnet}, CPM~\cite{liu2023composable}}, item]
    ]
    [{Visual}, sub_1
      [{Transporter Networks~\cite{zeng2021transporter}, VAPO~\cite{borja2022affordance}}, item]
    ]
    [{Semantic}, sub_1
      [{Early affordance learning~\cite{lopes2007affordance}}, item]
    ]
    [{Multimodal}, sub_1
      [{CLIPort~\cite{shridhar2022cliport}, RoboPoint~\cite{yuan2025robopoint}, MOKA~\cite{liu2024moka}, ManipLLM~\cite{li2024manipllm}}, item]
    ]
  ]
  [{\new{3D Representation-based} \\ \new{Planning}~\S\ref{subsec: 3d_as_planners}},sub
    [{Gaussian Splatting}, sub_1
      [{MSGField~\cite{sheng2024msgfield}, RoboSplat~\cite{yang2025novel}}, item]
    ]
    [{Implicit or Descriptor Fields}, sub_1
      [{{NDF~\cite{simeonov2022neural}, F3RM~\cite{shen2023distilled}, $D^{3}$Fields~\cite{wang2025d}}}, item]
    ]
    [{Generative World Models}, sub_1
      [{Imagination Policy~\cite{huang2025imagination}}, item]
    ]
    [{Structured Scene Graphs}, sub_1
      [{RoboEXP~\cite{jiang2025roboexp}}, item]
    ]
  ]
]
\end{forest}
\vskip -0.05in
\caption{
Taxonomy of high-level planning approaches for robotic manipulation, organized into six categories: LLM-based task planning, MLLM-based task planning, programmatic planning, geometric constraint-based planning, affordance-based planning, and 3D representation-based planning.
}
\label{fig: high-level_planner_taxonomy_colored}
\end{figure*}

%% file: figures/5_high-level_planner/high-level_planner.tex
\begin{figure*}[t]
\centering
\includegraphics[width=0.9\textwidth]{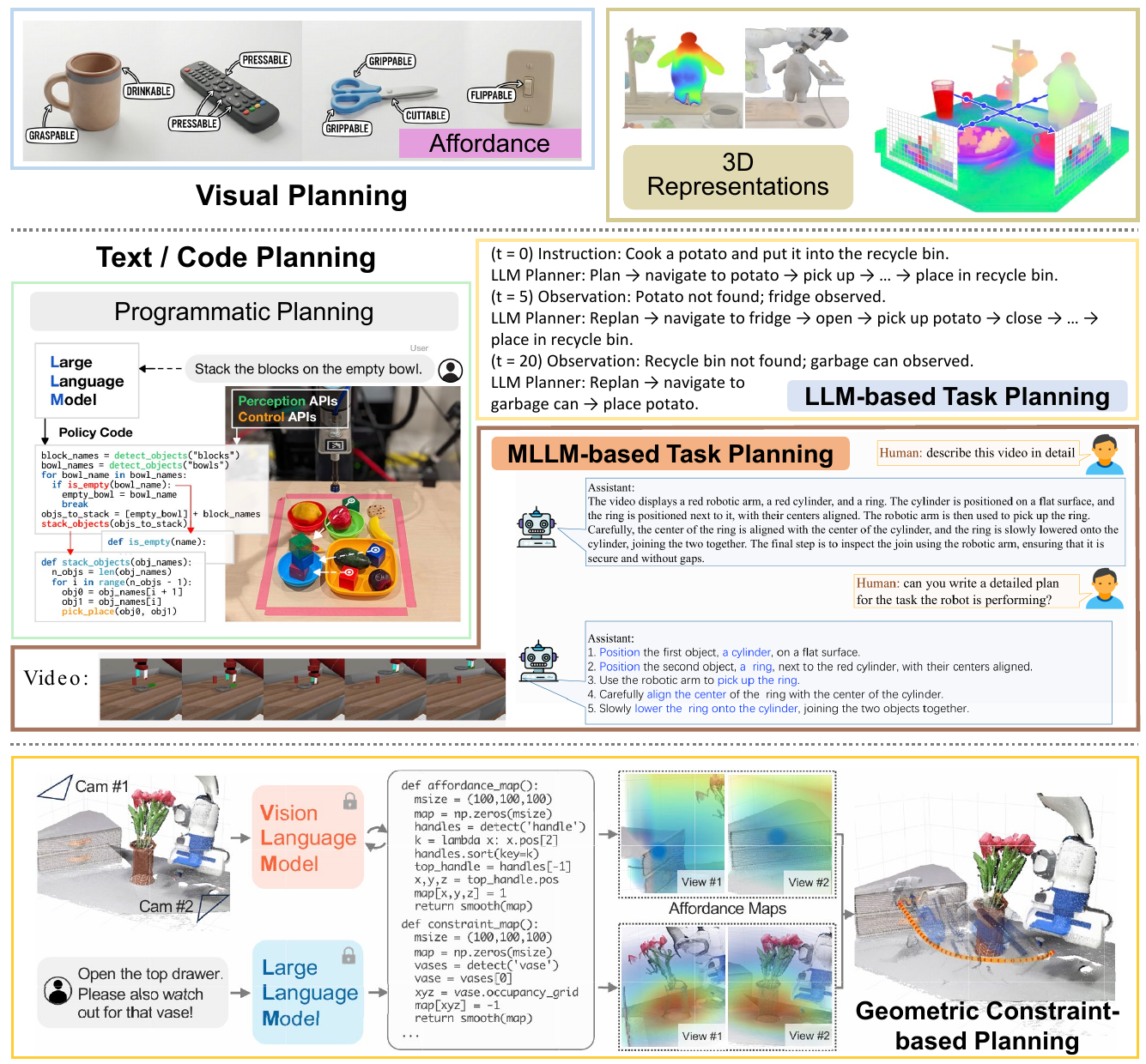}  
\vskip -0.05in
\caption{Overview of the taxonomy of high-level planners, highlighting six core components: LLM- and MLLM-based, programmatic, geometric constraint-based, affordance-based, and 3D representation-based planning. \new{Figure are adapted from~\cite{song2023llm, mu2023embodiedgpt, liang2023code, huang2023voxposer, jiang2022ditto, shen2023distilled}.}
}
\label{fig: high-level_planner}
\end{figure*}

%% file: sections/6-low_level_controller/low-level_learning-based_control.tex
\section{Low-Level Action Modeling}
\label{sec: low_level_control}

Low-Level Action Modeling concerns how observations and task specifications are transformed into executable actions, thereby grounding high-level reasoning in physical execution. 
\new{
Unlike High-Level Planning, which produces structured guidance, this layer outputs motion commands or action sequences that can be executed by downstream controllers. For representative vision-language-action models, this process can be written as
\begin{equation}
A_{\text{action}} = f_{\text{act}}(o,l \mid A_{\text{plan}}),
\end{equation}
where $o$ denotes observations, $l$ denotes task specifications, and $A_{\text{plan}}$ is optional. More broadly, this formulation also applies to a wider family of learning-based action models beyond VLA systems.
}
Following the learning pipeline of this layer, as shown in Figure~\ref {fig: low_level_control}, we organize the literature into three components: input modeling, latent learning, and policy learning. Together, these components provide a unified perspective on how learning-based action models couple perception, intermediate representation, and action generation.

\input{sections/6-low_level_controller/learning_strategy}

\input{sections/6-low_level_controller/input_learning}

\input{sections/6-low_level_controller/latent_learning}

\input{sections/6-low_level_controller/policy_learning}

%% file: sections/6-low_level_controller/learning_strategy.tex

\input{figures/6_low-level_control/summary}

\subsection{Learning Strategies}
\label{sec: learning_strategy}

Learning Strategies describe how supervision, feedback, and optimization signals are used to acquire manipulation policies. Rather than specifying what is perceived or how actions are parameterized, they determine how experience is converted into executable behavior, thereby affecting data efficiency, generalization, and robustness. Existing methods mainly differ in the extent to which they rely on environmental interaction, expert demonstrations, or auxiliary training objectives. As shown in Figure~\ref{fig: learning_strategy}, we group them into three categories: reinforcement learning (RL), imitation learning (IL), and learning with auxiliary tasks.

\subsubsection{Reinforcement Learning}
\label{subsubsec: rl}
In robotic manipulation, RL is an important paradigm for acquiring complex control policies through trial-and-error interaction with the environment. This section briefly reviews RL-based methods for robotic manipulation and groups them into two broad classes: model-free and model-based approaches, depending on whether an explicit or learned dynamics model is used during learning.

\paragraph{Model-free Methods}
Model-free reinforcement learning (RL) learns manipulation policies directly from interaction without relying on explicit environment models. Although broadly applicable to high-dimensional control, such methods are often limited by poor sample efficiency. Recent work therefore increasingly combines model-free RL with large-scale pretraining and efficient post-training, using offline robotic data, pretrained policies, or reinforcement-learning-based adaptation to improve generalization and online refinement~\cite{kalashnikov2018scalable, kumar2022pre, guo2025improving}. Overall, the role of model-free RL in robotic manipulation has gradually shifted from training task-specific policies from scratch toward scalable adaptation of pretrained generalist models.

\paragraph{Model-based Methods}
Model-based reinforcement learning improves data efficiency by leveraging explicit or learned models of environment dynamics for imagination, planning, or gradient-based optimization. Representative directions include latent world models for imagined rollouts, planning-oriented methods that integrate learned dynamics into model-predictive control, and differentiable simulation frameworks that enable gradient-based policy refinement~\cite{hafner2020dream, wu2023daydreamer, hansen2022temporal, xing2025stabilizing}. While these methods offer advantages in sample efficiency and interpretability, their effectiveness ultimately depends on model accuracy, which remains challenging in contact-rich and high-dimensional manipulation settings.

\input{figures/6_low-level_control/learning_strategy}


\subsubsection{Imitation Learning}
\label{subsubsec: il}

IL acquires manipulation behaviors from expert demonstrations, avoiding explicit reward design and reducing extensive trial-and-error interaction~\cite{schaal1999imitation}. From early control-theoretic formulations to modern deep visuomotor policies, it has become a widely used paradigm for robotic manipulation. Recent approaches further benefit from large-scale pretraining and foundation-model priors, improving multimodal perception, generalization, and transfer across tasks and embodiments~\cite{dalal2023imitating}.

\paragraph{Imitation from Action}
When expert state-action pairs are available, behavior cloning (BC) remains the most direct and widely used approach, learning policies that map observations to low-level actions~\cite{huang2019non, bai2025rethinking}. To improve long-horizon performance and robustness, recent methods often incorporate hierarchical structure, skill decomposition, pose-level action abstractions, or planner-guided supervision~\cite{katz2016imitation, sun2025exploring, mcdonald2022guided}. Beyond direct action cloning, some approaches instead infer the objectives underlying expert behavior through inverse or adversarial formulations, enabling policies to imitate demonstrations at the level of rewards, constraints, or state distributions~\cite{englert2017inverse, ho2016generative}.

\paragraph{Imitation from Observation (IfO)}
IfO removes action supervision and learns from state or visual trajectories alone, requiring the agent to infer executable behavior from observed transitions. Existing methods typically recover reward-like signals or objectives from observations, align demonstrations and agent behavior in learned representation spaces, or extract intermediate goals from videos for downstream control~\cite{alakuijala2023learning, ma2023vip, haldar2025point}. These approaches are particularly attractive when action labels are unavailable or difficult to transfer across embodiments, although the lack of direct action supervision often makes learning more ambiguous and challenging.

\subsubsection{Learning with Auxiliary Tasks}

Auxiliary-task learning enhances manipulation policies with additional self-supervised or weakly supervised objectives, providing structured learning signals beyond sparse task rewards or demonstrations. One major direction is \emph{world modeling}, where agents learn predictive representations of environment dynamics to support planning, control, or latent rollout-based optimization~\cite{zhao2024vlmpc, lu2024manigaussian, du2024video}. Another direction focuses on \emph{goal extraction and representation shaping}, transforming raw observations into compact and actionable abstractions such as masks, keypoints, waypoints, motion-centric representations, or language-grounded subgoals~\cite{gao2023k, shi2023waypoint, bharadhwaj2024track2act, jia2024chain}. In addition, auxiliary objectives such as multimodal alignment and reconstruction-based pretraining help improve spatial and temporal consistency in learned representations~\cite{nair2023r3m, xiao2022masked, ze2023gnfactor}. Overall, these methods improve data efficiency and representation quality by converting weak or indirect supervision into more informative training signals.

%% file: figures/6_low-level_control/summary.tex
\begin{figure}[t]
\centering
\includegraphics[width=1.0\linewidth]{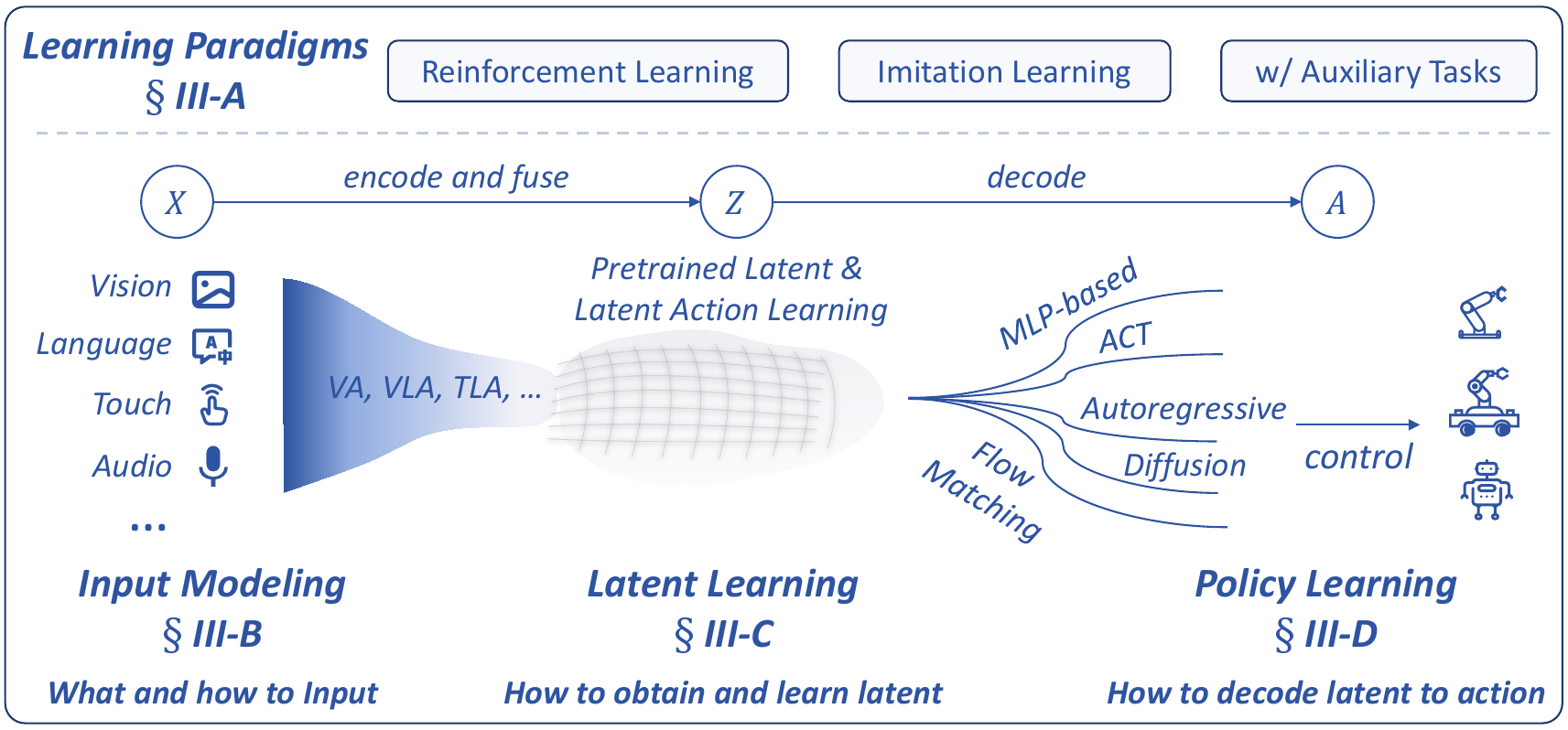}  
\vskip -0.05in
\caption{Overview of the taxonomy of learning-based Low-Level Action Modeling. It is organized into three components: Input Modeling, which specifies how multimodal inputs are represented and utilized; Latent Learning, which learns intermediate representations; and Policy Learning, which maps them to executable actions. Representative learning paradigms include reinforcement learning, imitation learning, and learning with auxiliary tasks.}
\label{fig: low_level_control}
\end{figure}

%% file: figures/6_low-level_control/learning_strategy.tex
\begin{figure}[t]
\centering
\includegraphics[width=0.90\linewidth]{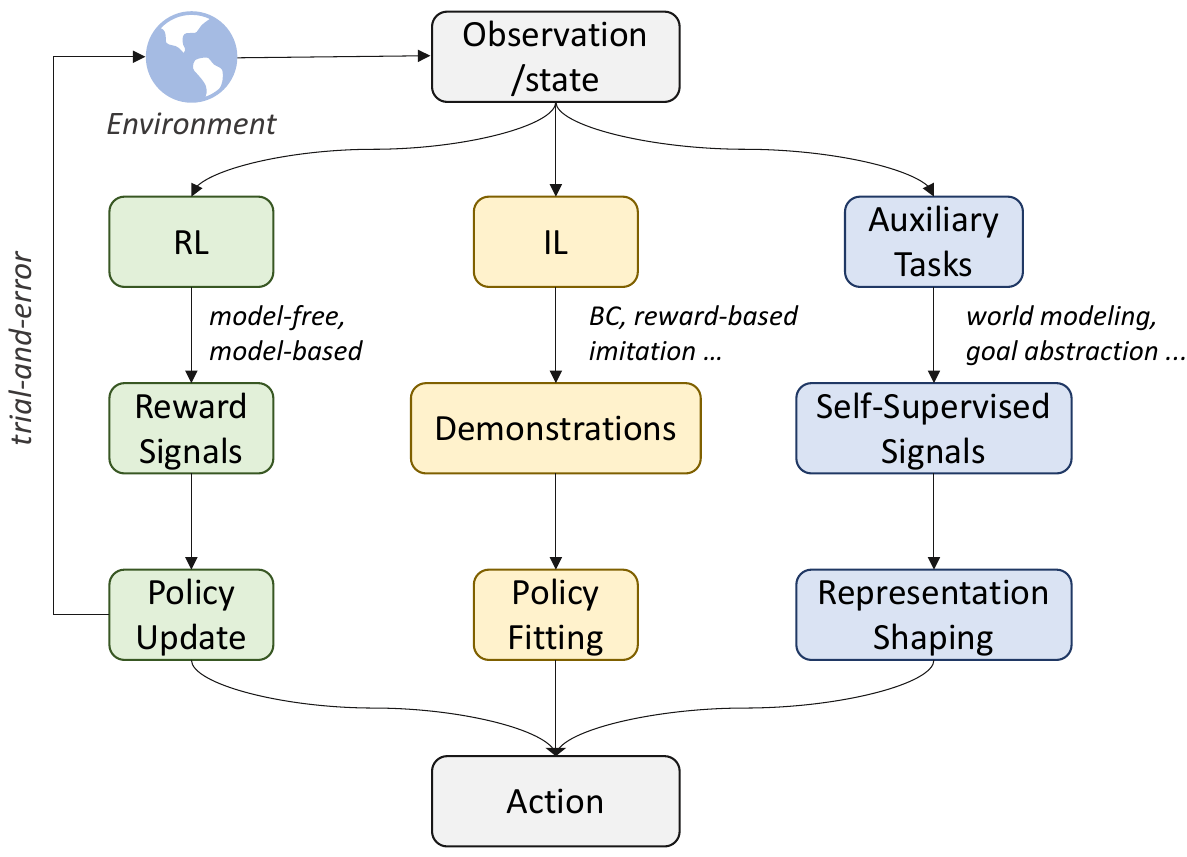}  
\vskip -0.05in
\caption{Comparison of learning paradigms for low-level robotic manipulation.
RL optimizes policies through trial-and-error interaction using reward signals, IL learns direct policy mappings from expert demonstrations, and auxiliary-task learning shapes representations via self-supervised objectives such as world modeling and goal abstraction.
}
\label{fig: learning_strategy}
\end{figure}

%% file: sections/6-low_level_controller/input_learning.tex
\subsection{Input Modeling}
\label{sec: input_modeling}

Input modeling defines how a robot perceives the world by specifying which sensory modalities are used and how they are encoded before policy learning. It includes the selection, alignment, and fusion of multimodal observations, such as vision, language, touch, and force, and transforms raw inputs into structured representations for control and decision making. Effective input modeling preserves essential spatial, temporal, and semantic information, providing a reliable perceptual foundation for robust manipulation across diverse tasks.

\subsubsection{Vision--Action Models}

Vision--Action models tightly couple visual perception with action generation, learning direct mappings from visual observations to motor commands. Rather than relying on explicit symbolic planning, these methods typically encode task-relevant information implicitly within end-to-end visuomotor policies. Recent advances have expanded this paradigm from early convolutional designs to transformer- and diffusion-based frameworks with improved temporal modeling and policy expressiveness.

\paragraph{2D Vision as Input}
Most Vision--Action models operate on 2D visual observations, typically RGB images from single or multi-view cameras~\cite{chi2023diffusion, wang2024scaling, prasad2024consistency}. Representative methods include diffusion-based policies such as Diffusion Policy~\cite{chi2023diffusion}, hierarchical extensions such as HDP~\cite{wang2025hierarchical}, and transformer-based models such as HPT~\cite{wang2024scaling}. These methods have demonstrated strong scalability and empirical effectiveness, but they generally lack explicit geometric grounding. As a result, they may struggle with precise spatial reasoning, contact geometry, and robust occlusion handling in complex physical environments.

\paragraph{3D Vision as Input}
To address these limitations, a growing line of work incorporates explicit 3D visual information into Vision--Action models~\cite{goyal2023rvt, wang2024gendp, ze20243d}. By lifting 2D observations into 3D scene representations, these methods align action generation more directly with scene geometry. Representative examples include RVT~\cite{goyal2023rvt}, which infers 3D action targets through multi-view transformations, GenDP~\cite{wang2024gendp}, which conditions diffusion policies on reconstructed 3D semantic fields, and DP3~\cite{ze20243d}, which integrates point clouds into the policy input. Although these methods substantially improve spatial reasoning and execution accuracy, most remain primarily perception-driven and still lack strong semantic grounding, leaving generalization across open-ended goals and instructions an open challenge.

\subsubsection{Vision-Language-Action Models}
\label{subsec: vla}

\input{tables/6_low-level_controller/vla_tax}

Recent advances in VLA models have established a unified paradigm for mapping multimodal perception to executable robotic behaviors. Unlike earlier vision-only or language-conditioned controllers, VLAs integrate semantic grounding, spatial reasoning, and sequential action generation within a single framework. As summarized in Figure~\ref{fig: vla-taxonomy}, we organize existing methods by input modality (2D vs.\ 3D) and methodological orientation (model-oriented vs.\ model-agnostic), which together reveal the main design trends of modern VLA systems.

\paragraph{2D Vision as Input}
Most existing VLA models primarily rely on 2D RGB observations, sometimes combined with multi-view images, as their main perceptual input. Under this setting, methods can be broadly divided into model-oriented approaches and model-agnostic strategies. We first review the model-oriented approaches:

\noindent \textbf{Non-LLM-based VLA.}  
Early VLA systems directly mapped visual observations and textual commands to actions through language-conditioned sequence models. RT-1~\cite{brohan2023rt} introduced a transformer-based discretized action policy that scaled to thousands of demonstrations, enabling diverse manipulation skills. Subsequent work such as VIMA~\cite{jiang2022vima} and HULC~\cite{mees2022matters} extended this paradigm to compositional instructions and multimodal imitation. 
While effective in relatively closed settings, these approaches remained limited in generalization.

\noindent \textbf{LLM/VLM-based VLA.}  
The emergence of LLMs and VLMs has introduced substantially stronger semantic priors into VLA systems. RT-2~\cite{zitkovich2023rt} co-trains web-scale vision-language data with robot trajectories, enabling improved semantic generalization and the transfer of web-scale knowledge to robotic control. RoboFlamingo~\cite{li2024vision} demonstrates the effectiveness of leveraging pretrained VLMs through lightweight action interfaces, whereas OpenVLA~\cite{kim2025openvla} scales end-to-end VLA training over diverse robot demonstrations. Building on these advances, $\pi_0$~\cite{black2025pi_0} and $\pi_{0.5}$~\cite{intelligence2025pi_} further improve robustness and generalization through stronger continuous action modeling and large-scale multimodal training. Despite this progress, a persistent gap remains between high-level semantic reasoning and precise continuous motor execution.

\noindent \textbf{Latent Learning for VLA.}
A parallel line of work introduces latent action spaces that compress task-relevant motion into compact intermediate representations. UniVLA~\cite{bu2025univla} learns language-conditioned latent actions in visual feature space and decodes them into embodiment-specific controls. villa-X~\cite{chen2026villa} jointly learns vision-based latent actions and proprioceptive forward dynamics within a diffusion framework, while AgiBot-GO1~\cite{bu2025agibot} inserts a latent-action planner between a vision-language backbone and low-level controllers. Collectively, these methods use latent actions to improve efficiency and transferability.

\noindent \textbf{Action Quantization for VLA.}
Another direction discretizes continuous trajectories into action tokens, reducing action-space complexity and stabilizing training. VQ-VLA~\cite{wang2025vq} and MiniVLA~\cite{belkhale2024minivla} learn quantized motor codebooks, while \newer{FAST}~\cite{pertsch2025fast} maps control signals into frequency-domain tokens. Such quantized interfaces help bridge high-level intent and continuous execution, although maintaining semantic consistency and avoiding codebook collapse remain challenging.

\noindent \textbf{Dual- and Multi-System VLA.} 
Inspired by dual-process cognitive theories, recent work proposes VLA architectures that separate fast reactive control from slower deliberative reasoning. Typically, a lightweight policy handles low-latency execution, while a larger vision-language model performs higher-level planning, reasoning, or memory management. Representative examples include LCB~\cite{shentu2024llms}, HiRT~\cite{zhang2025hirt}, RationalVLA~\cite{song2026dual}, HiRobot~\cite{shi2025hi}, and OpenHelix~\cite{cui2025openhelix}, as well as multi-system extensions such as TriVLA~\cite{liu2025trivla} and G0~\cite{jiang2025galaxea}. These architectures improve robustness and flexibility, but also introduce new challenges in arbitration, credit assignment, and cross-system consistency.

Second, we consider \emph{model-agnostic strategies}, which improve training or inference without modifying the underlying policy architecture.

\noindent \textbf{Inference-Time Optimization.}
Methods such as RoboMonkey~\cite{kwok2025robomonkey} and CronusVLA~\cite{li2026cronusvla} refine action selection at inference through voting, sampling, or calibration strategies. These approaches are particularly attractive because they can improve robustness without retraining and can be readily applied across diverse VLA backbones.

\noindent \textbf{Reinforcement Learning and Post-training.} 
Recent work explores interactive post-training of pretrained VLAs by introducing RL signals tailored to multimodal, long-horizon policies. SimpleVLA-RL~\cite{li2025simplevla} combines verifiable outcome-based rewards with group-level policy optimization, enabling stable refinement from sparse task feedback. VLA-RL~\cite{lu2025vla} emphasizes large-scale rollout-based RL to better align pretrained action decoders with execution-level requirements in generalist manipulation. RIPT-VLA~\cite{tan2025interactive} formulates post-training as lightweight interaction, using rollout feedback to improve performance without relying on a critic. ConRFT~\cite{chen2025conrft} bridges offline imitation and online reinforcement learning through a consistency-based objective, reducing distribution shift during refinement.

\noindent \textbf{Learning with Auxiliary Tasks.}
Auxiliary-task designs introduce structured supervision beyond action prediction to enrich intermediate reasoning and decision making. Representative directions include chain-of-thought-style intermediate reasoning~\cite{zawalski2025robotic, zhao2025cot}, explicit subgoal extraction~\cite{zheng2025tracevla, tian2025predictive}, and reconstruction-based consistency objectives~\cite{song2026reconvla}. By making intermediate structures more explicit, these approaches can improve reasoning horizon and execution transparency.

\noindent \textbf{Efficiency, Robustness, and Long-Horizon Generalization.}
Another practical line of work focuses on making VLAs more deployable in real-world settings. Methods such as TinyVLA~\cite{wen2025tinyvla}, VLA-Cache~\cite{xu2025vla}, and CEED-VLA~\cite{song2026ceed} improve efficiency through lightweight design, adaptive computation, or parallel decoding, while other approaches improve robustness through safety alignment, failure awareness, or perturbation defense~\cite{gu2025safe, xu2025can, zhou2025badvla}. Long-VLA~\cite{fan2025long} further targets long-horizon and compositional manipulation by improving temporal consistency and memory. Together, these efforts reflect a shift from single-step policy prediction toward stable and reliable long-horizon deployment.

\input{tables/6_low-level_controller/tactile}

\paragraph{3D Vision as Input}
Compared with 2D observations, 3D representations provide richer spatial grounding for contact-rich manipulation and long-horizon planning. Since most vision-language backbones are pretrained on large-scale 2D image-text corpora and lack intrinsic 3D understanding, increasing effort has gone into equipping VLA systems with explicit 3D perception. Existing methods can be broadly grouped into \emph{model-oriented} approaches, which directly incorporate 3D information into the policy architecture, and \emph{model-agnostic} strategies, which improve 3D VLAs without modifying the backbone.

\noindent \textbf{3D Embedding and Fusion.}
A primary model-oriented direction augments VLAs with explicit 3D embeddings, such as point clouds, depth maps, or voxel grids, and fuses them with 2D vision-language features. SpatialVLA~\cite{qu2025spatialvla} and GeoVLA~\cite{sun2025geovla} improve spatial precision through 3D positional or point-based geometric embeddings, while FP3~\cite{yang2025fp3} and RoboMM~\cite{yan2025robomm} explore large-scale 3D training and multimodal pretraining with explicit 3D inputs. Overall, these methods improve spatial awareness and robustness, though often at higher data and computation cost.

\noindent \textbf{Spatial Alignment.}
Another line of work addresses occlusion and viewpoint ambiguity through explicit spatial alignment and multi-view reasoning. BridgeVLA~\cite{li2025bridgevla} improves geometric consistency under partial visibility by aligning point clouds across viewpoints via cross-view heatmap prediction.

\noindent \textbf{3D Reasoning and Prediction.}
A third direction incorporates explicit 3D reasoning and predictive modeling to improve spatial understanding and long-horizon manipulation. 3D-VLA~\cite{zhen20243d} employs language-guided point-cloud diffusion to predict future scene states, while Evo-0~\cite{lin2025evo}, ChainedDiffuser~\cite{xian2023chaineddiffuser}, and SGR~\cite{zhang2023universal} introduce complementary mechanisms for implicit spatial reasoning, joint keypose and trajectory prediction, and semantic-geometric representation learning, respectively. Together, these approaches enrich VLA policies with structured 3D priors that support more spatially grounded action generation and planning.

\noindent \textbf{Model-Agnostic Strategies.}
Compared with the growing model-oriented literature, model-agnostic enhancements for 3D VLAs remain relatively limited. Existing efforts mainly focus on inference-time view selection or feasibility filtering, as exemplified by TVVE~\cite{bai2025learning}, while more systematic advances in calibration, consensus decoding, caching, and post-training for 3D settings are still scarce. This makes efficient inference and adaptation for 3D VLAs an important open direction.

Across both 2D and 3D modalities, recent VLA research is moving toward richer representations and more structured action generation. In 2D settings, this mainly appears through stronger LLM/VLM priors, latent abstractions, and multi-system reasoning, while 3D approaches emphasize spatial embeddings, alignment mechanisms, and explicit world models. 
\new{Nevertheless, current VLA systems still face important challenges in robustness, efficiency, and generalization, especially under long-horizon tasks and real-world deployment. Promising directions include more standardized 3D representations, hybrid architectures that combine reactive control with deliberative planning, and joint training across 2D and 3D modalities.}

\subsubsection{Tactile-based Action Models}
Tactile sensing provides fine-grained feedback on contact, geometry, and material properties, making it particularly important for precise and contact-rich manipulation. Unlike vision, tactile signals directly reflect physical interaction and remain informative under occlusion, slippage, and uncertainty. Recent advances in tactile sensing and representation learning have therefore motivated a growing body of work that incorporates tactile feedback into action models to improve robustness, accuracy, and adaptability. We summarize this line of work in Figure~\ref{fig: tactile}.

\noindent \textbf{Tactile Latent Learning.}
A first line of work focuses on learning transferable tactile representations. CLTP~\cite{ma2025cltp} aligns tactile geometry with language through contrastive pretraining, while Sparsh~\cite{higuera2025sparsh} learns generalizable tactile embeddings from large-scale self-supervised visuotactile data.

\noindent \textbf{Tactile-Action Models.}
Another direction directly incorporates tactile feedback into action generation. Representative examples include Feel the Force~\cite{adeniji2025feel}, which transfers contact-force patterns from human tactile-glove demonstrations to robot policies, Seq2Seq Imitation~\cite{yang2023seq2seq}, which models tactile signals as temporal sequences, and RoboPack~\cite{ai2024robopack}, which integrates tactile sensing into learned dynamics for MPC.

\noindent \textbf{Tactile-Vision-Action Models.} 
A growing body of work studies visuotactile fusion for robotic manipulation. Representative methods such as T-DEX~\cite{guzey2023dexterity}, RotateIt~\cite{qi2023general}, and Multimodal-SeeThrough~\cite{ablett2024multimodal} combine vision and touch to improve dexterous manipulation and imitation learning, while VTTB~\cite{gu2024vttb} and VITaL~\cite{george2025vital} show the value of joint visuotactile pretraining. Reactive Diffusion Policy~\cite{xue2025reactive} further demonstrates responsive contact-rich control with a slow--fast visuotactile diffusion architecture.

\noindent \textbf{Tactile-Language-Action Models.} 
Several recent works align tactile perception with natural language. TLA~\cite{hao2025tla} learns tactile-language-action mappings from sequential tactile signals, while Octopi~\cite{yu2024octopi} uses tactile-language modeling to reason about object properties beyond visual appearance.

\noindent \textbf{Tactile-Vision-Language-Action Models.} More recently, tactile sensing has been incorporated into full VLA pipelines. Tactile-VLA~\cite{huang2025tactile} extends VLA models with tactile inputs to improve physical interaction, VTLA~\cite{zhang2025vtla} combines vision and touch with language-conditioned preference learning for robust insertion, and Touch Begins~\cite{zhao2025touch} adopts a two-stage strategy that uses vision-language models for localization and tactile feedback for execution.

\subsubsection{Extra Modalities as Input}
Beyond vision and language, force and audio provide complementary physical cues that are valuable for contact-rich and partially observable manipulation. Research on force sensing has progressed from early imitation-learning approaches that jointly modeled motion and force~\cite{kormushev2011imitation} and bilateral teleoperation for position--force coordination~\cite{adachi2018imitation} to more recent force-aware visuomotor policies and VLA extensions for physically grounded control~\cite{he2025foar, yu2026forcevla}.
Audio provides a complementary sensing modality that captures contact events and state transitions often inaccessible to vision, supporting manipulation under occlusion~\cite{du2022play}, feedback for hidden interactions~\cite{li2023see}, audio-visual imitation from in-the-wild demonstrations~\cite{liu2025maniwav}, and audio-enhanced VLA models~\cite{zhao2025vlas}. Together, these modalities enrich action models with physically informative signals and improve robustness in real-world manipulation.

%% file: tables/6_low-level_controller/vla_tax.tex
\tikzset{
    my-box/.style={
        rectangle,
        draw=black,
        rounded corners,
        text opacity=1,
        minimum height=1.5em,
        minimum width=5em,
        inner sep=2pt,
        align=center,
        fill opacity=.5,
    },
    leaf-2d/.style={
        my-box, 
        fill=yellow!32, 
        text=black,
        font=\normalsize,
        inner xsep=5pt,
        inner ysep=4pt,
        align=left,
        text width=45em,
    },
    leaf-3d/.style={
        my-box, 
        fill=hidden-blue!32, 
        text=black,
        font=\normalsize,
        inner xsep=5pt,
        inner ysep=4pt,
        align=left,
        text width=45em,
    },
    leaf-summary/.style={
        my-box, 
        fill=yellow!32, 
        text=black,
        font=\normalsize,
        inner xsep=5pt,
        inner ysep=4pt,
        align=left,
        text width=45em,
    },
}

\begin{figure*}[t]
\centering
\resizebox{\textwidth}{!}
{
    \begin{forest}
        forked edges,
        for tree={
            grow=east,
            reversed=true,
            anchor=base west,
            parent anchor=east,
            child anchor=west,
            base=left,
            font=\large,
            rectangle,
            draw=black,
            rounded corners,
            align=center,
            minimum width=4em,
            edge+={darkgray, line width=1pt},
            s sep=3pt,
            inner xsep=2pt,
            inner ysep=4pt,
            line width=1.1pt,
            ver/.style={rotate=90, child anchor=north, parent anchor=south, anchor=center, inner xsep=5pt},
        },
        where level=1{text width=10em, font=\normalsize, align=center, inner xsep=5pt, inner ysep=3pt}{},
        where level=2{text width=7em, font=\normalsize, align=center, inner xsep=3pt}{},
        where level=3{text width=15em, font=\normalsize, align=left, inner xsep=5pt}{},
        where level=4{font=\normalsize, align=left}{},
[{Vision-Language-Action\\ Models (\S\ref{subsec: vla})}, ver
    [{2D Vision as Input}, ver, text width=8em
        [{Model-oriented}
            [{Non-LLM-based VLA}
                [{\eg RT-1~\cite{brohan2023rt}, Octo~\cite{ghosh2024octo}, VIMA~\cite{jiang2022vima}, HULC~\cite{mees2022matters}, 
                RoboBERT~\cite{wang2025robobert}, Dita~\cite{hou2025dita}, OTTER~\cite{huang2025otter}, \\ RoboGround~\cite{huang2025roboground}, BAKU~\cite{haldar2024baku}, 
                MDT~\cite{reuss2024multimodal}, 
                GR-1~\cite{wu2024unleashing}}, leaf-2d]
            ]
            [{LLM/VLM-based VLA}
                [{\eg RT-2~\cite{zitkovich2023rt}, RoboFlamingo~\cite{li2024vision}, OpenVLA~\cite{kim2025openvla}, OpenVLA-OFT~\cite{kim2025fine}, $\pi$0~\cite{black2025pi_0}, $\pi$0.5~\cite{intelligence2025pi_},\\ 
                UniVLA~\cite{bu2025univla}, 
                DexVLA~\cite{wen2025dexvla}, CogACT~\cite{li2024cogact}, Diffusion-VLA~\cite{wen2025diffusion}
                }, leaf-2d]
            ]
            [{Latent Learning for VLA}
            [{\eg AgiBot-GO1~\cite{bu2025agibot}, UniVLA~\cite{bu2025univla}, villa-X~\cite{chen2026villa}, LaRA-VLA~\cite{bai2026latent}}, leaf-2d]
            ]
            [{Action Quantization for VLA}
                [{\eg VQ-VLA~\cite{wang2025vq},  MiniVLA~\cite{belkhale2024minivla}, FAST~\cite{pertsch2025fast}}, leaf-2d]
            ]
            [{Dual- and Multi-system VLA}
                [{\eg LCB~\cite{shentu2024llms}, PIVOT-R~\cite{zhang2024pivot}, RoboDual~\cite{bu2024towards}, HiRT~\cite{zhang2025hirt}, RationalVLA~\cite{song2026dual}, TriVLA~\cite{liu2025trivla}, \\G0~\cite{jiang2025galaxea}, 
                OpenHelix~\cite{cui2025openhelix},  HiRobot~\cite{shi2025hi}}, leaf-2d]
            ]
        ]
        [{Model-agnostic}
            [{Inference-time Optimization}
                [{\eg RoboMonkey~\cite{kwok2025robomonkey}, CronusVLA~\cite{li2026cronusvla}, OC-VLA~\cite{zhang2026grounding}, PD-VLA~\cite{song2025pd}, CEED-VLA~\cite{song2026ceed}}, leaf-2d]
            ]
            [{Reinforcement Learning\\ and Post-training}
                [{\eg VLA-RL~\cite{lu2025vla}, RIPT-VLA~\cite{tan2025interactive}, ConRFT~\cite{chen2025conrft}, ThinkAct~\cite{huang2025thinkact}, 
                 V-GPS~\cite{nakamoto2025steering}, RLRC~\cite{chen2025rlrc}}, leaf-2d]
            ]
            [{Learning with Auxiliary Tasks}
                [{\eg ECoT~\cite{zawalski2025robotic}, CoT-VLA~\cite{zhao2025cot}, TraceVLA~\cite{zheng2025tracevla}, Seer~\cite{tian2025predictive}, 
                ReconVLA~\cite{song2026reconvla},  FlowVLA~\cite{zhong2025flowvla},  \\ GraspVLA~\cite{deng2025graspvla}, $\pi$0.5~\cite{intelligence2025pi_},  UP-VLA~\cite{zhang2025up},  RT-H~\cite{belkhale2024rt}, 
                LLARVA~\cite{niu2024llarva}, GO-1~\cite{bu2025agibot}}, leaf-2d]
            ]
            [{Efficiency}
            [{\eg VLA-Adapter~\cite{wang2026vla}, 
            TinyVLA~\cite{wen2025tinyvla}, 
            EfficientVLA~\cite{yang2025efficientvla}, 
            PD-VLA~\cite{song2025pd}, 
            \\ VLA-Cache~\cite{xu2025vla}, 
            CEED-VLA~\cite{song2026ceed}
            }, leaf-2d]
        ]
        [{Robustness}
            [{\eg SAFE~\cite{gu2025safe}, 
            BadVLA~\cite{zhou2025badvla}, 
            FAIL-Detect~\cite{xu2025can}, 
            }, leaf-2d]
        ]
            [{Long-horizon Generalization}
                [{\eg LongVLA~\cite{fan2025long}}, leaf-2d]
            ]
        ]
    ]
    [{3D Vision as Input}, ver, text width=8em
        [{Model-oriented}
            [{3D Embedding and Fusion}
                [{\eg SpatialVLA~\cite{qu2025spatialvla},  GeoVLA~\cite{sun2025geovla}, FP3~\cite{yang2025fp3}, RoboMM~\cite{yan2025robomm}}, leaf-3d]
            ]
            [{Spatial Alignment and\\ Multi-view Guidance}
                [{\eg BridgeVLA~\cite{li2025bridgevla}}, leaf-3d]
            ]
            [{3D Reasoning and Prediction}
                [{\eg  Evo-0~\cite{lin2025evo}, 
                ChainedDiffuser~\cite{xian2023chaineddiffuser}, SGR~\cite{zhang2023universal}}, leaf-3d]
            ]
        ]
        [{Model-agnostic}
            [{Task-aware view planning (\eg TVVE~\cite{bai2025learning})}, leaf-3d]
        ]
    ]
]
    \end{forest}
}
\vskip -0.05in
\caption{A taxonomy of VLA models organized along two dimensions, input modality (2D vs.\ 3D) and methodological orientation (model-oriented vs.\ model-agnostic), highlighting representative directions in architectural design, training, and inference.
}
\label{fig: vla-taxonomy}
\end{figure*}

%% file: tables/6_low-level_controller/tactile.tex
\tikzset{
    my-box/.style={
        rectangle,
        draw=black,
        rounded corners,
        text opacity=1,
        minimum height=1.5em,
        minimum width=5em,
        inner sep=2pt,
        align=center,
        fill opacity=.5,
    },
    leaf-my/.style={
        my-box, 
        fill=yellow!32, 
        text=black,
        font=\normalsize,
        inner xsep=5pt,
        inner ysep=4pt,
        align=left,
        text width=45em,
    },
    leaf-3d/.style={
        my-box, 
        fill=hidden-blue!32, 
        text=black,
        font=\normalsize,
        inner xsep=5pt,
        inner ysep=4pt,
        align=left,
        text width=45em,
    }
}
\begin{figure*}[t]
\centering
\resizebox{\textwidth}{!}{
    \begin{forest}
        forked edges,
        for tree={
            grow=east,
            reversed=true,
            anchor=base west,
            parent anchor=east,
            child anchor=west,
            base=left,
            font=\large,
            rectangle,
            draw=black,
            rounded corners,
            align=center,
            minimum width=4em,
            edge+={darkgray, line width=1pt},
            s sep=3pt,
            inner xsep=2pt,
            inner ysep=4pt,
            line width=1.1pt,
            ver/.style={rotate=90, child anchor=north, parent anchor=south, anchor=center, inner xsep=5pt},
        },
        where level=1{text width=18em, font=\normalsize,align=center, inner xsep=5pt, inner ysep=3pt}{},
        where level=2{text width=7em, font=\normalsize, align=left, inner xsep=3pt}{},
        where level=3{text width=15em, font=\normalsize, align=left, inner xsep=5pt}{},
        where level=4{font=\normalsize, align=left}{},
[{Tactile-based \\ Action Models}, ver
    [{Tactile Latent Learning}
        [{CLTP~\cite{ma2025cltp}, Sparsh~\cite{higuera2025sparsh}}, leaf-2d]
    ]
    [{Tactile-Action Models}
        [{Feel the Force~\cite{adeniji2025feel}, Seq2Seq Imitation~\cite{yang2023seq2seq}, RoboPack~\cite{ai2024robopack}, MimicTouch~\cite{yu2025mimictouch}}, leaf-my]
    ]
    [{Tactile-Vision-Action Models}
        [{T\mbox{-}DEX~\cite{guzey2023dexterity}, RotateIt~\cite{qi2023general}, Multimodal\mbox{-}SeeThrough~\cite{ablett2024multimodal}, VTTB~\cite{gu2024vttb}, VITaL~\cite{george2025vital},  
        RDP~\cite{xue2025reactive}}, leaf-my]
    ]
    [{Tactile-Language-Action Models}
        [{TLA~\cite{hao2025tla}, Octopi~\cite{yu2024octopi}}, leaf-my]
    ]
    [{Tactile-Vision-Language-Action Models}
        [{
        Tactile\mbox{-}VLA~\cite{huang2025tactile}, 
        VTLA~\cite{zhang2025vtla}, Touch Begins~\cite{zhao2025touch}, 
        VLA-Touch~\cite{bi2026vla}
        }, leaf-my]
    ]
]
    \end{forest}
}
\vskip -0.05in
\caption{A structured overview of tactile-based action models.}
\label{fig: tactile}
\end{figure*}

%% file: sections/6-low_level_controller/latent_learning.tex
\subsection{Latent Learning}
\label{subsec: latent_learning}

\input{figures/6_low-level_control/latent_learning}

Latent learning examines how robotic models can acquire robust and generalizable representations from input data, and how these representations can be effectively leveraged by the policy head to decode actions. We categorize existing work into two directions: pretrained latent learning, which focuses on learning high-quality representations for downstream tasks, and latent action learning, which addresses not only representation learning but also how to optimize the decoding of these representations into actions. We summarize these in Figure~\ref{fig: latent_learning}.

\subsubsection{Pretrained Latent Learning} 
Learning transferable visual representations, often referred to as robotic representations, is an important foundation for robust real-world visuomotor control. Inspired by the success of large-scale pre-training in computer vision~\cite{he2022masked} and natural language processing~\cite{devlin2019bert}, recent robotics research increasingly relies on pre-training visual encoders on diverse, domain-relevant data. Depending on the data source, these representations can be broadly grouped into three categories: models trained on general-purpose image datasets (\textit{e.g.}, ImageNet~\cite{deng2009imagenet}), human--object interaction datasets that provide rich manipulation priors (\textit{e.g.}, Ego4D~\cite{grauman2022ego4d}), and robot-centric datasets collected from embodied interaction (\textit{e.g.}, BridgeData V2~\cite{walke2023bridgedata}).

\noindent \textbf{Training on General Datasets.}
General-purpose visual pre-training offers a scalable alternative to robot-specific data collection. Parisi \emph{et al.}~\cite{parisi2022unsurprising} show that multi-layer feature fusion from vision models pre-trained on large image datasets can provide strong control-oriented representations, in some cases rivaling state-based inputs. Theia~\cite{shang2025theia} further distills multiple vision foundation models into a compact and transferable representation for downstream robot learning.

\noindent \textbf{Training on Human Egocentric Datasets.}
Human egocentric video provides rich priors on object interaction, hand motion, and task structure, making it a valuable source for transferable robotic representations. VC-1~\cite{majumdar2023we} and R3M~\cite{nair2023r3m} show that large-scale egocentric pre-training can yield general visual backbones and support few-shot real-robot learning, while Voltron~\cite{karamcheti2023language} and HRP~\cite{srirama2024hrp} further incorporate language grounding and affordance priors to improve higher-level semantics and cross-embodiment generalization. Together, these works highlight the value of egocentric pre-training for bridging human demonstrations and robot manipulation.

\noindent \textbf{Training on Robotic Datasets.}
The growing availability of large-scale robot-collected datasets has enabled representation learning directly grounded in embodied interaction. MVP~\cite{radosavovic2023real} uses masked visual pre-training on diverse in-the-wild videos to learn transferable visual representations. Premier-TACO~\cite{zheng2024premier} further improves multitask representations through temporally contrastive objectives for better few-shot policy learning. More recently, manipulation-centric analysis~\cite{jiang2025robots} explicitly aligns visual representations with downstream manipulation performance through large-scale robotic pre-training.

\subsubsection{Latent Action Learning}
Recent advances in latent action learning connect video representation and world modeling, enabling robots to acquire action abstractions beyond explicit supervision. Dual-system and latent-learning approaches for VLAs, as discussed in Section~\ref{subsec: vla}, also fall into this category. Broadly, latent actions can be grouped into two forms: discrete representations, typically learned through quantization, and continuous vector representations. We review these two classes below.

\paragraph{Discretization and Vector Quantization}
These methods convert continuous actions or representations into discrete tokens, yielding compact and reusable action primitives that simplify policy learning and improve stability. Early work such as ILPO~\cite{edwards2019imitating} and LAPO~\cite{schmidt2023learning} inferred latent action structure from observation-only demonstrations via inverse dynamics. Later methods discretized continuous control into latent tokens for efficient generative control and scalable multitask learning~\cite{lee2024behavior, wu2025discrete, li2025star}. More recent approaches combine discretized latent actions with vision--language models: LAPA~\cite{ye2025latent} learns latent actions from visual transitions, while DreamGen~\cite{jang2025dreamgen} scales this paradigm using video world models and inverse dynamics to synthesize large robotic datasets. Overall, this line trades fine-grained continuity for compositionality, robustness, and scalability.

\paragraph{Continuous Latent Action Representations}
Continuous latent action representations encode actions as vectors in a continuous space, providing a compact interface between perception and control. By abstracting low-level motor commands into smooth latent trajectories, these methods support interpolation, generalization, and transfer across behaviors, embodiments, and task contexts.

\noindent \textbf{Latent Dynamics Representations.}
A first line of work represents task-relevant dynamics as latent variables that condition action generation. MimicPlay~\cite{wang2023mimicplay} predicts latent actions from goal images derived from human play for long-horizon imitation, CLAM~\cite{liang2025clam} learns continuous latent actions from unlabeled demonstrations with robust motor alignment, and CoMo~\cite{yang2026learning} scales latent motion embeddings from internet videos to improve cross-domain generalization.

\noindent \textbf{Implicit World Modeling.}
Another class exploits latent predictions from learned world models to guide control. VPP~\cite{hu2025video} adapts a video foundation model for manipulation and aggregates predicted visual latents to condition a diffusion policy. FLARE~\cite{zheng2025flare} aligns future latent predictions with implicit world models to improve generalization from human demonstrations, while Genie Envisioner~\cite{liao2026genie} enforces consistency between video-diffusion latents and action-diffusion policies, strengthening prediction–control coupling.

\noindent \textbf{Latent Diffusion Policies.}
Several approaches integrate continuous latent spaces into diffusion-based control. LAD~\cite{bauer2025latent} trains diffusion policies in a shared latent action space for cross-embodiment transfer, while LaDi-WM~\cite{huang2025ladi} learns a latent diffusion world model whose predicted semantics and geometry condition downstream action generation.

\noindent \textbf{Koopman-based Latent Dynamics.}
Koopman-based methods learn linear latent dynamics that approximate nonlinear manipulation processes. KoDex~\cite{han2023utility} studies Koopman representations for dexterous manipulation, KOROL~\cite{chen2024korol} learns interpretable object-centric latents through Koopman rollouts for stable and explainable control, and KOAP~\cite{bi2025imitation} combines Koopman-based control with diffusion planning for long-horizon execution.

\noindent \textbf{Goal- and Instruction-conditioned Latents.}
Some methods jointly encode observations and instructions into task-centric latent actions. Procedure Cloning~\cite{yang2022chain} introduces latent procedural abstractions inspired by chain-of-thought reasoning for long-horizon tasks, while UniVLA~\cite{bu2025univla} learns task-centric latent actions from cross-embodiment data in an unsupervised manner, enabling VLA prediction without action labels.

%% file: figures/6_low-level_control/latent_learning.tex
\begin{figure*}[t]
\centering
\includegraphics[width=\linewidth]{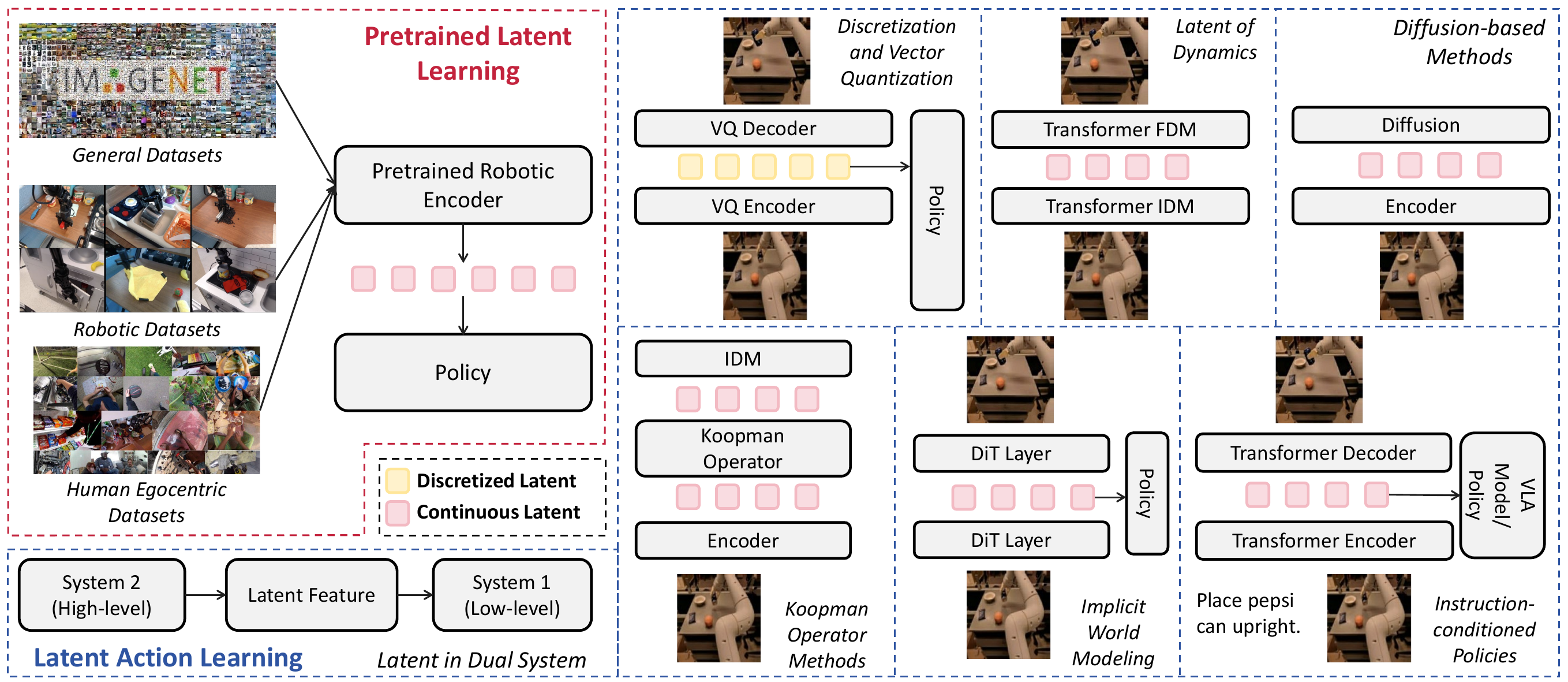}  
\vskip -0.1in
\caption{Overview of latent learning. \textbf{Top left}: Robotic encoders are pretrained on general datasets, human egocentric datasets, and robotic datasets to produce latent representations for downstream policies. \textbf{Bottom left}: In a dual-system architecture, System 2 outputs latent features to guide System 1 in action generation. \textbf{Right}: Latent action learning can be implemented using either discretized latents through \new{Vector Quantization (VQ, yellow)} or continuous latents (pink). 
\new{Representative modules include Inverse Dynamics Models (IDM), Forward Dynamics Models (FDM), and Diffusion Transformers (DiT).}
}
\label{fig: latent_learning}
\end{figure*}

%% file: sections/6-low_level_controller/policy_learning.tex
\subsection{Policy Learning}
\label{subsec: policy_learning}

Policy learning specifies how learned representations are decoded into executable robot actions, defining the functional interface between perception, internal representations, and motor execution.

\subsubsection{MLP-based Policy}
Early learning-based manipulation policies adopt multilayer perceptrons (MLPs) to directly map encoded observations to actions, serving as simple and efficient visuomotor controllers when paired with strong representations~\cite{nair2023r3m, radosavovic2023real}.

\subsubsection{Transformer-based Policy}
Transformer-based policies use self- and cross-attention to model temporal dependencies and multimodal context, enabling sequence-to-sequence action generation from variable-length histories. Within this paradigm, ACT~\cite{zhao2023learning} introduced action chunking to reduce error accumulation and stabilize bimanual manipulation, with later extensions scaling skills and incorporating richer supervision~\cite{bharadhwaj2024roboagent, haldar2024baku}. In parallel, autoregressive policies treat manipulation as next-token prediction over action sequences, supporting in-context imitation and long-horizon reasoning, as exemplified by ICRT~\cite{fu2024context} and CARP~\cite{gong2025carp}.

\subsubsection{Diffusion Policy (DP)}
DP formulates action generation as an iterative denoising process, enabling multimodal trajectory synthesis and strong generalization in robotic manipulation~\cite{chi2023diffusion}. Recent extensions improve this paradigm along several directions: 3D-aware variants enhance spatial grounding for contact-rich and long-horizon manipulation~\cite{ze20243d}; mixture-of-experts and equivariant architectures improve multitask generalization by encoding task structure and symmetry priors~\cite{wang2025sparse, yang2025equibot}; and integration with video prediction and world modeling supports planning-oriented and data-efficient learning~\cite{zhu2025unified}. These advances make diffusion policy a flexible and extensible paradigm for modern robot manipulation.

\subsubsection{Flow Matching (FM) Policy}
FM policies replace stochastic denoising with deterministic transport dynamics, yielding faster inference and smoother action trajectories. Early work demonstrated FM as a stable and efficient alternative to diffusion for imitation learning~\cite{zhang2024affordance}. Building on this foundation, RTC~\cite{black2025real} enables real-time execution by overlapping action generation with ongoing control, further improving efficiency and temporal consistency.

\input{tables/7-bottlenecks/method_comparisons}

\subsubsection{Other Policy Paradigms}
Beyond the above families, alternative designs explore different inductive biases. State-space-model-based policies leverage long-sequence modeling for efficient control~\cite{jia2024mail}, while frequency-domain policies encode actions spectrally to improve long-horizon stability~\cite{zhong2025freqpolicy}. These approaches highlight complementary directions for structuring action generation.

Across these paradigms, policy learning has evolved from direct regression toward structured sequence generation that models uncertainty, temporal dependencies, and long-horizon coherence. MLP- and Transformer-based policies emphasize representation decoding and context aggregation, while diffusion and flow-based methods treat action generation as a generative process over trajectories, improving robustness and multimodality. Together, these advances point toward policies that more tightly integrate perception, memory, and planning for scalable robotic manipulation.

%% file: tables/7-bottlenecks/method_comparisons.tex
\begin{table*}[htbp]
\centering
\scriptsize
\caption{Summary of major foundation-model-based robotic manipulation methods across high-level planning and low-level action modeling. For each category, we summarize a representative method, its core idea, modalities, data regime, advantages and use cases, and limitations, highlighting the capability boundaries and characteristic failure modes of different approaches.}
\vskip -0.05in
\label{tab: method_comparison}
\resizebox{\textwidth}{!}{
\begin{tabular}{
>{\centering\arraybackslash}p{0.3cm}
>{\raggedright\arraybackslash}p{1.8cm}
>{\raggedright\arraybackslash}p{2.0cm}
>{\raggedright\arraybackslash}p{3.6cm}
>{\raggedright\arraybackslash}p{2.0cm}
>{\raggedright\arraybackslash}p{4.0cm}
>{\raggedright\arraybackslash}p{2.4cm}
>{\raggedright\arraybackslash}p{2.2cm}}
\toprule
\textbf{Type}& \textbf{Category} & \textbf{Representative Method} & \textbf{Core Idea} & \textbf{Modalities} & \textcolor{black}{\textbf{Training Data Regime}} & \textbf{Advantages \& Use Cases} & \textbf{Limitations} \\
\midrule

\multirow{20}{*}{\rotatebox{90}{High-level Planner}}

& \makecell[tl]{LLM-based \\ Task Planning \\ (Section II-A)} & \makecell[tl]{SayCan \\ \cite{brohan2023can}} & LLM-planned global reasoning with affordance-grounded feasibility & RGB images, language & \textcolor{black}{(I) 68K robot demos; (II) 12K successful autonomous robot trajectories} & Strong semantic reasoning and task decomposition & High cost; may violate physical constraints \\
\cmidrule(lr){2-8}
& MLLM-based \newline Task Planning \newline (Section II-B)& \makecell[tl]{PaLM-E \\ \cite{driess2023palm}} & Embodied multimodal LLM for end-to-end task reasoning & RGB image, state estimation, language & \textcolor{black}{(I) Large-scale language data; (II) Internet-scale vision-language data; (III) Real robot demos} & Unified perception and reasoning; positive transfer across tasks & High cost; may lose language capabilities \\
\cmidrule(lr){2-8}
& Code Generation (Section II-C) & \makecell[tl]{Code as \\ Policies \\ \cite{liang2023code}} & LLM generates executable code; calls perception and control APIs & language, perception API, control primitives & \textcolor{black}{No training data} & Unsafe or invalid code may be generated \\
\cmidrule(lr){2-8}
& Geometric \newline Constraint-based\newline Planning\newline (Section II-D)& \makecell[tl]{VoxPoser \\ \cite{huang2023voxposer}} & Builds 3D value maps from language-vision for smooth motion & RGB images, language, 3D voxel maps & \textcolor{black}{No training data} & Strong physical grounding & Relies on predefined geometry; limited semantics \\
\cmidrule(lr){2-8}
& Affordance\newline Learning\newline (Section II-E)& \makecell[tl]{CLIPort \\ \cite{shridhar2022cliport}} & Decouples semantics and localization: CLIP grounding, Transporter pick-and-place & RGB images, language & \textcolor{black}{Simulated robot demos} & Action-centric interaction & Affordance ambiguity in cluttered or novel scenes \\
\cmidrule(lr){2-8}
& 3D\newline Representations\newline as Planner\newline (Section II-F) & \makecell[tl]{F3RM \\ \cite{shen2023distilled}} & Distills CLIP into 3D fields for language-based pick-and-place & RGB images, language, 3D geometry & \textcolor{black}{10 robot demos across 5 tasks} & Rich spatial reasoning & Sensitive to reconstruction errors and dynamics \\
\midrule

\multirow{40}{*}{\rotatebox{90}{Low-level Action Modeling}}
& 2D\newline Non-LLM-based\newline VLA\newline (Section III-B2a) & \makecell[tl]{RT-1 \\ \cite{brohan2023rt}} & Transformer maps vision and language to actions; discretization enables low-level control & RGB images, language & \textcolor{black}{130K robot demos across 744 tasks from 13 robots} & Demonstration scalability and diverse manipulation skills & Weak semantic priors; limited generalization and cross-domain transfer \\
\cmidrule(lr){2-8}
& 2D \newline LLM/VLM-based\newline VLA\newline (Section III-B2a)& \makecell[tl]{OpenVLA \\ \cite{kim2025openvla}} & Fine-tunes Prismatic VLM on robot data; fuses SigLIP and DINOv2 with discrete action tokens & RGB images, language & \textcolor{black}{970K real robot demos curated from the Open X-Embodiment dataset} & Strong language grounding and scalability & Action discretization limits precision \\
\cmidrule(lr){2-8}
& Latent Learning\newline for VLA\newline (Section III-B2a)& \makecell[tl]{LAPA \\ \cite{ye2025latent}} & Learns discrete latent actions via VQ-VAE; VLM predicts latent actions & RGB images, language & \textcolor{black}{(I) 220K action-free human manipulation videos; (II) robot demos from BridgeData, Open X-Embodiment; (III) 100 robot trajectories in SIMPLER} & Efficient and transferable latent action abstractions & Inference latency \\
\cmidrule(lr){2-8}
& Action\newline Quantization\newline for VLA\newline (Section III-B2a)& \makecell[tl]{MiniVLA \\ \cite{belkhale2024minivla}} & Learns quantized motor codebooks; trajectory discretization for stable training & RGB images, language & \textcolor{black}{(I) Real robot demos from Open X-Embodiment; (II) robot demos from LIBERO-90 and BridgeData} & Reduced complexity and stabilized training & Difficulty in semantic consistency and collapse avoidance \\
\cmidrule(lr){2-8}
& Dual-/Multi-system VLA (Section III-B2a)& \makecell[tl]{OpenHelix \\ \cite{cui2025openhelix}} & Dual-system VLA: frozen MLLM and pre-trained policy; separates fast reactive control and slow deliberative reasoning & language, 3D scene representation, RGB, proprioception & \textcolor{black}{Robot demos from CALVIN} & Efficient real-time control & Limited high-level reasoning \\
\cmidrule(lr){2-8}
& Learning with Auxiliary Tasks (Section III-B2a) & \makecell[tl]{$\pi_{0.5}$ \\ \cite{intelligence2025pi_}} & Hierarchical inference with subtask prediction; flow matching action generation & RGB images, language, proprioception & \textcolor{black}{(I) Real mobile manipulator demos in homes; (II) multi-environment robot trajectories; (III) vision-language data} & Continuous action modeling with smooth trajectories & Higher computational cost \\
\cmidrule(lr){2-8}
& Reinforcement \newline Learning\newline (Section III-B2a)& \makecell[tl]{VLA-RL \\ \cite{lu2025vla}} & Large-scale rollout RL refinement, multi-turn conversation formulation; PPO with robotic reward densification & RGB images, language, proprioception & \textcolor{black}{(I) Robot demos for 40 LIBERO tasks; (II) online simulation trajectories collected with RL; (III) segmented trajectories with automatically generated process rewards} & Scalable refinement from sparse feedback and execution-aligned action decoder & Extensive online interaction and critic dependence \\
\cmidrule(lr){2-8}
& Efficient VLA\newline (Section III-B2a)& \makecell[tl]{VLA-Adapter \\ \cite{wang2026vla}} & Parameter-efficient adapters; all-layer latents and bridge attention for efficient action generation & RGB images, language, proprioception & \textcolor{black}{Robot demos from LIBERO and CALVIN} & Parameter-efficient adaptation & Strongly depends on backbone quality \\
\cmidrule(lr){2-8}
& Robustness \newline (Section III-B2a)& \makecell[tl]{BadVLA \\ \cite{zhou2025badvla}} & Objective-decoupled backdoor attacks; reveals VLA security vulnerabilities & RGB images, language & \textcolor{black}{(I) Robot demos from LIBERO and SimplerEnv; (II) poisoned robot trajectories with visual or multimodal triggers} & Systematic backdoor vulnerability analysis & Limited to training-time injection \\
\cmidrule(lr){2-8}
& 3D VLA\newline (Section III-B2b)& \makecell[tl]{SpatialVLA \\ \cite{qu2025spatialvla}} & Introduces 3D positional encodings and adaptive action grids to capture transferable spatial priors & 3D point clouds or depth, RGB, language & \textcolor{black}{1.1M real robot demos from multiple datasets} & Improved 3D spatial reasoning & Depends on accurate scene reconstruction \\
\bottomrule
\end{tabular}
}
\end{table*}

%% file: sections/9-future_research.tex
\input{tables/3_benchmarks/benchmarks}

\section{Prospective Future Research Directions}

\new{As discussed in Sections~\ref{sec: high_level_planner} and \ref{sec: low_level_control}, and summarized in Table~\ref{tab: method_comparison}, foundation-model-based manipulation systems have advanced in semantic reasoning, task decomposition, and action generation.} However, most approaches remain constrained by narrow task and embodiment coverage, limited sensory and action interfaces, and insufficient real-world robustness. Future research should therefore address four tightly coupled challenges: building a general robot brain that generalizes across tasks and embodiments, collecting diverse large-scale data, enabling richer multimodal interaction beyond vision-dominant perception, and ensuring safety and collaboration in open-world environments.

\subsection{Core Challenge 1: Building a True Robot Brain}

\new{A central long-term goal of robotic foundation models is to move beyond task-specific fitting toward a true ``robot brain'' that can generalize across tasks, objects, environments, and embodiments. While recent large-scale VLA systems and policy learning approaches have shown promising progress, most existing results are still obtained under relatively controlled variations, such as adapting to new object instances within familiar task families, recombining known task primitives, or handling moderate environmental changes~\cite{dreczkowski2025learning, wen2025diffusion}. These settings are closer to interpolation than true extrapolation.}

\new{A true robot brain, however, should support stronger forms of generalization, including robust transfer to unseen task structures, object categories, and substantially different environments, as well as more compositional and adaptive behavior under novelty. Achieving this capability requires more than scaling current policies within narrow benchmarks. It instead calls for general-purpose architectures that support flexible modality and embodiment interfaces and scale effectively with data and compute~\cite{ji2025robobrain,intelligence2025pi_}; stronger generalization and adaptation through continual and lifelong learning, memory, replay, and efficient representation updates~\cite{liu2026pretrained}; and more reliable long-horizon execution through tighter coupling between high-level planning and low-level closed-loop control~\cite{fan2025long}. Together, these directions point toward robotic systems that acquire reusable structure for flexible open-world manipulation, rather than merely performing well within fixed distributions.}

\subsection{Core Challenge 2: Data and Evaluation Bottlenecks}

Despite recent progress, current robotic foundation models still rely heavily on data-driven fitting. 
\newer{As summarized in Table~\ref{tab:manipulation_simulator_benchmark}, existing benchmarks evaluate different aspects of robot manipulation. CALVIN emphasizes long-horizon language-conditioned execution but has limited environment diversity and a fixed embodiment; LIBERO focuses on multi-task and lifelong learning but uses a fixed embodiment and limited out-of-distribution settings; SimplerEnv evaluates real-to-sim consistency rather than real-world performance. Other benchmarks mainly target low-level control, offline imitation learning, large-scale task diversity, or language-guided reasoning, each with corresponding limitations in scene, embodiment, horizon, or interaction complexity. Benchmark scores should therefore be interpreted with respect to the capabilities and constraints of the specific evaluation setting, rather than as universal measures of robotic intelligence.}
In addition, the lack of standardized evaluation protocols hinders fair comparison across works. This challenge is compounded by the scarcity and cost of real-world manipulation data and the limited fidelity of current simulators for contact-rich, articulated, and deformable interactions. Future progress therefore requires more reproducible evaluation frameworks~\cite{yakefu2025robochallenge}, scalable ``data flywheels'' for autonomous experience collection and high-value data extraction~\cite{li2026roboclaw}, and higher-fidelity differentiable simulators for policy optimization and sim-to-real transfer~\cite{yang2025robot}.

\subsection{Core Challenge 3: Multimodal Physical Interaction}

\new{Vision-centric manipulation alone is insufficient for real-world physical intelligence. As summarized in Section~\ref{sec: input_modeling}, existing action models can incorporate additional modalities such as language, tactile signals, and audio~\cite{huang2025tactile}. However, unlike humans, who adaptively integrate multiple sensory cues during task execution, most current models support only one extra modality beyond vision and language, making unified multimodal perception and action difficult to achieve. Future robotic foundation models should therefore fuse richer sensory streams, such as touch, audio, and thermal signals, into unified, adaptive, and temporally coherent representations despite heterogeneous sampling rates and noise characteristics.}

\new{Beyond multimodal input modeling, robotic manipulation also requires stronger physical interaction capabilities. A major frontier is the manipulation of complex materials, such as cloth, cables, fluids, and granular media~\cite{zhang2025robochemist}, where state spaces are effectively high-dimensional and contact dynamics dominate. Progress in this direction will likely require new object representations, such as graph- or field-based models~\cite{ze2023gnfactor,li2026information}, together with stronger physics-informed inference and learning algorithms that remain stable under partial observability and uncertain contact.}

\subsection{Core Challenge 4: Safety and Collaboration}

\new{As robots enter human environments, safety must be treated as a first-class design constraint rather than a secondary feature. As discussed in Section~\ref{subsec: vla}a, existing research has begun to address safety-related issues such as failure prediction, backdoor attacks, and language-instruction interference~\cite{gu2025safe,zhou2025badvla,xu2025can}. However, future robotic foundation models will require stronger guarantees at multiple levels. At the single-robot level, they should provide intrinsic safety through self-constrained control that respects kinematic and dynamic limits while regulating smoothness, force, and energy in real time~\cite{song2026dual}. At the multi-robot level, they should support safe collaboration through predictive coordination and shared protocols. For real-world deployment, safety must also be paired with autonomy in fault detection and recovery, including continuous monitoring and reliable fallback policies~\cite{zhou2025code}. More broadly, practical systems will likely rely on hybrid paradigms that combine the adaptability of learning-based models with the stability of classical methods, such as rule-based logic or MPC~\cite{hansen2024td}, especially in safety-critical settings.}

%% file: tables/3_benchmarks/benchmarks.tex
\begin{table*}[t]
\centering
\caption{Summary of representative robot manipulation simulators and benchmarks.  Here, demos denote the number of demonstrations.}
\label{tab:manipulation_simulator_benchmark}
\vskip -0.05in

\scriptsize

\resizebox{\textwidth}{!}{
\begin{tabular}{
    >{\raggedright\arraybackslash}p{0.105\textwidth}
    >{\centering\arraybackslash}p{0.075\textwidth}
    >{\centering\arraybackslash}p{0.050\textwidth}
    >{\centering\arraybackslash}p{0.060\textwidth}
    >{\centering\arraybackslash}p{0.072\textwidth}
    >{\raggedright\arraybackslash}p{0.135\textwidth}
    >{\raggedright\arraybackslash}p{0.225\textwidth}
    >{\raggedright\arraybackslash}p{0.225\textwidth}
}
\toprule
\textbf{Name}
& \textbf{Simulator}
& \textbf{Objects}
& \textbf{Tasks}
& \textbf{Data Scale}
& \textbf{Robot}
& \newer{\textbf{Evaluation Scope}}
& \newer{\textbf{Main Limitation}} \\
\midrule

MetaWorld~\cite{yu2020meta}
& MuJoCo
& 80
& 50
& 2.5K demos
& Sawyer
& \newer{Multi-task reinforcement learning and policy generalization}
& \newer{State-based single-arm manipulation only} \\

RLBench~\cite{james2020rlbench}
& CoppeliaSim
& 28
& 100
& 2.7K demos
& Franka Panda
& \newer{Vision-based multi-task manipulation}
& \newer{Waypoint-based demonstrations simplify execution} \\

CALVIN~\cite{mees2022calvin}
& PyBullet
& 28
& 34
& 40M frames
& Franka Panda
& \newer{Long-horizon language-conditioned manipulation}
& \newer{Limited environment diversity} \\

Robomimic~\cite{mandlekar2022matters}
& MuJoCo
& 15
& 8
& 6K demos
& Franka Panda
& \newer{Offline imitation learning}
& \newer{Limited task diversity and data scale} \\

LIBERO~\cite{liu2023libero}
& MuJoCo
& 67
& 130
& 6.5K demos
& Franka Panda
& \newer{Lifelong and multi-task learning}
& \newer{Limited embodiment and out-of-distribution coverage} \\

RoboCasa~\cite{nasiriany2024robocasa}
& MuJoCo
& 2509
& 100
& 100K+ demos
& Franka Panda, Fourier GR1, Tiago, etc.
& \newer{Large-scale household manipulation}
& \newer{Limited scene diversity} \\

SimplerEnv~\cite{li2025evaluating}
& SAPIEN
& 16
& 10
& \newer{140K+ demos}
& Google Robot, WidowX
& \newer{Real-to-sim evaluation of VLA policies}
& \newer{Limited scene and task diversity} \\

RoboTwin~\cite{chen2025robotwin}
& SAPIEN
& 731
& 50
& 25K demos
& ALOHA, UR5, Franka, ARX-X5
& \newer{Cross-embodiment manipulation}
& \newer{Limited long-horizon evaluation} \\

ManiSkill3~\cite{tao2025maniskill3}
& SAPIEN
& 10K+
& 12
& 1M frames
& Franka Panda, Unitree H1, AnyMAL-C etc.
& \newer{Large-scale general manipulation}
& \newer{Limited language-conditioned evaluation} \\

VLABench~\cite{zhang2025vlabench}
& MuJoCo
& 2K+
& 100
& 5K demos
& Franka Panda
& \newer{Language-guided reasoning and manipulation}
& \newer{Limited contact-rich manipulation} \\
\bottomrule
\end{tabular}
}
\end{table*}

%% file: sections/10-conclusion.tex
\section{Conclusion}

This survey has reviewed recent progress in robot manipulation through a unified perspective on high-level planning, learning-based low-level action modeling, and future research directions. By organizing existing methods along planning and learning dimensions, we aim to clarify the algorithmic structure of modern robotic manipulation systems. Despite substantial progress, the field remains far from achieving robust and general-purpose capability. Key challenges include building unified architectures that couple perception, reasoning, and control, overcoming data, evaluation, and perception bottlenecks that limit scalability and generalization, and ensuring safety and reliability in open-world human environments. Addressing these challenges will be critical for moving beyond specialized manipulation pipelines toward more adaptive and trustworthy robotic foundation models. We hope this abstraction-driven survey provides a useful reference for understanding current methods and framing future research in robot manipulation.

%% file: sections/0-abstract_appendix.tex
\begin{abstract}
In this supplementary document, we provide additional discussions, clarifications, and comparisons:
\begin{itemize}
    \item How foundation models benefit robotic manipulation compared with conventional methods (see Section~\ref{sec: comparision}).
    \item A summary of major manipulation task types and the corresponding representative methods (see Section~\ref{sec: manipulation_tasks}).
    \item A clarification of generalization in foundation-model-based robotic manipulation, together with representative methods that address this challenge (see Section~\ref{sec: generalization}).
    \item A quantitative comparison across major method categories (see Section~\ref{sec: quantitative_comparisons}).
\end{itemize}
\end{abstract}

%% file: sections/11-appendix.tex
\appendix

\subsection{Comparison Between Foundation Model-Based and Conventional Methods}
\label{sec: comparision}

Compared with conventional manipulation pipelines, foundation models broaden robotic manipulation from a largely task-specific control problem to one that tightly couples semantic understanding, spatial grounding, and action generation. Traditional methods often rely on carefully engineered task representations, manually designed rewards or constraints, and fixed object or task settings, which limits their flexibility in open-ended environments. By contrast, foundation models introduce large-scale semantic and multimodal priors that enable robots to interpret free-form language instructions, reason over unseen objects and scenes, infer task-relevant affordances or subgoals, and support more transferable policy learning. Their contribution to manipulation is therefore not merely improved perception, but a new interface between high-level intent understanding and downstream motion generation.

\noindent\textbf{Problem Formulation.}
Conventional manipulation methods are typically formulated around well-specified control objectives, such as trajectory optimization, feedback control, reward maximization, or expert action imitation. In contrast, foundation model-based methods often reformulate manipulation as a joint problem of instruction understanding, semantic grounding, intermediate reasoning, and action generation. This shift is particularly important in open-world settings, where the robot must determine not only how to execute a motion, but also what to do, where to act, and in what order.

\noindent\textbf{Capability Differences.}
Foundation model-based methods differ from conventional approaches in several important aspects. First, they enable more flexible task specification through natural language, reducing the reliance on manually engineered symbolic interfaces or task-specific reward design. Second, they improve semantic generalization across unseen objects, instructions, and scenes by leveraging large-scale pretraining. Third, they provide stronger support for long-horizon and compositional manipulation, since they can produce intermediate artifacts such as task plans, affordance priors, subgoals, code, or structured scene descriptions that guide downstream execution. By contrast, conventional methods are often more specialized and may require explicit redesign when the task distribution changes.

\noindent\textbf{Relationship to Conventional Execution Modules.}
Foundation models do not completely replace conventional motion generation and control. In most robotic manipulation systems, they primarily contribute high-level reasoning, semantic abstraction, or structured intermediate guidance, while low-level trajectory generation and control are still handled by motion planners, trajectory optimization methods, feedback controllers, or learned action policies. This is because precise execution remains constrained by embodiment, dynamics, safety, latency, and closed-loop stability, all of which are areas where conventional control methods remain essential.

\noindent\textbf{Limitations and Failure Modes.}
Foundation model-based manipulation methods also introduce new challenges. Their semantic predictions may not always correspond to physically feasible actions; inferred affordances or subgoals may be plausible at the language level but geometrically invalid; and long-horizon reasoning still often suffers from error accumulation. In addition, large models often incur higher computational cost and latency, which may conflict with strict real-time manipulation. These limitations suggest that foundation models and conventional methods are best viewed not as competing paradigms, but as complementary components in a hierarchical manipulation stack.

\subsection{Manipulation Tasks}
\label{sec: manipulation_tasks}

Early grasping methods primarily focused on identifying stable grasp configurations through geometric analysis, force closure conditions, or task-specific heuristics. Candidate grasp poses were analytically generated from object geometry, contact normals, or predefined grasp templates, and then executed using inverse kinematics (IK) and motion planning. Building on such established grasps, early approaches to dexterous, deformable, mobile, quadrupedal, and humanoid manipulation typically assumed a known target pose or a task-specific goal. The emphasis was on optimizing motion trajectories or control commands to achieve these goals under physical and kinematic constraints, rather than learning end-to-end action generation from raw sensory input as in modern RL or IL methods. The comparison is summarized in Table~\ref{tab: method_comparison}.

\input{tables/4_manipulation_tasks/learning_vs_non-learning}

\subsubsection{Grasping}

In the narrow sense considered in this work, grasping specifically refers to the tasks of grasp detection and grasp generation. These tasks involve identifying feasible grasp configurations from sensor inputs such as images or point clouds, allowing robotic end-effectors to securely pick up objects. The primary focus is on predicting the position and orientation of the gripper to ensure stable and reliable grasps, even in the presence of diverse object shapes, varying poses, and cluttered environments.

\noindent \textbf{Non-Learning-based Grasp.} 
Early methods generate grasp poses by explicitly analyzing object geometry~\cite{jones1990planning}, performing contact-driven force analysis~\cite{miller2004graspit}, or applying task-specific rules~\cite{stoytchev2005behavior}, in combination with the gripper model. 
However, due to their limited generalization ability in handling complex shapes, occlusions, and unseen objects, these approaches have gradually been replaced by data-driven, learning-based methods, as discussed below. The taxonomy of learning-based grasping approaches is illustrated in Figure~\ref{fig: grasp_methods}.

\noindent \textbf{Rectangle-based Grasp.}
Grasping rectangles were first introduced by Jiang et al.~\cite{jiang2011efficient}. While they are visually similar to bounding boxes, their representational semantics are fundamentally different. A grasping rectangle is defined as a 5-dimensional representation. 
Subsequent methods have progressively incorporated deep learning techniques, ranging from basic convolutional neural networks (CNNs)~\cite{redmon2015real, kumra2017robotic, morrison2018closing} to more advanced architectures such as ResNet~\cite{he2016deep}, GR-ConvNet~\cite{kumra2020grconvnet} and Transformer~\cite{wang2022transformer}, and further to CLIP-based models that integrate textual information through feature fusion methods~\cite{nguyen2024lightweight}. More recently, diffusion models have also been employed~\cite{vuong2024language}. Over time, the models have evolved from simple to complex, and the modalities have shifted from unimodal to increasingly multimodal.

\input{figures/4_manipulation_tasks/grasp_methods}

\noindent \textbf{6-DoF Grasp.}
Two-dimensional rectangle-based representations, which typically assume parallel-jaw grippers executing top-down grasps, are limited to 2 or 3 degrees of freedom (DoF). This restricts the gripper’s orientation and reduces applicability in unstructured or complex environments. To address these limitations, researchers have proposed 6-DoF grasp representations~\cite{yan2018learning} that define a grasp as a complete 6-dimensional pose in 3D space. This formulation enables the robot to grasp objects from arbitrary orientations, significantly improving flexibility and robustness. The 6-DoF representation is especially beneficial in cluttered scenes, non-planar object poses, and scenarios involving complex geometries.

Apart from a few early 2D-based approaches~\cite{zhai2023monograspnet}, the majority of recent methods adopt 3D representations. In particular, methods based on point cloud inputs have explored a wide range of architectures, ranging from mapping convolutional networks~\cite{fang2023anygrasp} and transformer-based models~\cite{fan2025miscgrasp}, to generative autoencoders~\cite{yan2018learning}, variational autoencoders~\cite{mousavian20196dof, iwase2025zerograsp}, and UNet-style frameworks~\cite{xie2025rethinking}. 
To improve efficiency, various methods have been proposed, including applying instance segmentation or heatmap to focus on task-relevant manipulation regions~\cite{murali20206, chen2023efficient}, reducing the grasp search space by representing grasp poses with contact points on the object surface~\cite{sundermeyer2021contact}, incorporating graph neural networks for geometric reasoning on point cloud data~\cite{moghadam2023grasp}, and adopting an economic supervision paradigm that selects key annotations and leverages focal representation to reduce training cost and improve performance~\cite{wu2024economic}.
Some works further address structural distortions commonly found in real-world point cloud data by introducing completion or denoising modules to convert the input into a clean and consistent style~\cite{cheng2025pcf, cai2025real}. 
Several methods also address the SE(3)-equivariance problem~\cite{lim2025equigraspflow, hu2025orbitgrasp} by modeling grasp generation as a continuous normalizing flow over SE(3) with equivariant vector fields, or by predicting per-point grasp quality over the orientation sphere using spherical harmonic basis functions.
Meanwhile, several methods also leverage 3D representations such as SDF~\cite{breyer2021volumetric, song2025implicit, jauhri2024learning}, NeRF~\cite{rashid2023language, dai2023graspnerf}, and 3DGS~\cite{zheng2024gaussiangrasper, ji2025graspsplats, yu2024sparsegrasp} to extract geometric features or to sample grasp points for downstream prediction.

\noindent \textbf{Language-driven Grasp.}
In recent years, language-driven grasping has gained increasing attention, aiming to achieve object-specific and instruction-guided manipulation. Existing approaches can be broadly categorized into three groups. The first adopts multimodal feature fusion~\cite{xu2023joint, nguyen2024language}, where textual and visual modalities are jointly encoded, often via cross-attention mechanisms. The second leverages existing grasp models to generate large numbers of grasp candidates, followed by ranking or scoring with LLMs or VLMs to select the most confident grasps. For instance, LLMs can generate task-specific descriptions~\cite{tang2023graspgpt}, while VLMs are used for visual grounding to compute grasp confidence scores~\cite{qian2025thinkgrasp, tang2025affordgrasp}. Representative methods include VL-Grasp, which employs VLMs to attend to the target object and generate grasps~\cite{lu2023vl}; OWG, which leverages VLM-based semantic priors for planning under occlusion~\cite{tziafas2025towards}; and Reasoning-Grasping, which integrates visual-linguistic inference for improved object-level understanding~\cite{jin2025reasoning}. Other efforts adapt MLLMs for environment-aware error correction~\cite{luo2025roboreflect}, or learn object-centric attributes to enable rapid grasp adaptation across tasks~\cite{yang2024attribute}. Affordance-driven approaches also ground grasp generation in language and vision cues~\cite{dong2025rtagrasp, song2025learning, deshpande2025graspmolmo}. The third line of work directly fine-tunes MLLMs on large-scale grasping datasets to build grasp foundation models~\cite{xu2024rt, zhang2025vcot}.

\noindent \textbf{Challenges.}
First, the aforementioned grasp detection and generation methods were originally designed for 2-DoF parallel-jaw grippers. 
However, with the increasing use of dexterous hands, research has shifted toward dexterous grasping, which requires more complex annotations such as hand joint angles and contact force maps. To handle the high dimensionality of this task, various generative approaches—including diffusion-based models—have been proposed~\cite{xu2024dexterous, weng2024dexdiffuser, wei2024grasp}. Some approaches represent grasp configurations via contact points or maps~\cite{li2023gendexgrasp}, or leverage interaction-based representations like D(R, O)~\cite{wei2024d} to infer grasps from geometric relationships.
Second, traditional grasping strategies typically rely on a single end-effector (e.g., suction or parallel grippers), which limits adaptability to diverse object geometries. To address this, several works have explored bimanual grasping using dual-arm or dual-gripper configurations~\cite{mahler2019learning, yamada2025combo}.
Finally, grasping transparent objects remains a significant challenge, as depth sensors often fail to detect or localize such materials accurately. Recent efforts address this limitation by incorporating alternative modalities, such as LiDAR~\cite{deng2025fusegrasp} or 3D reconstruction~\cite{duisterhof2024residual, shi2024asgrasp, kim2025t}, to infer the geometry of transparent or reflective objects.

\input{tables/4_manipulation_tasks/manipulation_tasks}

\subsubsection{Dexterous Manipulation}

Dexterous manipulation refers to the capability of robotic systems equipped with multi-fingered or anthropomorphic hands to achieve precise and coordinated object control through complex contact interactions. It involves in-hand reorientation, fine force modulation, and multi-point contact, enabling actions such as twisting, grasping, and rotating, as illustrated in Figure~\ref{fig: case_dex}. This capability is critical for tasks requiring high precision and adaptability, such as tool use, assembly, and manipulation of small or irregular objects. Human hand models typically include 20–25 DoF, with each finger modeled by 4 DoF, the thumb by 4–5 DoF, and additional DoF from the palm and wrist for enhanced realism~\cite{welte2025interactive}.

\input{figures/4_manipulation_tasks/case_dexterous}

\noindent \textbf{Non-Learning-based Methods.}
Early work on dexterous manipulation predominantly employed non-learning approaches, including optimization- and control-based techniques such as heuristic search and constrained optimization~\cite{bai2014dexterous}. These methods typically assume access to known dynamics and object models, and focus on trajectory planning under physical constraints.

\noindent \textbf{Learning-based Methods.} 
With the advent of RL, both model-based and model-free paradigms have been applied to dexterous manipulation. Model-based methods such as PDDM~\cite{nagabandi2020deep} exploit learned dynamics for efficient planning and control, while model-free approaches directly optimize policies from interaction~\cite{rajeswaran2018learning}, sometimes enhanced with few-shot imitation to accelerate learning on real hands~\cite{zhu2019dexterous}. More recently, human-in-the-loop frameworks such as HIL-SERL~\cite{luo2025precise} combine demonstrations, online corrections, and reinforcement learning to achieve sample-efficient training of precise and dexterous skills directly on real robots.
IL has gained attention for its data efficiency, exemplified by DexMV~\cite{qin2022dexmv}, which employs kinematic retargeting from human videos to robot hands. Extensions incorporate auxiliary tasks such as contact prediction~\cite{liang2025dexhanddiff, fu2025cordvip} and inverse dynamics modeling~\cite{radosavovic2021state}, enabling better generalization and action consistency.
Hybrid IL–RL approaches aim to combine these strengths, typically using IL for policy initialization followed by RL refinement~\cite{kumar2016learning, arunachalam2023dexterous}. Integration can vary: DexMV applies IL first and fine-tunes with RL, whereas ViViDex~\cite{chen2025vividex} performs RL in a privileged state space before distilling policies via IL.

Recent advances in VLA models further expand generalization. DexGraspVLA~\cite{zhong2025dexgraspvla} integrates semantic reasoning with diffusion-based policies for grasping in cluttered scenes, while OFA~\cite{li2025object}, LBM~\cite{barreiros2025careful}, and Being-H0~\cite{luo2025being} leverage multimodal prompts and human videos to generate dexterous motions and follow language instructions.

Human guidance provides another supervision source: Chen et al.~\cite{chen2025object} use object and wrist trajectories from videos to guide RL, and Mandi et al.~\cite{mandi2025dexmachina} introduce functional retargeting to transition from demonstrations to autonomous control. Affordance reasoning further supports grasp-specific strategies~\cite{kannan2023deft, agarwal2023dexterous}, guiding object selection under semantic constraints.
Finally, recent works emphasize robustness and autonomy. Task decomposition reduces complexity by structuring subtasks~\cite{dasari2023learning}, while recovery mechanisms such as REBOOT~\cite{hu2023reboot} introduce IL-trained reset policies to handle failures in long-horizon settings.

\noindent \textbf{Challenges.} 
Beyond high-dimensional control and contact dynamics, dexterous manipulation faces broader challenges.
First, many real-world tasks are long-horizon and compositional, requiring sequential execution of fine-grained skills. Chen et al.~\cite{chen2023sequential} address this by decomposing tasks into discrete skills trained with RL and linking them via a high-level policy.
Second, sim-to-real transfer remains a bottleneck due to discrepancies in perception, dynamics, and actuation. CyberDemo~\cite{wang2024cyberdemo} mitigates this through data augmentation, improving robustness under domain shift.
Third, accurate perception is difficult in cluttered or partially observed scenes, where occlusion hampers object tracking. DexPoint~\cite{qin2023dexpoint} leverages point cloud completion to recover missing geometry and enhance spatial awareness.

\subsubsection{Soft Robotic Manipulation}

Soft manipulators, built with compliant materials or structures, are well-suited for human–robot collaboration, operation in uncertain environments, and safe, adaptive grasping, as illustrated in Figure~\ref{fig: case_soft}. They overcome the limitations of rigid manipulators when handling delicate or deformable objects~\cite{chen2022review}. To this end, a wide range of designs have been developed to support diverse applications~\cite{ansari2017towards, thuruthel2018stable, puhlmann2022rbo, liu2022touchless, wang2025spirobs}.

\input{figures/4_manipulation_tasks/case_soft}

\noindent \textbf{Non-Learning-based Methods.}
For soft manipulator control, analytical kinematic models based on the constant-curvature assumption map shape parameters to end-effector poses via virtual rigid-link chains and Denavit–Hartenberg transformations~\cite{jones2006kinematics}. Similarly, three-dimensional continuum manipulators can be approximated as rigid-jointed robots, with computed-torque control achieving closed-loop dynamics~\cite{wang2020dynamic}. Other approaches employ model predictive control with Koopman operators~\cite{bruder2019modeling, lv2025multi}, or rely on accurate Lagrangian models combined with adaptive dynamic sliding mode control to enhance robustness~\cite{kazemipour2022adaptive}. Hybrid schemes also integrate learning into non-learning frameworks: forward dynamics can be approximated with machine learning and combined with trajectory optimization for open-loop control~\cite{thuruthel2017learning}, while feedback-driven strategies exploit learned models to stabilize and restore system states~\cite{thuruthel2018stable}.

\textbf{Learning-based Methods.} 
RL and IL have been widely applied to soft manipulator control. 
Model-free RL avoids explicit physical modeling by training policies in simulation for closed-loop endpoint control~\cite{you2017model}. Forward dynamics learned with recurrent networks can be integrated into trajectory optimization to generate samples and train predictive controllers~\cite{thuruthel2018model}, while LSTM-based dynamics models enable feedback policy learning~\cite{centurelli2022closed}. Domain randomization combined with incremental offline training has improved task-space accuracy and adaptability~\cite{li2022towards}, and wavelet-based dynamics approximations paired with LSTM and TD3 controllers further enhance generalization under variability~\cite{zhou2024cable}.
IL approaches include Soft DAgger~\cite{nazeer2023soft}, which employs dynamic behavior mapping for online expert-like action generation, and KineSoft~\cite{yoo2025kinesoft}, which integrates kinesthetic teaching with diffusion-based policy learning. Hybrid methods also emerge, such as SS-ILKC~\cite{meng2025sensor}, which combines multi-objective RL with adversarial IL to learn goal-directed control strategies in sensor space, supported by sim-to-real pre-calibration for zero-shot transfer.

\noindent \textbf{Challenges.} 
Soft-hand-based manipulation faces several challenges, including modeling highly nonlinear and underactuated dynamics, achieving precise force and pose control with deformable or fragile objects, ensuring robustness under environmental uncertainty and sensor noise, and lowering the cost and complexity of data collection and teleoperation for policy training. To improve sample efficiency, Soft DAgger~\cite{nazeer2023soft} enables online imitation learning from limited demonstrations through dynamic behavior mapping. To reduce hardware complexity in teleoperation, Liu et al.~\cite{liu2022touchless} propose a flexible bimodal sensory interface that combines vision-based perception with wearable sensors, enabling intuitive and low-cost control without bulky equipment.

\subsubsection{Deformable Object Manipulation}

\input{figures/4_manipulation_tasks/case_deformable}

Deformable object manipulation (DOM) requires robots to perceive and control non-rigid objects whose shapes vary under applied forces. Unlike rigid-body manipulation, it demands reasoning over high-dimensional, continuous state spaces, coping with uncertain deformation dynamics, and interpreting subtle visual or tactile cues. As illustrated in Figure~\ref{fig: case_deformable}, tasks such as cloth folding, rope and cable tying, and food handling involve diverse deformations—including tension, compression, and bending—making DOM a complex yet crucial challenge in robotic manipulation~\cite{sanchez2018robotic, blanco2024t, gu2023survey}.

\noindent \textbf{Non-Learning-based Methods.}
Traditional approaches to DOM, such as path planning~\cite{saha2007manipulation} and model-based control~\cite{luque2024model}, often rely on simplified physical models or analytical solutions. However, these methods typically struggle with generalization and real-time adaptability in complex environments, leading to a shift toward data-driven techniques such as RL, IL, and hybrid learning-control frameworks.

\noindent \textbf{Learning-based Methods.}
Matas et al.~\cite{matas2018sim} propose a deep RL method based on a modified DDPG algorithm that learns from visual and robot state inputs and achieves zero-shot sim-to-real transfer via domain randomization. In contrast, IL methods directly exploit demonstrations: DexDeform~\cite{li2023dexdeform} extracts latent skills from human demonstrations and fine-tunes them with limited robot data, DefGoalNet~\cite{thach2024defgoalnet} predicts goal configurations from few-shot demonstrations conditioned on state and context, and MPD~\cite{scheikl2024movement} generates movement primitives with diffusion models. To combine both paradigms, DMfD~\cite{salhotra2022learning} incorporates expert data into RL through an advantage-weighted BC loss, expert-initialized replay buffers, and reset-to-state initialization, achieving stable and efficient policy optimization.

Beyond learning paradigms, DOM has explored diverse strategies spanning geometric modeling, affordance prediction, and structure-aware perception. DeformGS~\cite{duisterhof2024deformgs} represents deformable objects with canonical Gaussians and learns a time-conditioned deformation field for temporally consistent 3D pose tracking. Affordance-based approaches include DeformerNet~\cite{hu20193}, which predicts end-effector displacements from partial point clouds, Foresightful Affordance~\cite{wu2023learning}, which integrates dense per-pixel affordances with long-horizon value estimation, and APS-Net~\cite{luan2025learning}, which ranks standardized folding and flattening trajectories guided by affordance heatmaps. Language-conditioned affordance prediction has also been explored~\cite{deng2024learning}.
Finally, estimating object-specific physical properties has emerged as a complementary direction~\cite{caporali2024deformable, kuroki2024gendom}. GenDOM~\cite{kuroki2024gendom} learns a parameter-conditioned policy and leverages a single human demonstration with differentiable simulation to infer unseen object properties, enabling generalization to novel deformable instances.

\noindent \textbf{Challenges.}
Deformable objects lack a fixed pose, and key deformation regions are often occluded during manipulation. To improve visibility and perception, recent work jointly optimizes camera and manipulator motion, guided by structure-of-interest cues~\cite{weng2024interactive}. Deformation dynamics also exhibit delayed responses, complicating real-time control; this is addressed by encoding states into latent spaces and predicting their dynamics. For example, DeformNet~\cite{li2024deformnet} encodes object geometry with PointNet and a conditional NeRF, and models temporal evolution with a recurrent state-space model for latent-level MPC. An even greater challenge is modeling topological changes. DoughNet~\cite{bauer2024doughnet} extends latent-space modeling to jointly capture geometric and topological variations from point clouds, enabling long-horizon planning with CEM for tool selection and manipulation.

\subsubsection{Mobile Manipulation}

\input{figures/4_manipulation_tasks/case_mobile}

Mobile manipulation is a robotic paradigm that combines navigation and manipulation capabilities within a single system. It allows robots to physically interact with objects beyond a fixed workspace by actively navigating the environment. This integration poses significant challenges in perception, planning, and control, as it requires coordinating whole-body motion, handling long-horizon tasks, and dealing with dynamic, partially observable scenes. 
Several studies have focused on designing robots specifically for mobile manipulation~\cite{holmberg2000development, hebert2015mobile, fu2025mobile}.
As illustrated in Figure~\ref{fig: case_mobile}, mobile manipulation is critical for real-world applications such as household assistance, warehouse automation, and service robotics~\cite{thakar2023survey}.

\noindent \textbf{Non-Learning-based Methods.}
Traditionally, mobile manipulation is decomposed into two separate problems: navigation and manipulation. Navigation is typically handled by classical path planners, such as grid-based methods (e.g., Dijkstra) and sampling-based planners (e.g., RRT). Manipulation is often addressed using grasp models, motion primitives, or trajectory planning based on point clouds.
Some control methods perform joint optimization of navigation and manipulation. For example, Berenson et al.~\cite{berenson2008optimization} optimize full-body configurations and grasp poses simultaneously, then plan motions via sampling. Chitta et al.~\cite{chitta2012mobile} coordinate base-arm planning with ROS-based navigation and point cloud grasping. Others embed manipulation goals directly into MPC cost functions~\cite{pankert2020perceptive}.

\noindent \textbf{Learning-based Methods.}
These methods have emerged to learn policies for whole-body control. 
For RL, Wang et al.~\cite{wang2020learning} use RGB-D inputs to estimate object pose and train a policy that jointly predicts base velocity, arm trajectories, and gripper actions. HarmonicMM~\cite{yang2024harmonic} extends this by integrating visual and pose information into a unified RL framework, while Wu et al.~\cite{wu2020spatial} propose spatial Q-value learning at the map level for navigation guidance. To improve reward signals, Honerkamp et al.~\cite{honerkamp2021learning} design dense rewards based on end-effector reachability, and Causal MoMa~\cite{hu2023causal} introduces causality-aware modeling of control–reward relations to stabilize policy optimization.
For IL, MOMA-Force~\cite{yang2023moma} learns motion and force policies from visual-force demonstrations, while HoMeR~\cite{sundaresan2025homer} decomposes tasks into global keypose prediction and local refinement to map end-effector targets to whole-body actions. Wang et al.~\cite{zhicheng2025object} leverage SAM2-based perception for object-centric IL, and Skill Transformer~\cite{huang2023skill} treats manipulation as a skill prediction problem, learning both skill categories and corresponding low-level actions.
Hybrid methods combine IL with RL to enhance generalization. For example, Xiong et al.~\cite{xiong2024adaptive} initialize policies via behavior cloning and refine them through online RL to handle unseen articulated objects.

Beyond learning paradigms, recent work integrates high-level reasoning and multimodal representations. VLA-based methods such as MoManipVLA~\cite{wu2025momanipvla} process multimodal instructions to predict end-effector waypoints while delegating base motion to a trajectory optimizer. LLM-driven frameworks, including MoMa-LLM~\cite{honerkamp2024language} and SayPlan~\cite{rana2023sayplan}, combine open-vocabulary language, scene graphs, and reasoning for object search and high-level task planning. Complementary efforts in \textit{3D scene modeling} guide manipulation through active perception: ActPerMoMa~\cite{jauhri2024active} optimizes viewpoint selection and grasp reachability via incremental TSDF mapping, while TaMMa~\cite{hou2025tamma} employs sparse Gaussian localization and depth completion to generate accurate target poses.

\noindent \textbf{Challenges.}
In addition to perception–action coordination, dynamic obstacles, and long-horizon decision-making, mobile manipulation faces several additional challenges.
The first challenge is enabling effective human–robot collaboration. Ciocarlie et al.~\cite{ciocarlie2012mobile} developed an assistive system that combines user interfaces, shared control, and autonomy to transform a PR2 robot into an in-home assistant. Extending this line of work, Robi Butler~\cite{xiao2025robi} introduces a closed-loop household control framework that supports multimodal interaction, where high-level planners such as LLMs and VLMs enable natural language and gesture-based commands for intuitive and remote collaboration.
The second challenge is real-time manipulation during navigation, where robots must coordinate mobility and manipulation under environmental constraints. Whole-body motion control frameworks~\cite{haviland2022holistic, burgess2023architecture} address this problem by enabling reactive planning and execution for on-the-move manipulation.

\subsubsection{Quadrupedal Manipulation}

\input{figures/4_manipulation_tasks/case_quadrupedal}

Quadrupedal manipulation is an emerging paradigm that combines the agile mobility of quadruped robots with the ability to physically interact with objects. Unlike traditional manipulators or mobile bases, quadrupeds can traverse complex and unstructured terrains while maintaining dynamic stability, making them particularly suitable for applications such as search and rescue, field robotics, and autonomous exploration. \newer{Representative tasks are illustrated in Figure~\ref{fig: case_quad}}.
Manipulation can be realized through several embodiments: whole-body loco-manipulation~\cite{jeon2023learning, huang2024hilma, ouyang2024long, ding2024quar, song2024germ, zhao2025more}; leg-as-manipulator designs that repurpose one or more legs for interaction~\cite{cheng2023legs, arm2024pedipulate, he2024learning}; back-mounted arms~\cite{wolfslag2020optimisation, fu2023deep, arcari2023bayesian, ferrolho2023roloma, yokoyama2023asc, qiu2024learning, mendonca2025continuously, zhang2024gamma, liu2025visual, jiang2024learning, wang2025quadwbg, pan2025roboduet, zhi2025learning, hou2025efficient}; and grippers integrated into front legs for simultaneous locomotion and manipulation~\cite{lin2024locoman}. Each embodiment introduces unique challenges in whole-body coordination, dynamic control, and perception-driven planning.

\noindent \textbf{Non-Learning-based Methods.}
Recent advances in quadrupedal manipulation have explored diverse control strategies, ranging from model-based optimization to learning-based approaches. Optimization-based methods provide interpretable and physically grounded control with strong task generalization. For example, Arcari et al.~\cite{arcari2023bayesian} combine MPC with Bayesian multi-task error learning for real-time dynamics adaptation. Wolfslag et al.~\cite{wolfslag2020optimisation} incorporate SUF stability metrics and contact constraints into a quadratic programming framework for robust, support-leg-aware planning, a concept further extended by RoLoMa~\cite{ferrolho2023roloma} to improve trajectory robustness. LocoMan~\cite{lin2024locoman} demonstrates a hardware–control co-design approach, equipping quadrupeds with lightweight front-leg manipulators and employing unified WBC to achieve agile locomotion and precise manipulation.

\noindent \textbf{Learning-based Methods.}
In the realm of RL, recent research has focused on developing structured and informative control representations. Jeon et al.~\cite{jeon2023learning} introduce a hierarchical RL framework that encodes interaction experience, robot morphology, and action history into latent representations to facilitate effective policy learning. Fu et al.~\cite{fu2023deep} decouple leg and arm rewards in the policy gradient formulation to enhance coordination between locomotion and manipulation, while Zhi et al.~\cite{zhi2025learning} present a unified force–position controller that estimates forces from perception instead of sensors. Hou et al.~\cite{hou2025efficient} incorporate explicit arm kinematics and feasibility-based rewards to promote physically plausible behaviors. Two-stage training schemes have also been explored: RoboDuet~\cite{pan2025roboduet} sequentially trains locomotion and arm policies with reward adaptation for coordination. GAMMA~\cite{zhang2024gamma} improves grasp precision by conditioning policies on grasp poses, while Wang et al.~\cite{wang2025quadwbg} propose GORM, a metric for grasp reachability under varying base poses, guiding base movements for optimal grasping. Teacher–student frameworks have further advanced base–arm coordination, as in VBC~\cite{liu2025visual} and Jiang et al.~\cite{jiang2024learning}, where visual or pose-tracking guidance is distilled into student policies. Other works combine hierarchical and hybrid designs, such as HiLMa-Res~\cite{huang2024hilma}, which uses high-level RL to control Bézier parameters and base motion while relying on hybrid CPG–Bézier controllers for leg movements, and Pedipulate~\cite{arm2024pedipulate}, which demonstrates end-to-end RL for single-foot manipulation.
In the IL domain, Human2LocoMan~\cite{niu2025human2locoman} enables cross-embodiment transfer from XR-driven human demonstrations using a modular Transformer policy, effectively transferring human manipulation skills to quadrupeds.
Hybrid IL–RL frameworks further improve efficiency and generalization. He et al.~\cite{he2024learning} employ BC to train a high-level planner for grasp trajectories, while coordinates locomotion and single-leg manipulation. WildLMa~\cite{qiu2024wildlma} builds an IL-based skill library from VR demonstrations, sequencing and composing tasks under LLM guidance.

VLA-based frameworks have also emerged for high-level semantic reasoning. QUAR-VLA~\cite{ding2024quar} pioneers the application of VLA models to quadrupeds, while GeRM~\cite{song2024germ} and MoRE~\cite{zhao2025more} integrate mixture-of-experts architectures with offline RL to learn generalist visuomotor policies and Q-functions with enhanced generalization and decision quality. QUART-Online~\cite{tong2024quart} advances this line by introducing action-chunk discretization and semantic alignment training, enabling latency-free inference for quadrupedal VLA tasks.

Finally, advances in 3D semantic perception and high-level planning extend quadrupedal capabilities. GeFF~\cite{qiu2024learning} performs real-time NeRF-based reconstruction with semantic relevance fields to guide locomotion, while manipulation is executed via learned grasp models. At the task level, LLMs have been combined with policy libraries: Ouyang et al.~\cite{ouyang2024long} propose a hierarchical framework where LLMs parse long-horizon, multi-skill instructions into structured subgoals executed by RL-based skills, bridging symbolic reasoning and continuous control.

\noindent \textbf{Challenges.}
In addition to common challenges such as balance–manipulation coupling, terrain adaptability, and perception delays, real-world deployment remains difficult due to the sim-to-real gap caused by modeling inaccuracies and sensor noise. To mitigate this, fully autonomous RL pipelines have been developed. Mendonca et al.~\cite{mendonca2025continuously} propose a framework that integrates on-policy data collection with continuous training, while ASC~\cite{yokoyama2023asc} addresses long-horizon tasks by decomposing them into modular skills trained via RL, coordinating them through a skill-switching policy jointly trained with IL and RL, and introducing a corrective policy to improve robustness during deployment.
Long-horizon task execution itself poses another significant challenge. Cheng et al.~\cite{cheng2023legs} propose a stage-wise RL framework that independently learns locomotion and single-leg manipulation skills, which are later composed via a behavior tree to accomplish temporally extended tasks.

\subsubsection{Humanoid Manipulation}

Humanoid manipulation involves robotic platforms with human-like morphology, typically including a torso, two arms, and either simplified 2-DoF or fully dexterous hands, with the lower body implemented as either a mobile base or bipedal legs. These systems are designed to perform object interaction tasks in human environments.
These systems seek to replicate or extend human manipulation capabilities, enabling actions such as grasping, lifting, tool use, and coordinated bimanual operations, as illustrated in Figure~\ref{fig: case_human}. While the humanoid form offers natural compatibility with tools and environments built for humans, it also introduces challenges in balance control, whole-body coordination, and fine motor skills, making humanoid manipulation a central research focus in robotics and embodied intelligence.

\input{figures/4_manipulation_tasks/case_humanoid}

\noindent \textbf{Non-Learning-based Methods.}
Early humanoid control primarily relied on traditional approaches such as rule-based and analytical controllers~\cite{harada2006dynamics, bouyarmane2012humanoid}. Some works also explored the integration of learning-based models into control frameworks. For example, OKAMI~\cite{li2025okami} leverages 3D human pose estimation from a single human video to infer joint positions and derive executable actions for humanoid robots via inverse kinematics.

\noindent \textbf{Learning-based Methods.}
The recent trend in humanoid manipulation has shifted from hand-tuned controllers toward fully learning-based approaches. As highlighted in~\cite{gu2025humanoid}, the field remains underexplored, with existing methods falling broadly into RL, IL, and VLA frameworks.

For RL-based approaches, FLAM~\cite{zhang2025flam} leverages a pre-trained human motion foundation model to score the stability of humanoid poses, using this score as an auxiliary reward to encourage balanced behaviors. Xie et al.~\cite{xie2023hierarchical} combine diffusion-based motion generation with reinforcement learning to achieve whole-body humanoid manipulation.
In the domain of IL, TRILL~\cite{seo2023deep} directly maps RGB images to pose commands, while OmniH2O~\cite{he2025omnih2o} abstracts unified motion goals from diverse modalities (e.g., language, RGB, MoCap, VR) using a diffusion policy. iDP3~\cite{ze2025generalizable} adapts the classic DP3 diffusion policy from third-person to egocentric views, enabling teleoperation of the head, torso, and arms. Qiu et al.~\cite{qiu2025humanoid} leverage large-scale egocentric demonstrations to align human and robot actions in a shared representation space, and TACT~\cite{murooka2025tact} incorporates tactile feedback to improve manipulation robustness.
Hybrid RL+IL strategies have also been proposed. For example, André et al.~\cite{schakkal2025hierarchical} decouple locomotion and manipulation by training a mid-level trajectory generator with IL to specify motion goals, while using RL to optimize a low-level tracking policy for accurate execution.

Finally, VLA methods aim to develop end-to-end language-conditioned policies for joint locomotion and manipulation. \newer{HumanVLA~\cite{xu2024humanvla}} encodes images and language separately, concatenates features, and decodes actions through an MLP. GR00T N1~\cite{bjorck2025gr00t} further integrates visual and tokenized language features via a VLM before feeding them into a diffusion-based head. \newer{Humanoid-VLA~\cite{ding2025humanoid}} aligns language and action spaces using cross-attention to fuse visual features, while TrajBooster~\cite{liu2025trajbooster} introduces a loco-manipulation VLA trained with retargeted data from real-world to simulation.

\noindent \textbf{Challenges.}
In addition to challenges such as multimodal perception, balance–manipulation coupling, and high degrees of freedom, humanoid manipulation also struggles with enabling multi-agent collaboration. For instance, CooHOI~\cite{gao2024coohoi} exploits object dynamics as an implicit communication channel to achieve coordinated manipulation across multiple agents. To further address the sim-to-real gap, Lin et al.~\cite{lin2025sim} introduce a generalizable reward formulation and a decoupled RL architecture, which improve sample efficiency through staged training.

\subsection{Generalization}
\label{subsec:generalization_tasks}

Robotic manipulation generalization can be broadly categorized into three dimensions: \textbf{environment generalization}, \textbf{task generalization}, and \textbf{cross-embodiment generalization}. Environment generalization concerns robustness to variations in conditions such as Sim2Real transfer, spatial transformations, lighting, and background or distractor changes. Task generalization emphasizes maintaining performance across different task configurations, including long-horizon tasks, few-shot or meta-learning, continual learning, and skill composition. Cross-embodiment generalization focuses on transferring skills across robots with diverse morphologies, kinematics, dynamics, or sensing modalities, which is crucial for building general-purpose embodied agents. In what follows, we structure our survey along these three perspectives.

\input{figures/7_bottlenecks/generalization}

\subsubsection{Environment Generalization}

Environment generalization in robotic manipulation refers to the ability of policies trained under specific conditions to maintain performance across varying environments, scenes, viewpoints, and physical settings. Such generalization is challenged by discrepancies between simulation and reality, spatial transformations of objects and cameras, and visual appearance changes such as lighting, backgrounds, and distractors. Existing approaches address these challenges from several complementary perspectives, including sim-to-real and real-to-sim-to-real transfer, geometric equivariance and viewpoint robustness, and robustness to other visual environmental variations.

\textbf{i) Sim2Real and Real2Sim2Real Generalization}

Simulation offers a scalable alternative for training and evaluating manipulation policies, alleviating the cost, operational risk, and safety concerns associated with large-scale real-world data collection. However, discrepancies in dynamics, sensing, visual appearance, and environmental complexity often result in poor transferability, commonly referred to as the sim-to-real gap~\cite{lyu2024scissorbot}. Existing approaches can be broadly organized into three paradigms according to how simulated and real-world experience are combined: simulation-only training, sim--real adaptation and co-training, and real2sim2real reconstruction.

\noindent \textbf{Simulation-Only Training.}
This paradigm assumes that sufficient variability introduced during simulation training can enable policies to transfer directly to the real world without further adaptation. Domain randomization is a representative strategy, where physical parameters such as mass, friction, and damping~\cite{chen2021understanding}, as well as perceptual factors such as textures~\cite{tobin2017domain}, lighting~\cite{tremblay2018training}, and object poses~\cite{ren2019domain}, are systematically perturbed to prevent policies from overfitting to a specific simulator configuration. Peng et al.~\cite{peng2018sim} demonstrate that randomized dynamics can substantially improve real-world transfer, while more recent systems such as DexScale~\cite{liu2025dexscale} scale simulation diversity and augmentation to increasingly complex manipulation settings. Nevertheless, zero-shot transfer remains difficult when simulated variability fails to capture fine-grained visual, contact, or dynamic properties of real environments.

\noindent \textbf{Sim--Real Adaptation and Co-Training.}
Rather than relying exclusively on simulation, another line of work introduces a limited amount of real-world experience to anchor policy learning. Earlier approaches primarily adapt simulation-trained policies through real-world fine-tuning or continual learning; for example, Josifovski et al.~\cite{josifovski2025safe} study safe continual adaptation under deployment constraints. Natural Language Can Help Bridge the Sim2Real Gap~\cite{yu2024natural} instead uses language descriptions as a shared semantic signal across simulated and real observations, encouraging domain-invariant representations while training with abundant simulated and sparse real demonstrations. More recently, Sim-and-Real Co-Training~\cite{maddukuri2025sim} systematically demonstrates that directly training a shared policy on mixed simulated and real datasets can substantially improve real-world manipulation performance even when noticeable visual and behavioral discrepancies remain between domains. Generalizable Domain Adaptation~\cite{cheng2025generalizable} further aligns the joint distributions of observations and actions across simulation and reality through optimal-transport-based objectives, learning task-relevant domain-invariant representations and enabling simulated experience to improve real-world performance even in scenarios not represented by the limited real demonstrations. These developments shift sim-to-real learning from post-hoc policy correction toward joint exploitation of large-scale simulation and sparse real-world supervision.

\noindent \textbf{Real2Sim2Real Reconstruction.}
A complementary paradigm closes the loop by reconstructing simulatable or renderable environments from real-world observations and subsequently using these reconstructions for policy learning~\cite{torne2024reconciling,yu2025real2render2real}. RialTo~\cite{torne2024reconciling} constructs task-specific digital twins from limited real-world sensing and introduces inverse distillation to transfer real demonstrations into simulation, where reinforcement learning can safely improve policy robustness before deployment back in the real world. Real2Render2Real~\cite{yu2025real2render2real} reduces dependence on both dynamics simulation and physical robot collection by reconstructing object geometry and appearance from smartphone scans and tracking object motion from a single human demonstration, enabling large-scale rendering of robot-compatible training trajectories. Real2Edit2Real~\cite{zhao2026real2edit2real} further introduces a metric 3D control interface that edits reconstructed scenes and manipulation trajectories and uses depth-conditioned video generation to synthesize geometrically consistent demonstrations under new spatial configurations. These approaches illustrate an emerging transition from manually constructed simulation environments toward real-world-grounded digital reconstruction and editable synthetic environments for scalable policy generalization.

\textbf{ii) SE(3) and SIM(3)-Equivariance Generalization}

A complementary strategy for environment generalization is to explicitly encode geometric symmetry into policy architectures. SE(3)- and SIM(3)-equivariant models constrain representations or actions to transform predictably under changes in translation, rotation, and, for SIM(3), scale, thereby reducing the need to independently observe every spatial configuration during training. EquiBot~\cite{yang2024equibot} introduces a SIM(3)-equivariant diffusion policy that improves data efficiency and generalization across changes in object pose and scale. ET-SEED~\cite{tie2025seed} develops an efficient trajectory-level SE(3)-equivariant diffusion policy and relaxes the requirement that every transition in the diffusion process must independently employ expensive equivariant computation. More recently, E3Flow~\cite{zhang2026efficient} combines spherical-harmonic-based equivariant representations with rectified flow and multimodal visual features, extending geometric equivariance toward more efficient flow-based visuomotor policy learning. Together, these methods demonstrate how explicit geometric structure can provide strong inductive biases for spatial generalization.

\noindent \textbf{Viewpoint Robustness Beyond Explicit Equivariance.}
Related approaches address viewpoint changes without requiring the policy to satisfy explicit SE(3) or SIM(3) equivariance. Adapt3R~\cite{wilcox2025adapt3r} combines pretrained semantic visual features with end-effector-relative 3D localization, enabling imitation policies to transfer across unseen camera viewpoints and robot embodiments. Another direction actively changes the observation process itself rather than passively requiring the policy to tolerate arbitrary viewpoints. Vision in Action~\cite{xiong2025vision} learns task-dependent searching, tracking, and focusing behaviors from human demonstrations using an actuated robotic neck, allowing the robot to recover informative viewpoints when objects become occluded. SaPaVe~\cite{liu2026sapave} extends active perception to VLA-based manipulation by decoupling camera motion from manipulation actions and coordinating semantic camera control with geometry-aware execution. These methods complement explicit equivariance by either constructing viewpoint-robust spatial representations or actively controlling perception to maintain informative observations during manipulation.

\textbf{iii) Others}

Lighting, background, and distractor variations constitute another major source of environment shift, as visuomotor policies can overfit to visual cues that correlate with demonstrations but are irrelevant to task execution. Existing approaches address these variations through augmentation, invariant representation learning, and training across heterogeneous visual contexts. Physically based augmentation methods explicitly vary illumination during training to improve robustness under lighting changes~\cite{jin2025physically}. Decomposing the Generalization Gap~\cite{xie2024decomposing} systematically analyzes how individual visual factors contribute to imitation-learning generalization failures, highlighting the importance of separating task-relevant information from nuisance variations. Maniwhere~\cite{yuan2024learning} combines multi-view representation learning with curriculum-based randomization and augmentation to improve robustness across combinations of visual disturbances and support sim-to-real transfer. Other approaches reduce dependence on incidental scene context through position-invariant regularization~\cite{yin2023spatial} or by learning from demonstrations collected under changing contexts~\cite{yang2020cross,qian2023robot}. At a larger scale, Diffusion-VLA~\cite{wen2025diffusion} demonstrates that large multimodal foundation policies can retain manipulation performance under unseen objects, distractors, and new backgrounds, suggesting that broad pretraining can complement explicit augmentation and invariance mechanisms. Collectively, these methods aim to prevent policies from relying on spurious visual correlations and preserve task-relevant behavior under substantial changes in environmental appearance.

\subsubsection{Task Generalization}

Task generalization in robotic manipulation refers to the ability of learned policies to adapt to new, complex, or unseen tasks without extensive retraining. It encompasses the generalization of learned skills, structures, and adaptation mechanisms across variations in task composition, semantics, and temporal scope. Robust task generalization requires policies not only to execute individual skills but also to compose, transfer, and refine them under new configurations and objectives.

\textbf{i) Generalization for Long-horizon Tasks}

Long-horizon robotic manipulation represents a core challenge for embodied intelligence. Unlike short-horizon tasks that typically involve single-skill execution, long-horizon problems demand the coordination of multiple sub-skills, hierarchical reasoning, and temporally abstract decision-making. Robust generalization requires agents not only to plan and execute extended action sequences but also to adapt across variations in task structure, object arrangement, and dynamic environments. Recent advances approach this challenge along three interconnected axes: skill compositionality, semantic task decomposition, and structure-aware representation learning.

\noindent \textbf{Skill Compositionality.}
At the foundation lies the ability to compose and reuse modular skills~\cite{chenu2022divide,lee2021adversarial,zhu2022bottom,belkhale2023hydra,agia2022stap,zhang2023bootstrap,liu2024learning}. STAP~\cite{agia2022stap} addresses sequencing by optimizing the feasibility of action chains. Generative frameworks such as BOSS~\cite{zhang2023bootstrap} and BLADE~\cite{liu2024learning} leverage large language models to autonomously expand skill libraries and incorporate semantic grounding into action spaces, thereby enabling flexible skill reuse and extension.

\noindent \textbf{Semantic Task Decomposition.}
Beyond static skill composition, multimodal semantics have been introduced to parse and decompose tasks into modular components~\cite{mu2025look, myers2024policy, dalal2025local, lin2024hierarchical, sundaresan2025s, mao2024dexskills}. PALO~\cite{myers2024policy} employs vision-language models to translate high-level task descriptions into reusable sub-tasks, enabling rapid adaptation with minimal supervision. ManipGen~\cite{dalal2025local} integrates local policies that encode invariances in pose, skill order, and scene layout, combining them with foundational models in vision and motion planning to generalize across unseen long-horizon sequences.

\noindent \textbf{Structure-aware Representation Learning.}
A finer level of generalization is achieved through task-level abstraction and representation learning~\cite{triantafyllidis2023hybrid, tanwani2021sequential, wu2020squirl, chen2025backbone, chen2025robohorizon, yang2025bootstrapping}. TBBF~\cite{chen2025backbone} decomposes complex tasks into primitive therbligs, facilitating efficient action-object mappings and trajectory synthesis. RoboHorizon~\cite{chen2025robohorizon} formalizes a Recognize-Sense-Plan-Act pipeline by fusing LLMs with multi-view world models, addressing sparse reward supervision and perceptual complexity. HD-Space~\cite{yang2025bootstrapping} identifies atomic sub-task boundaries from demonstrations, improving sample efficiency and policy robustness with limited data.

\textbf{ii) Compositional Generalization}

Compositional generalization in robotic manipulation emphasizes the ability to solve novel tasks by systematically recombining known skills, objects, and instructions. Recent research explores diverse strategies to achieve this capability. One line of work focuses on data-driven efficiency, where targeted data collection strategies are designed to maximize coverage of compositional variations with minimal demonstrations, enabling scalable policy training~\cite{gao2024efficient}. Another approach grounds policies in programmatic or symbolic structures, allowing robots to leverage modular task representations for systematic recomposition and generalization to unseen task combinations~\cite{wang2023programmatically}. Complementary efforts investigate policy architectures that explicitly factorize control into entities or modules, facilitating the recombination of learned components to improve adaptability in complex environments~\cite{zhou2022policy}. Together, these methods illustrate a growing effort to move beyond rote memorization of tasks toward systematic generalization, enabling robots to flexibly adapt to combinatorial task spaces.

\textbf{iii) Few-shot Learning}

Few-shot learning enables robots to acquire skills from only a handful of demonstrations~\cite{chen2025vidbot, valassakis2022demonstrate, jain2024cobt, heppert2024ditto, li2025elastic}, alleviating the high cost of large-scale data collection. Research in this area can be broadly grouped into two directions: embedding structured priors and leveraging semantic transfer. 
One line of work embeds structured inductive biases into perception and control to maximize the utility of limited demonstrations. Domain-invariant constraints, including spatial equivariance, 3D geometry, or action continuity, are incorporated to reduce data requirements and improve generalization. For example, Ren et al.~\cite{ren2024learning} combine point cloud representations with a diffusion-based policy, achieving robust generalization from as few as ten demonstrations.
A complementary line of research emphasizes semantic transfer and compositional generalization. These methods extract transferable knowledge from alternative sources, such as human demonstrations or previously solved tasks, and adapt it to new objectives. For instance, YOTO~\cite{zhou2025you} maps human hand motions from video to dual-arm robotic skills, while TOPIC~\cite{song2025few} constructs task prompts and a dynamic relation graph to systematically reuse prior experience for new policy learning.

\textbf{iv) Meta Learning}

While closely related to few-shot learning, meta learning distinguishes itself by focusing on the ability to \emph{amortize} task adaptation. Instead of relying directly on structural priors or semantic transfer to generalize from a handful of demonstrations, meta learning trains over a distribution of tasks so that the model acquires a generic adaptation procedure. This paradigm equips robots with rapid adaptability to unseen tasks, even under extremely limited supervision~\cite{duan2017one,james2018task,di2023one,zhang2024one,bing2023meta,haldar2023teach,biza2023one}.

\noindent \textbf{Task-Conditioned Meta-Representations.}
One direction focuses on learning task-conditioned embeddings that serve as meta-representations. Early approaches such as Duan et al.~\cite{duan2017one} and TecNets~\cite{james2018task} encode demonstrations into low-dimensional vectors for policy conditioning. Later works extend this idea with relational graph networks~\cite{di2023one} or invariance-based objectives~\cite{zhang2024one}, enforcing structural consistency across tasks. These embedding-based approaches highlight the importance of compact task representations in accelerating policy adaptation.

\noindent \textbf{Cross-Modal and Instruction-Driven Adaptation.}
Another direction emphasizes the integration of additional modalities and task instructions into the adaptation process. MILLION~\cite{bing2023meta} incorporates natural language into the meta-learning loop, enabling semantically guided adaptation. FISH~\cite{haldar2023teach} demonstrates versatile skill acquisition from just one minute of demonstrations, while interaction-warping~\cite{biza2023one} aligns novel trajectories with past interaction structures to facilitate transfer. Collectively, these works establish meta learning as a principled framework for scalable and data-efficient generalization.

\textbf{v) Lifelong, Continual and Incremental Learning}

Endowing robots with the ability to continuously acquire new capabilities is a key step toward adaptive and autonomous embodied agents. Unlike conventional task generalization, which typically evaluates transfer to a fixed set of unseen tasks, lifelong and incremental learning consider a non-stationary setting in which new tasks or skills arrive sequentially. The central challenge is therefore twofold: newly acquired knowledge should benefit future learning, while previously learned capabilities should be retained without repeated full retraining. In robotic manipulation, existing approaches can be broadly distinguished into lifelong or continual learning, which emphasizes knowledge retention and transfer across a sequence of tasks, and skill-incremental learning, which focuses more explicitly on expanding and updating an evolving repertoire of manipulation skills.

\noindent \textbf{Lifelong and Continual Learning.}
Lifelong and continual learning aims to retain and reuse knowledge as robots encounter a sequence of new manipulation tasks~\cite{yao2025think,roy2025m2distill,gao2021cril,lei2026dynamic}. A major direction is to organize manipulation knowledge into reusable modular components. LOTUS~\cite{wan2024lotus} continually discovers recurring skills from incoming task demonstrations, incrementally expands or updates a skill library, and composes these skills through a hierarchical meta-controller, enabling forward and backward transfer across sequential tasks. PPL~\cite{yao2025think} instead represents shared motion primitives as reusable and extensible prompts; pretrained primitive prompts are preserved while new prompts are introduced for incoming skills, allowing previous knowledge to facilitate subsequent learning without overwriting existing capabilities. Complementary approaches focus more directly on preserving learned representations. M2Distill~\cite{roy2025m2distill} employs multimodal knowledge distillation to maintain consistent representations across sequential tasks and mitigate catastrophic forgetting, while CRIL~\cite{gao2021cril} uses generative replay to preserve experience from previously learned behaviors. Together, these approaches show that continual robot learning increasingly combines modular skill reuse with explicit mechanisms for knowledge preservation, balancing forward transfer to new tasks against retention of previously acquired capabilities.

\noindent \textbf{Incremental Learning.}
In contrast, skill-incremental learning focuses more explicitly on settings in which robots progressively acquire or refine atomic manipulation skills while preserving their existing skill repertoire. iManip~\cite{zheng2025imanip} formalizes this setting with a skill-incremental benchmark built on RLBench~\cite{james2020rlbench} and introduces temporal replay together with an Extendable PerceiverIO, whose expandable action prompts accommodate new action primitives while mitigating catastrophic forgetting of previously learned skills. SIL-C~\cite{lee2025policy} further reveals that incremental skill acquisition introduces an additional challenge beyond forgetting: as the underlying skill library evolves, previously learned high-level policies may become incompatible with newly added or updated skills. To address this issue, SIL-C introduces a lazy-learning interface that dynamically aligns policy-level subtasks with the evolving skill space according to trajectory-distribution similarity, allowing new skills to be incorporated while maintaining compatibility with existing downstream policies. These approaches highlight two complementary aspects of skill-incremental manipulation: extending the policy representation to accommodate new primitives and maintaining a stable interface between an evolving skill library and previously learned task policies.

\subsubsection{Cross-Embodiment Generalization}

Cross-embodiment generalization refers to the ability to transfer manipulation knowledge across agents or robotic platforms with different visual appearances, kinematics, action spaces, morphologies, or sensing configurations. Rather than categorizing existing approaches according to specific algorithmic mechanisms, we organize this direction according to the relationship between the source and target embodiments. Specifically, existing studies can be broadly divided into human-to-robot generalization, cross-robot transfer between functionally similar embodiments, and heterogeneous cross-embodiment generalization across substantially different robot morphologies and action spaces.

\textbf{i) Human-to-Robot Generalization}

Human demonstrations provide a scalable source of manipulation experience, but transferring them to robots requires bridging substantial differences in appearance, sensing, morphology, and action spaces. Early approaches therefore seek embodiment-invariant representations or transferable intermediate signals that preserve task semantics while suppressing embodiment-specific motion details. XSkill~\cite{xu2023xskill} discovers shared skill prototypes from unlabeled human and robot manipulation videos and maps them to robot actions through a skill-conditioned diffusion policy. Human2Sim2Robot~\cite{lum2025crossing} instead extracts object trajectories and hand configurations from a single human RGB-D demonstration and converts them into embodiment-agnostic rewards and initialization priors for reinforcement learning in simulation. X-Sim~\cite{dan2025x} similarly avoids direct human-to-robot action mapping by reconstructing photorealistic simulation from human videos and using object motion as a transferable learning signal before distilling the resulting behavior into a real-world policy.

More recent methods explicitly align human and robot representations during policy learning. UniSkill~\cite{kim2025uniskill} learns embodiment-agnostic skill representations from large-scale unlabeled cross-embodiment videos, allowing skills extracted from human video prompts to guide robot policies trained with robot actions. ImMimic~\cite{liu2025immimic} maps retargeted human trajectories to robot trajectories and constructs intermediate domains through trajectory interpolation, reducing visual, morphological, and physical discrepancies during co-training. EgoBridge~\cite{punamiya2025egobridge} further formulates human-to-robot transfer as domain adaptation and aligns joint latent feature--action distributions using optimal transport guided by trajectory similarity. Together, these approaches illustrate a progression from indirect motion and object-centric transfer toward explicit human--robot representation alignment.

\textbf{ii) Similar-Embodiment Cross-Robot Generalization}

A second setting considers transfer between robot platforms that retain similar functional structures, such as serial manipulators equipped with comparable end effectors. Although these robots may differ in appearance, degrees of freedom, kinematics, or control dynamics, substantial policy knowledge can often be reused without learning entirely new behavioral abstractions. Mirage~\cite{chen2024mirage} enables zero-shot transfer between different robot arms and grippers by separately correcting perceptual and control discrepancies through cross-painting, state alignment, and dynamics compensation. Wang et al.~\cite{wang2024cross} instead project heterogeneous state and action spaces into a shared latent control space, enabling transfer across manipulators with mismatched kinematics. Modularity through Attention~\cite{zhou2023modularity} takes a complementary approach by decomposing language-conditioned manipulation policies into reusable functional components and robot-specific modules, so that only embodiment-dependent components need to be adapted. These studies show that when source and target robots remain functionally similar, transfer can often be achieved through selective alignment of perception, control, or modular policy components.

\textbf{iii) Heterogeneous Cross-Embodiment Generalization}

The most challenging setting considers substantially heterogeneous embodiments whose observation spaces, action dimensions, kinematic structures, or functional capabilities may differ significantly. One direction addresses this problem through large-scale heterogeneous co-training and shared policy representations. Pushing the Limits of Cross-Embodiment Learning~\cite{yang2024pushing} jointly learns from manipulation and navigation data spanning robotic arms, mobile bases, quadrupeds, and aerial robots, demonstrating positive transfer across markedly different embodiments. CrossFormer~\cite{doshi2024scaling} further represents heterogeneous observations and actions as variable-length token sequences processed by a shared Transformer, supporting manipulation, navigation, locomotion, and aviation within a unified policy. HPT~\cite{wang2024scaling} adopts a modular alternative by combining embodiment-specific input stems and output heads with a shared Transformer trunk, allowing heterogeneous visual and proprioceptive data to contribute to common policy pretraining.

A complementary direction constructs explicit structural, functional, or system-level bridges between different morphologies. AnyBimanual~\cite{lu2025anybimanual} transfers pretrained unimanual skills to bimanual manipulation by composing reusable skill representations and aligning arm-specific visual observations. LEGATO~\cite{seo2025legato} introduces a shared grasping tool that provides a common interaction interface across embodiments, allowing manipulation behaviors to be retargeted through embodiment-specific inverse kinematics. CEI~\cite{wu2026cei} explicitly models functional similarity between different robot arms and end effectors and aligns trajectories in 3D space, including transfer between parallel-jaw grippers and dexterous hands. At a higher abstraction level, RoboOS~\cite{tan2025roboos} organizes heterogeneous robot capabilities through standardized interfaces for perception, control, communication, and skill execution, enabling reusable high-level skills to operate across different robotic platforms. Together, these methods show that heterogeneous cross-embodiment transfer can be supported through shared policies, functional correspondence, common physical interfaces, and system-level abstraction.

\subsection{Quantitative Comparison and Empirical Findings}
\label{sec: quantitative_comparisons}

{\color{red}
\subsubsection{Cross-Benchmark Patterns in Reported Results}

Table~\ref{tab:vla_quantitative_comparison} summarizes the reported performance of representative low-level action models on CALVIN~\cite{mees2022calvin}, LIBERO~\cite{liu2023libero}, and SimplerEnv~\cite{li2025evaluating}. High-level planning methods are not included in this quantitative aggregation because they typically generate heterogeneous intermediate outputs, such as subgoals, skill calls, programs, keypoints, or geometric constraints, which must be grounded and executed by downstream policies, controllers, or skill libraries. Their end-to-end success therefore reflects the joint effectiveness of planning and execution rather than the quality of the planner alone, making cross-paper task-success rates difficult to compare without a shared executor or standardized planning-specific metrics. This exclusion applies only to the quantitative table.

The reported results provide a useful overview of the performance levels in the literature, but they should not be interpreted as the outcome of a unified reproduction conducted under a fully matched training and evaluation protocol. Although the compared methods generally use the same nominal benchmark tasks and underlying demonstration sources, their effective pipelines may differ in data conversion, image preprocessing, action representation, normalization, checkpoint selection, and evaluation configuration. We therefore use Table~\ref{tab:vla_quantitative_comparison} to identify broad benchmark-level patterns, while relying on controlled empirical studies to analyze the effects of individual VLA design choices.

\paragraph{Different benchmarks measure different capabilities, and model rankings are not stable across them} CALVIN, LIBERO, and SimplerEnv emphasize different aspects of robotic manipulation. CALVIN evaluates language-conditioned sequential execution, with its principal metric measuring how many consecutive tasks a policy can complete within a long-horizon rollout. LIBERO evaluates task learning and execution across four suites that emphasize spatial relations, object variations, goal specifications, and long-horizon composition, respectively. SimplerEnv instead constructs simulation counterparts of real-robot environments and evaluates policies originally intended for real-world deployment, thereby providing a complementary real-to-sim perspective.

Consequently, performance on one benchmark does not reliably predict performance on another. For example, OpenVLA-OFT reports a higher average success rate than $\pi_0$ on LIBERO, whereas $\pi_0$ achieves higher reported scores in the Google Robot and WidowX settings of SimplerEnv. Similar ranking changes can be observed for other models and embodiments. These results indicate that current VLA performance is strongly benchmark- and embodiment-dependent. No individual benchmark should therefore be treated as a complete measure of general-purpose manipulation capability.

\input{tables/11-appendix/vla_evaluation}

\paragraph{Several LIBERO suites are approaching saturation, whereas long-horizon execution remains more discriminative} Recent methods commonly report success rates close to or above $98\%$ on the LIBERO-Spatial, LIBERO-Object, and LIBERO-Goal suites. The differences among advanced models on these suites have consequently become small, reducing their ability to distinguish improvements in modern VLA architectures. In contrast, LIBERO-Long exhibits larger performance gaps and remains more sensitive to temporal consistency, compositional execution, and error accumulation.

This pattern does not imply that spatial, object-level, or goal-conditioned generalization has been fully solved. The reported scores are obtained within a fixed benchmark ecosystem using benchmark-specific demonstrations and evaluation distributions. Rather, the results suggest that short-horizon in-suite task execution is becoming insufficient as the sole evaluation criterion and that harder long-horizon, out-of-distribution, and cross-embodiment settings are increasingly necessary for meaningful model comparison.

\paragraph{Parameter count is not a sufficient explanation of current VLA performance} The reported results do not exhibit a simple monotonic relationship between model size and task success. For example, the 0.9B-parameter X-VLA reports higher scores than several 3B- and 7B-parameter models on LIBERO and some SimplerEnv settings. This observation suggests that parameter count alone does not determine downstream manipulation performance.

However, it should not be interpreted as evidence that smaller models are intrinsically superior. Model size is confounded with the choice of vision-language backbone, robot pretraining data, policy architecture, action representation, optimization objective, and fine-tuning recipe. Moreover, many current benchmarks place substantial emphasis on adapting to a fixed set of benchmark tasks. Their scores may therefore reflect not only general-purpose representation capacity but also the efficiency with which a model specializes to the benchmark distribution.

\subsubsection{Controlled Evidence on VLA Design Choices}

Cross-paper differences in Table~\ref{tab:vla_quantitative_comparison} cannot be causally attributed to any individual design choice because the compared models differ simultaneously in their backbones, training data, action spaces, optimization objectives, and evaluation implementations. To obtain reliable insights, we synthesize controlled comparisons in which one or a small number of factors are varied within a shared experimental setup. The tables below report results using the metrics adopted by the corresponding studies; values from different experimental blocks should not be directly compared as absolute performance. Instead, the comparisons are used to identify recurring trends, conditional findings, and unresolved questions in VLA design.

\input{figures/11_appendix/ablation_arch}

\paragraph{Decoupling action generation and providing sufficient policy capacity are critical to VLA performance} A pretrained VLM does not automatically provide an effective interface for robot control, because representations optimized for visual-language understanding must still be transformed into temporally and dynamically grounded actions. As shown in Fig.~\ref{fig:abla_arch}(d), decoupling action generation from the VLM through a separate policy head increases the LIBERO success rate from $19.8\%$ to $30.2\%$, indicating that semantic representation learning and continuous control benefit from specialized computational pathways. Expanding the action head into a larger dedicated policy module further raises performance to $64.4\%$, demonstrating that action-module capacity constitutes a major bottleneck rather than a secondary implementation detail. Once a sufficiently capable policy module is established, the precise form of VLM--policy interaction has a comparatively smaller effect. As shown in Fig.~\ref{fig:abla_arch}(e), replacing loose coupling with tight layer-wise interaction leaves LIBERO performance unchanged at $90.0\%$, while producing a modest improvement from $53.7\%$ to $55.4\%$ on LIBERO-plus. This suggests that tighter latent interaction can improve robustness under more challenging visual and environmental variations, but it cannot substitute for adequate action-module capacity. Because the two ablations are conducted at different stages of model development, their absolute improvements should not be interpreted as directly matched effect sizes. Taken together, the results indicate that architectural separation and sufficient policy capacity are the primary prerequisites for effective vision-language-to-action transfer, whereas tighter VLM--policy coupling provides a smaller, setting-dependent refinement~\cite{wu2026vlanext}.

\input{tables/11-appendix/ablation_action}

\paragraph{Continuous action formulations generally outperform naive scalar discretization, whereas the relative merits of continuous objectives remain task-dependent} Controlled comparisons indicate that independently quantizing each action dimension into discrete bins can discard fine-grained control information and is particularly vulnerable to error accumulation in long-horizon execution. As summarized in Table~\ref{tab:abla_action}, replacing per-dimension discrete prediction with continuous actions increases the average number of consecutively completed CALVIN tasks from $1.85$ to $2.37$. A broader comparison on LIBERO and LIBERO-plus yields a consistent pattern: direct regression, DDIM, and flow matching all outperform the examined classification-based formulations, while VQ-VAE tokenization performs even worse than uniform binning in this setting. The results nevertheless do not establish a universal ordering among continuous objectives. Direct regression obtains the highest LIBERO score and is effectively tied with DDIM on LIBERO-plus, whereas flow matching improves the CALVIN sequence length only marginally over MSE+BCE, from $3.57$ to $3.68$. The most robust conclusion is therefore that retaining action precision through continuous prediction is more consistently beneficial than adopting any particular regression or generative objective. At the same time, these comparisons concern naive scalar binning and a specific VQ-VAE formulation and should not be interpreted as evidence that discrete action modeling is intrinsically inferior; structured sequence-level tokenizers may preserve temporal dependencies and action precision more effectively than independently quantized action dimensions~\cite{li2026matters,wu2026vlanext}.

\input{figures/11_appendix/ablation_action_horizon}

\paragraph{Action chunking is important, but the action horizon should be neither too short nor unnecessarily long} Jointly predicting multiple future actions is consistently more effective than single-step prediction, indicating that a horizon of one is insufficient to capture the short-term temporal dependencies required for coordinated manipulation. As shown in Figure~\ref{fig:abla_action_chunk}, increasing the horizon from 1 to 4 improves the success rate from $64.4\%$ to $75.4\%$ on LIBERO and from $34.0\%$ to $40.0\%$ on LIBERO-plus. Extending the horizon to 8 retains a clear advantage over single-step prediction, achieving $74.6\%$ and $43.4\%$, respectively, but does not provide a uniform improvement over horizon 4: it performs slightly worse on standard LIBERO while achieving the best result on LIBERO-plus. These results suggest that action chunks must be sufficiently long to represent locally coherent motion, but that further increasing the horizon yields setting-dependent rather than monotonic gains. An overly short horizon under-models temporal structure, whereas an unnecessarily long horizon may reduce the frequency at which the policy can incorporate new observations and correct execution errors, although this mechanism is not directly isolated by the reported ablation. The action horizon should therefore balance temporal coherence with closed-loop responsiveness and be selected according to the task dynamics and evaluation conditions~\cite{wu2026vlanext}.

\paragraph{VLM pretraining is essential, but backbone scale alone is insufficient and visual adaptation is critical} Initializing a VLA from a pretrained VLM consistently yields substantially stronger downstream performance than training the same adaptation architecture from random initialization, confirming that pretrained visual-language representations provide an important foundation for manipulation learning. However, as shown in Table~\ref{tab:abla_vlm}, the resulting control performance does not increase monotonically with backbone size. Within the Qwen3-VL family, the 2B model achieves the strongest CALVIN result of $4.142$, whereas the 8B model performs best on SimplerEnv with a success rate of $58.3\%$; increasing the backbone to 30B-A3B does not improve either benchmark and reduces the SimplerEnv score to $44.8\%$. Module-level ablations further reveal that this limitation cannot be explained solely by insufficient language-model capacity. Freezing the vision encoder substantially degrades performance, reducing Qwen2.5-VL-3B from $3.856$ to $2.855$ on CALVIN and from $48.00\%$ to $23.95\%$ on SimplerEnv, while freezing its word embeddings produces only minor changes. The same pattern is more pronounced for PaliGemma, whose performance falls from $3.506$ to $0.495$ on CALVIN and from $55.25\%$ to $13.25\%$ on SimplerEnv when the vision encoder is frozen. These results indicate that VLM pretraining is necessary but not sufficient: successful VLM-to-VLA transfer depends less on parameter count alone than on adapting pretrained visual representations to the geometric and control-relevant information required by manipulation~\cite{zhang2026vlm4vla}.

\input{tables/11-appendix/ablation_vlm}

\paragraph{Robot-data scaling depends on physical alignment and mixture composition rather than data volume alone} Heterogeneous robot pretraining can provide valuable transferable representations, but simply enlarging the pretraining mixture does not guarantee monotonic downstream improvement. As shown in Table~\ref{tab:abla_data_scaling}, under the aligned EEF-relative action representation and frozen-VLM setting, pretraining on OXE increases the LIBERO 5-shot average success rate from $66.9\%$ to $77.3\%$. However, cumulatively adding real-robot EEF data reduces performance to $73.8\%$, and further incorporating simulated EEF trajectories decreases it to $72.1\%$. These results indicate that greater embodiment diversity may introduce interference when differences in data distributions, sensing configurations, and skill frequencies are not sufficiently reconciled. Nevertheless, cross-embodiment pretraining remains useful as an initialization for low-data adaptation, improving CALVIN few-shot performance by $17.2$ percentage points in single-task success. Taken together, the evidence suggests that heterogeneous robot data is most effective when its action semantics are physically aligned and its mixture is matched to the target embodiment and downstream skill distribution, rather than when datasets are pooled solely to maximize nominal scale~\cite{wang2026rethinking,li2026matters}.

\input{tables/11-appendix/ablation_train}

\paragraph{Unified evaluation largely supports the empirical credibility of published VLA results} The unified harness reduces repository-specific variation through a common model--benchmark communication interface, isolated environments with pinned dependencies, versioned evaluation configurations and simulator assets, fixed random seeds and episode sets, and structured logging of the harness version, benchmark configuration, and per-episode metrics. Under these controlled conditions, the reproduced results remain close to the published values for the models and checkpoints summarized in Table~\ref{tab:abla_protocol}. On LIBERO, the absolute discrepancies are generally limited to $0.2$--$2.1$ percentage points, while the reproduced CALVIN results differ by only $0.04$--$0.13$ in average consecutive task length. These small gaps indicate that the published results on LIBERO and CALVIN are broadly reproducible and are unlikely to be driven primarily by independently maintained evaluation scripts. SimplerEnv exhibits larger residual discrepancies of up to $8.0$ percentage points. We hypothesize that this larger variation is mainly associated with the greater sensitivity of real-to-sim evaluation to simulator realization and with noisier benchmark conditions, rather than with the evaluated policy alone. Overall, the audit provides positive evidence that published VLA scores are broadly credible at the benchmark level, while also indicating that strict comparisons of small cross-paper differences still require matched evaluation implementations, particularly for real-to-sim benchmarks such as SimplerEnv~\cite{choi2026vla}.

\input{tables/11-appendix/ablation_protocol}

\subsubsection{Supported Conclusions and Remaining Limitations}

\paragraph{Supported conclusions} The combined cross-benchmark and controlled evidence supports several comparatively robust conclusions. First, current benchmark scores are capability-specific rather than universally predictive: model rankings vary across CALVIN, LIBERO, and SimplerEnv, while the near-saturation of several standard LIBERO suites increasingly limits their ability to discriminate among advanced policies. Second, policy-side design is a major determinant of VLA performance. Across the examined settings, separating action generation from the VLM, allocating sufficient capacity to the policy module, preserving fine-grained continuous action information, and predicting multi-step action chunks consistently improve upon direct action-token prediction, naive scalar discretization, and single-step control. Third, VLM pretraining provides a strong foundation for downstream policy learning, but neither backbone size nor generic vision-language capability alone reliably predicts manipulation performance; effective transfer additionally requires adapting visual representations to control-relevant geometry and state changes. Fourth, robot-data scaling is governed by action-space alignment, mixture composition, and downstream relevance rather than nominal data volume alone. Finally, unified reproduction recovers most published LIBERO and CALVIN results with relatively small discrepancies, providing positive evidence that the reported benchmark-level performance is broadly credible within the audited models and settings~\cite{li2026matters,wu2026vlanext,wang2026rethinking,zhang2026vlm4vla,choi2026vla}.

\paragraph{Setting-dependent findings} Several design choices exhibit conditional rather than universal advantages. Continuous action formulations generally outperform the particular scalar-binning and VQ-VAE formulations examined here, but the available evidence does not establish a consistent ordering among direct regression, diffusion, and flow matching. Likewise, action chunking is clearly preferable to single-step prediction in the reported ablation, yet the best horizon varies across standard and perturbed evaluation settings. Increasing VLM--policy interaction beyond a sufficiently expressive policy module yields only modest gains, and expanding a heterogeneous robot-data mixture can either improve transfer or introduce interference depending on its physical and statistical alignment with the target setting. Backbone scale, action horizon, generative objective, coupling strategy, and data-mixture size should therefore be treated as jointly dependent design variables rather than independently scalable quantities.

\paragraph{Remaining limitations} These conclusions should be interpreted as empirical design guidance rather than universal laws. The controlled evidence is drawn from a small number of studies and remains concentrated in the CALVIN, LIBERO, and SimplerEnv ecosystems. The reported ablations are not based on a common backbone, training budget, or factorial experimental design, and interactions among policy capacity, action representation, temporal horizon, pretraining data, and optimization objective are therefore not fully isolated. Moreover, many studies report point estimates without matched confidence intervals or repeated-seed statistics, limiting the interpretation of small numerical differences. The current synthesis also focuses on low-level VLA policies and does not establish corresponding conclusions for high-level planning, dexterous manipulation, whole-body control, safety-critical interaction, or large-scale real-world deployment. Future empirical studies should combine shared evaluation infrastructure with multi-seed controlled ablations, harder out-of-distribution and long-horizon tasks, and evaluations spanning both simulation and real robots.

}

%% file: tables/4_manipulation_tasks/learning_vs_non-learning.tex
\begin{table*}[t]
\centering
\caption{Comparison between non-learning and learning-based methods for grasping and manipulation. Here, grasping specifically refers to grasp detection and generation.}
\label{tab: method_comparison}
\vskip -0.05in
\resizebox{\textwidth}{!}{
\begin{tabular}{l
>{\centering\arraybackslash}p{3cm}
>{\centering\arraybackslash}p{4.5cm}
>{\centering\arraybackslash}p{4.5cm}
>{\centering\arraybackslash}p{2.5cm}
>{\centering\arraybackslash}p{2.5cm}
}
\toprule
\textbf{Mani. Type} & \textbf{Control Type} & \textbf{Pose Generation} & \textbf{Trajectory Generation} & \textbf{Generalization} & \textbf{Interpretability \& Stability} \\
\midrule
\multirow{2}{*}{Grasping} 
  & Non-learning & Analytical: geometric rules, force-closure analysis & IK + motion planning & Low  & High \\
  & Learning      & Learned from data                                & IK + motion planning & High & Low  \\
\midrule
\multirow{2}{*}{Manipulation} 
  & Non-learning & Task-specific or predefined goal pose             & IK + motion planning & Low  & High \\
  & Learning      & Learned implicitly via RL or IL & Learned policies via RL or IL & High & Low \\
\bottomrule
\end{tabular}
}
\end{table*}

%% file: figures/4_manipulation_tasks/grasp_methods.tex
\begin{figure}[t]
\centering
\includegraphics[width=1.0\linewidth]{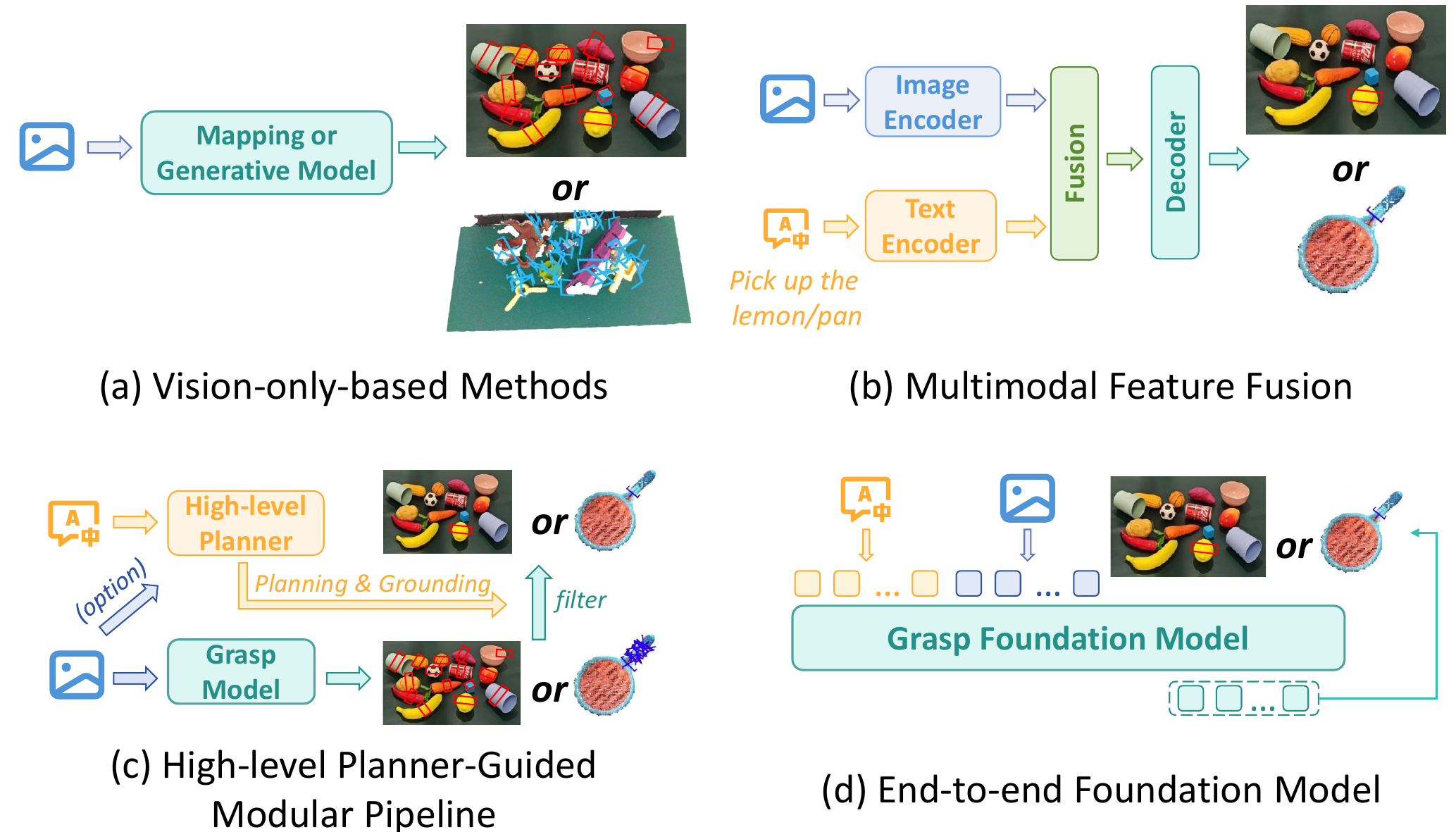}  
\vskip -0.05in
\caption{Comparison of different grasping methods. (a) Vision-only approaches, which map visual inputs to grasp poses using CNNs or spatial transformers, or generate poses via VAEs or diffusion models. With the introduction of language, three main categories emerge: (b) fusion of language and visual features through mechanisms such as cross-attention; (c) generation of multiple grasp candidates using pretrained grasp models, followed by selection using high-level reasoning modules, such as LLMs, VLMs, or 3D scene representations; and (d) end-to-end fine-tuning of grasp foundation models on large-scale grasping datasets. Figure adapted from~\cite{zhang2025vcot, fang2020graspnet, tang2025affordgrasp}.
}
\label{fig: grasp_methods}
\end{figure}

%% file: tables/4_manipulation_tasks/manipulation_tasks.tex
\begin{table*}[t]
\centering
\caption{Representative methods across manipulation types and learning paradigms.}
\label{tab: manipulation_methods}
\vskip -0.05in
\resizebox{\textwidth}{!}{
\begin{tabular}{lcccc}
\toprule
Mani. Type & RL & IL & RL+IL & VLA \\
\midrule
Dexterous & PDDM~\cite{nagabandi2020deep}, \cite{rajeswaran2018learning}, \cite{zhu2019dexterous} & DexHandDiff~\cite{liang2025dexhanddiff}, CordViP~\cite{fu2025cordvip} & REBOOT~\cite{hu2023reboot}, ViViDex~\cite{chen2025vividex} & OFA~\cite{li2025object}, LBM~\cite{barreiros2025careful} \\
Soft Robotics & \cite{thuruthel2018model}, \cite{li2022towards} & Soft DAgger~\cite{nazeer2023soft}, KineSoft~\cite{yoo2025kinesoft} & SS‑ILKC~\cite{meng2025sensor} & -- \\
Deformable Object & \cite{matas2018sim} & DeformerNet~\cite{hu20193}, DexDeform~\cite{li2023dexdeform} & DMfD~\cite{salhotra2022learning} &
-- \\
Mobile & \cite{wu2020spatial}, MoMa~\cite{hu2023causal} & MOMA-Force~\cite{yang2023moma}, Skill Transformer~\cite{huang2023skill} & \cite{xiong2024adaptive} & MoManipVLA~\cite{wu2025momanipvla} \\
Quadrupedal & VBC~\cite{liu2025visual}, GAMMA~\cite{zhang2024gamma} &
Human2LocoMan~\cite{niu2025human2locoman} & \cite{he2024learning}, WildLMa~\cite{qiu2024wildlma} & QUAR-VLA~\cite{ding2024quar}, GeRM~\cite{song2024germ} \\
Humanoid & \cite{xie2023hierarchical}, FLAM~\cite{zhang2025flam} & OmniH2O~\cite{he2025omnih2o}, iDP3~\cite{ze2025generalizable} & \cite{schakkal2025hierarchical} & GR00T N1~\cite{bjorck2025gr00t}, \newer{Humanoid-VLA~\cite{ding2025humanoid}} \\
\bottomrule
\end{tabular}
}
\end{table*}

%% file: figures/4_manipulation_tasks/case_dexterous.tex
\begin{figure}[t]
\centering
\includegraphics[width=1.0\linewidth]{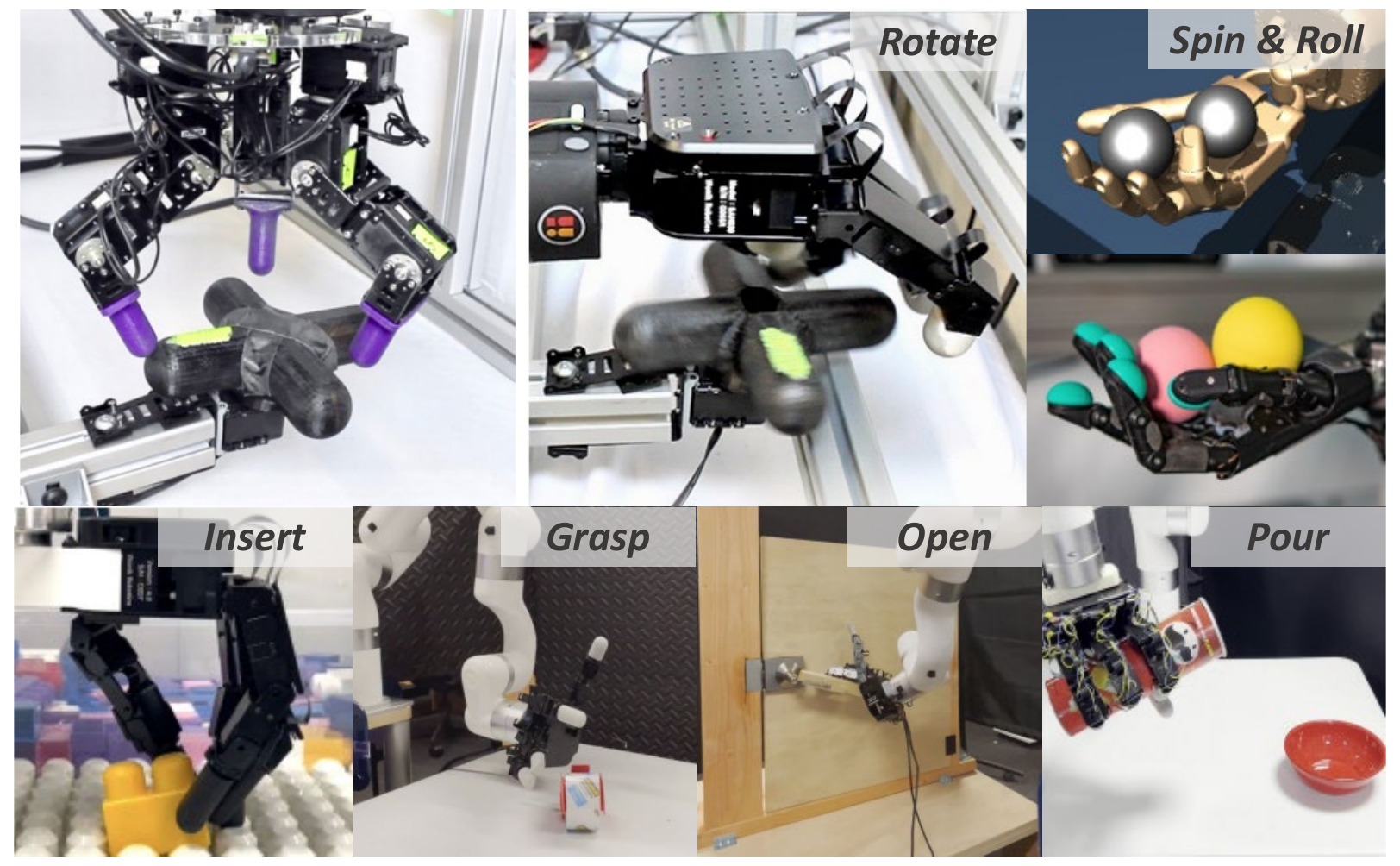}  
\vskip -0.05in
\caption{Tasks in dexterous manipulation, figure adapted from~\cite{zhu2019dexterous, nagabandi2020deep, qin2023dexpoint, chen2023sequential, wang2024cyberdemo}.
}
\label{fig: case_dex}
\end{figure}

%% file: figures/4_manipulation_tasks/case_soft.tex
\begin{figure}[t]
\centering
\includegraphics[width=1.0\linewidth]{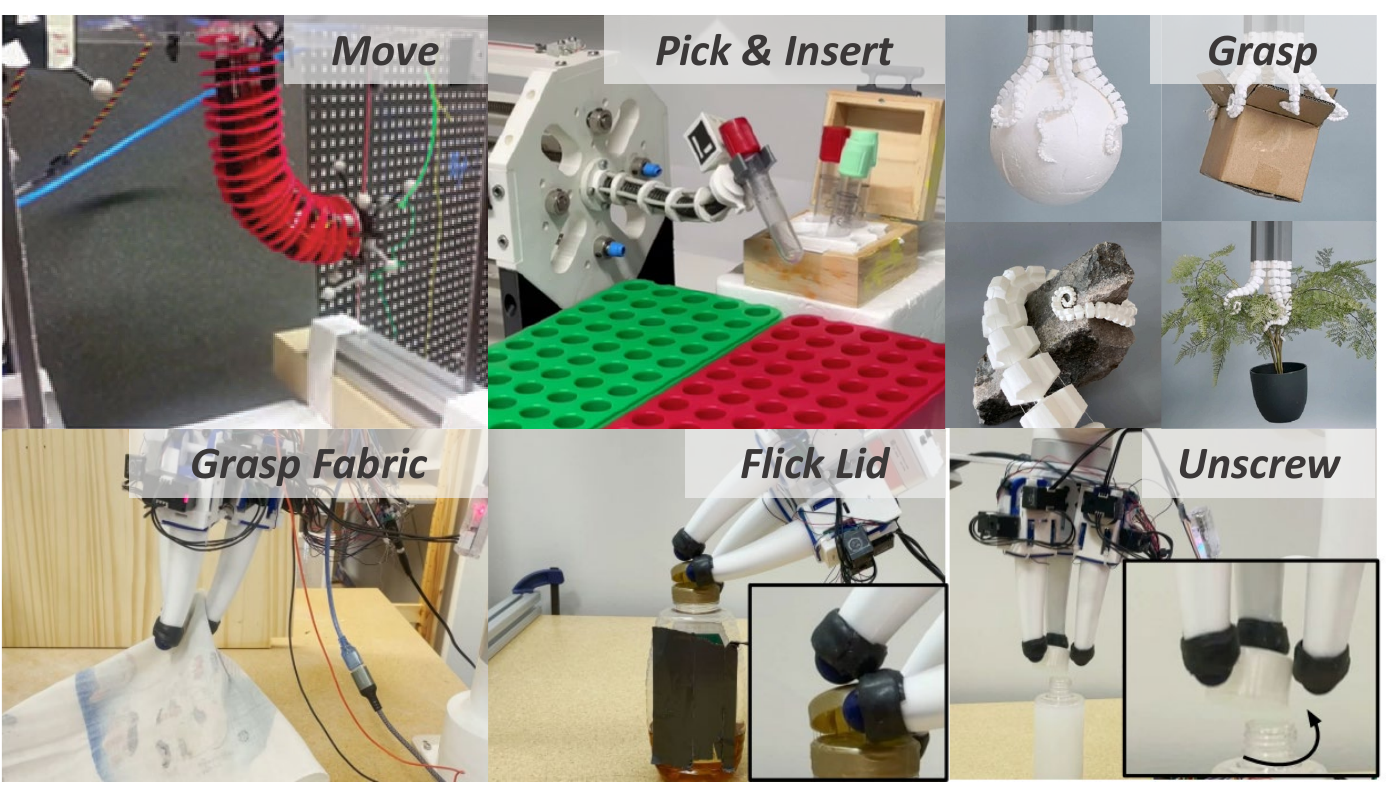}  
\vskip -0.05in
\caption{Tasks in soft robotic manipulation, figure adapted from~\cite{you2017model, yoo2025kinesoft, wang2025spirobs}.
}
\label{fig: case_soft}
\end{figure}

%% file: figures/4_manipulation_tasks/case_deformable.tex
\begin{figure}[t]
\centering
\includegraphics[width=1.0\linewidth]{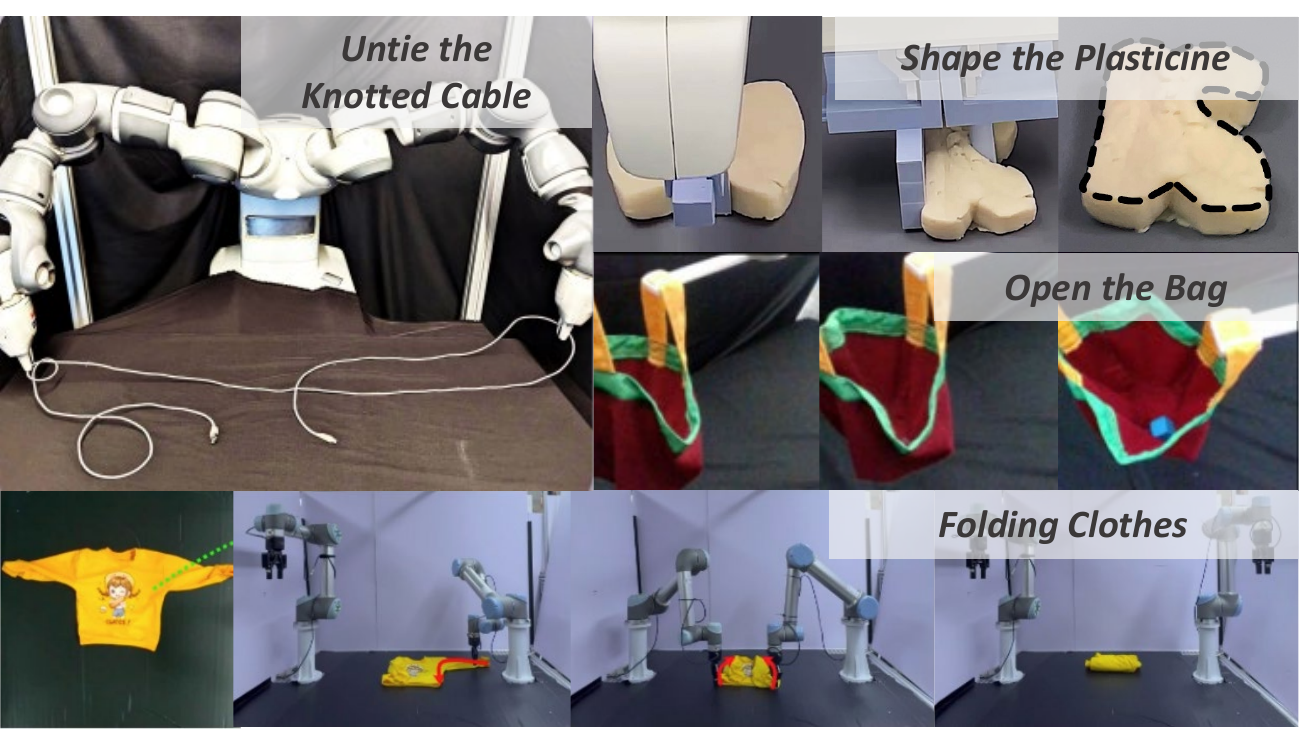}
\vskip -0.05in
\caption{Tasks in deformable object manipulation, figure adapted from~\cite{shi2023robocook, viswanath2023handloom, weng2024interactive, luan2025learning}.
}
\label{fig: case_deformable}
\end{figure}

%% file: figures/4_manipulation_tasks/case_mobile.tex
\begin{figure}[t]
\centering
\includegraphics[width=1.0\linewidth]{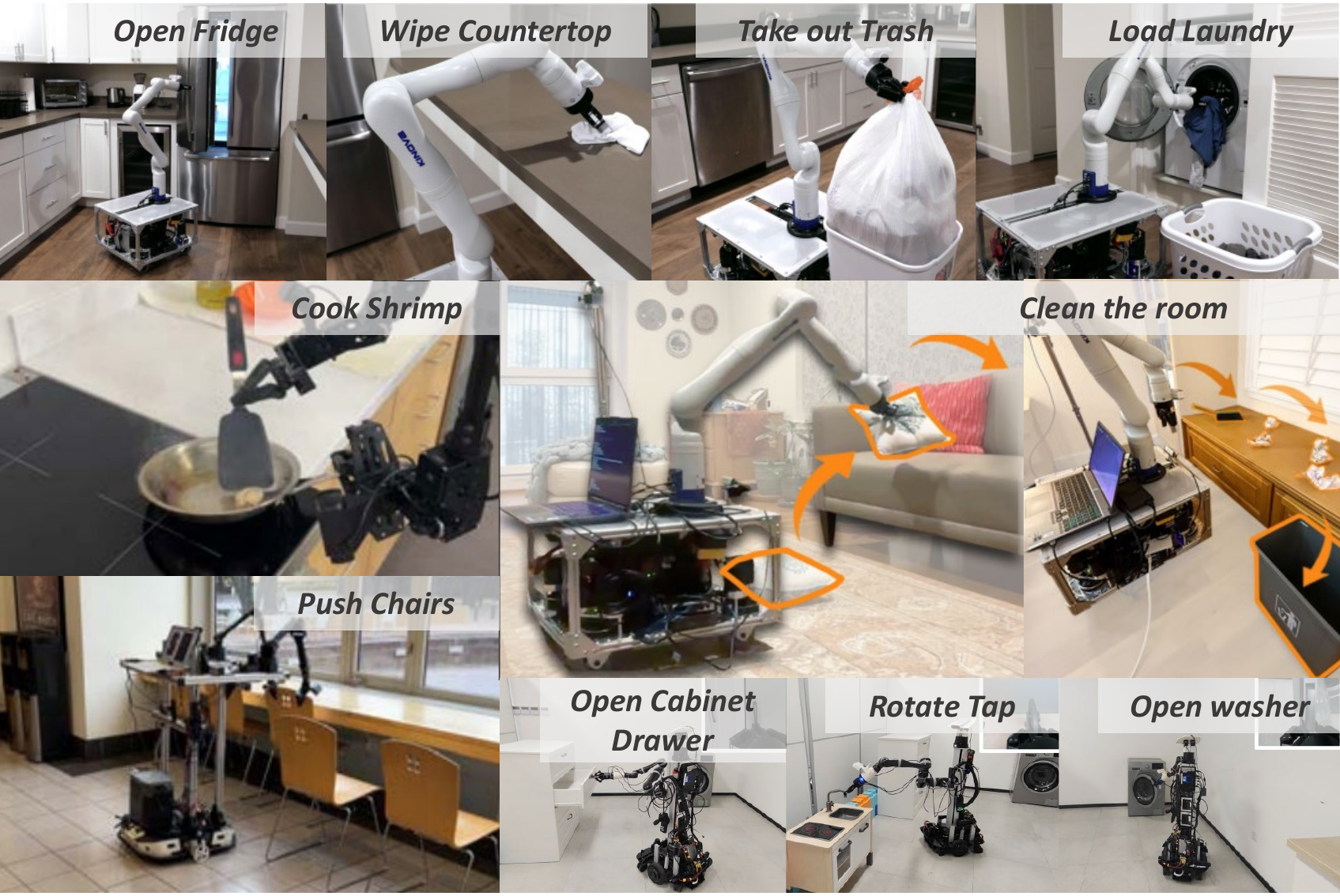}  
\vskip -0.05in
\caption{Tasks in mobile manipulation, figure adapted from~\cite{yang2023moma, fu2025mobile, wu2025tidybot++, sundaresan2025homer}.
}
\label{fig: case_mobile}
\end{figure}

%% file: figures/4_manipulation_tasks/case_quadrupedal.tex
\begin{figure}[t]
\centering
\includegraphics[width=1.0\linewidth]{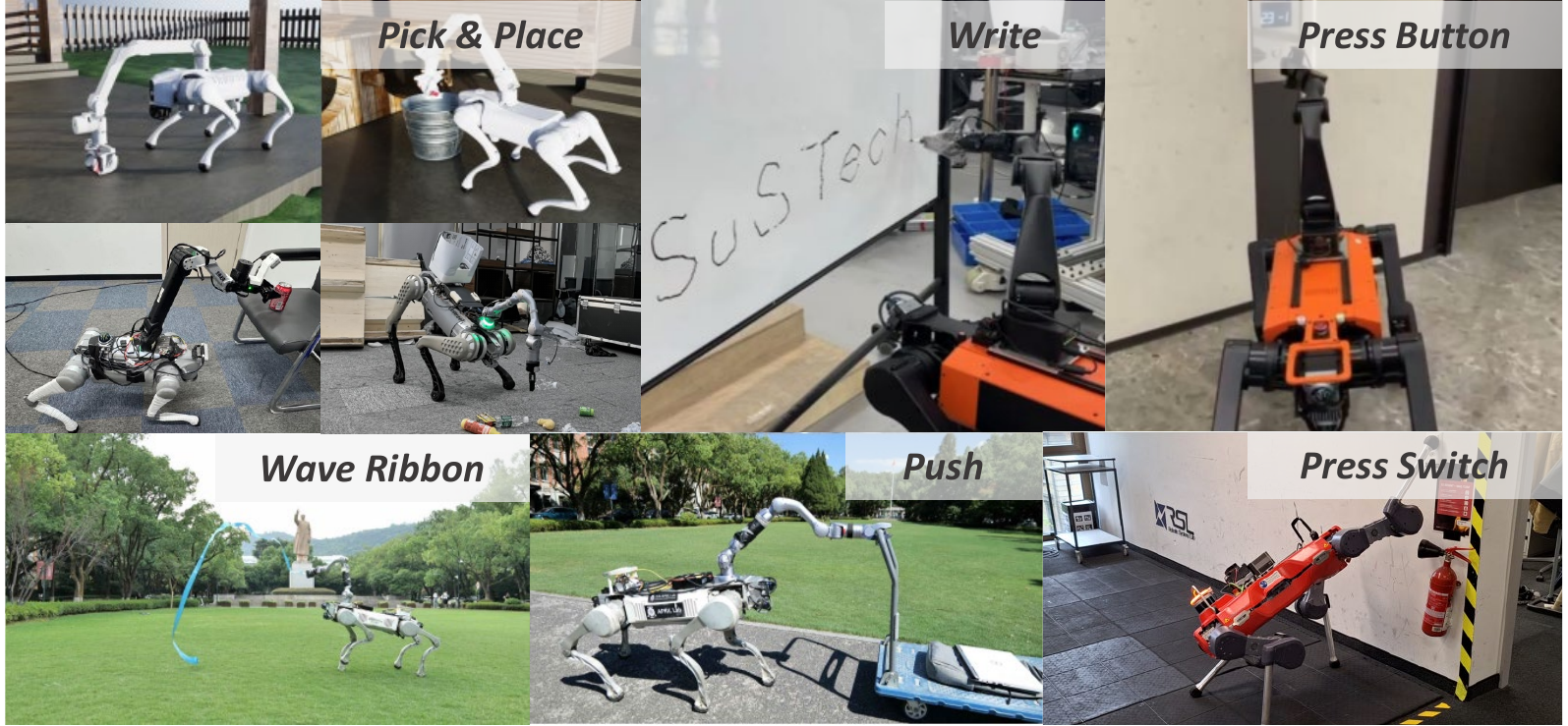}  
\vskip -0.05in
\caption{Tasks in quadrupedal manipulation, figure adapted from~\cite{jiang2024learning, arm2024pedipulate, hou2025efficient, wang2025quadwbg, wang2025odyssey}.
}
\label{fig: case_quad}
\end{figure}

%% file: figures/4_manipulation_tasks/case_humanoid.tex
\begin{figure}[t]
\centering
\includegraphics[width=1.0\linewidth]{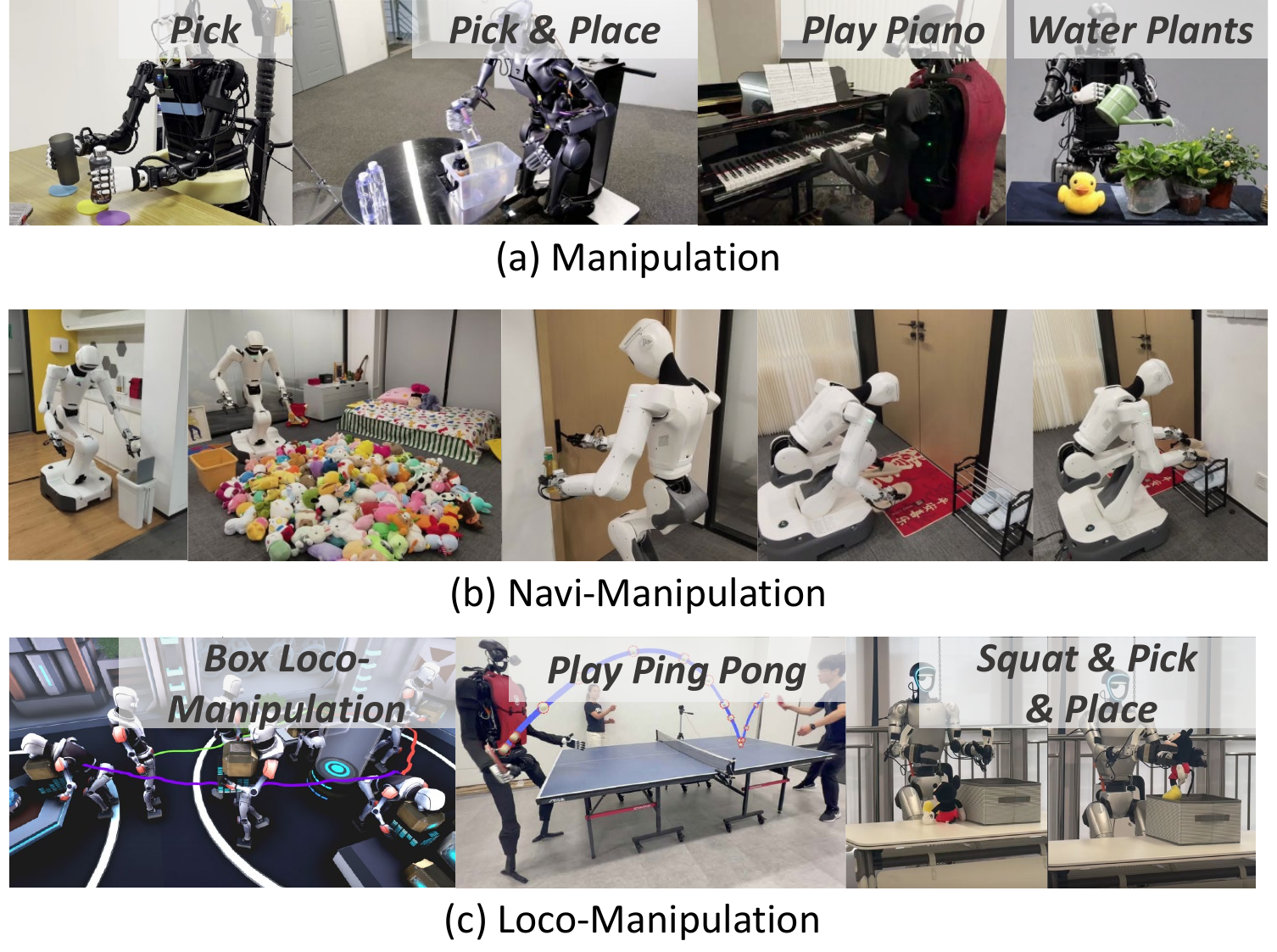}  
\vskip -0.1in
\caption{Tasks in humanoid manipulation, categorized into manipulation~\cite{xie2023hierarchical, fu2025humanplus, ze2025generalizable}, navi-manipulation~\cite{gao2025towards}, and loco-manipulation~\cite{qiu2025humanoid, fu2025humanplus, liu2025trajbooster}.
}
\label{fig: case_human}
\end{figure}

%% file: figures/7_bottlenecks/generalization.tex
\begin{figure*}[t]
\centering
\includegraphics[width=1.\linewidth]{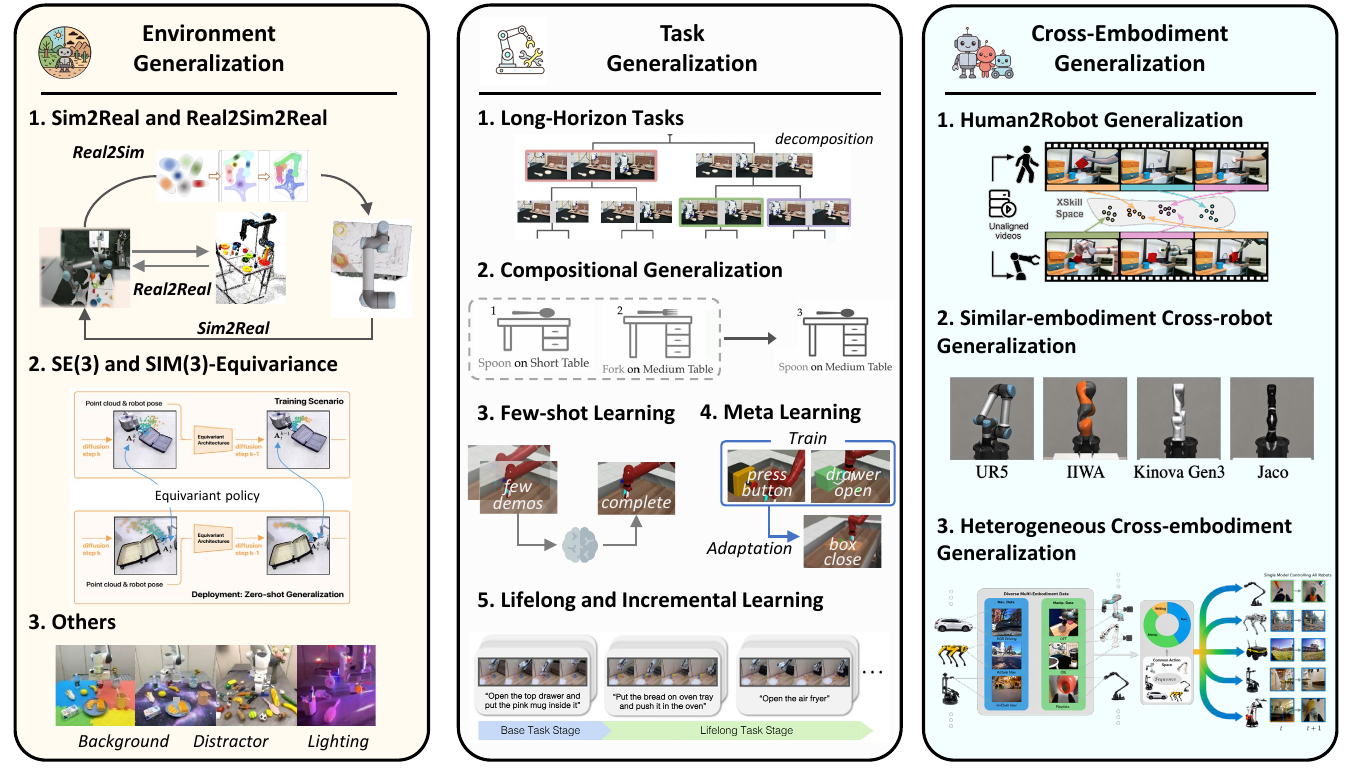}
\vskip -0.05in
\caption{Overview of generalization in robotic manipulation across three dimensions: environment generalization~\cite{li2024robogsim, yang2024equibot, deng2025graspvla}, task generalization~\cite{gao2024efficient, wan2024lotus}, and cross-embodiment generalization~\cite{xu2023xskill, yang2024pushing, chen2024mirage}. Representative settings include sim-to-real transfer and robustness to environmental variation, long-horizon and compositional task generalization, and transfer across humans, robots, and embodiments.
}   
\label{fig: generalization}
\end{figure*}

%% file: tables/11-appendix/vla_evaluation.tex
\begin{table*}[t]
\centering
\caption{Quantitative comparison of representative VLA and policy models on CALVIN~\cite{mees2022calvin}, LIBERO~\cite{liu2023libero}, SimplerEnv~\cite{li2025evaluating}.}
\label{tab:vla_quantitative_comparison}
\vskip -0.05in
\begin{tabular}{lcc|c|ccccc|ccc}
\toprule
\multirow{2}{*}{Method} & \multirow{2}{*}{Params.} & \multirow{2}{*}{Year} & CALVIN & \multicolumn{5}{c|}{LIBERO} & \multicolumn{3}{c}{SimplerEnv} \\
\cmidrule(lr){4-4}\cmidrule(lr){5-9}\cmidrule(lr){10-12}
& & & ABC$\rightarrow$D & Spatial & Object & Goal & Long & Avg & \makecell[c]{Google\\Robot VM} & \makecell[c]{Google\\Robot VA} & \makecell[c]{WidowX\\VM} \\
\midrule
OpenVLA~\cite{kim2025openvla}        & 7B   & CoRL 2024    & 3.270 & 84.7 & 88.4 & 79.2 & 53.7 & 76.5 & 45.8 & 52.4 & 4.2 \\
CogACT~\cite{li2024cogact}           & 7B   & 2024.12      & 3.250 & 95.2 & 98.6 & 95.2 & 89.6 & 94.7 & 82.7 & 66.2 & 51.3 \\
OpenVLA-OFT~\cite{kim2025fine}       & 7B   & RSS 2025     & 4.100 & 97.6 & 98.4 & 97.9 & 94.5 & 97.1 & 63.0 & 54.3 & 31.3 \\
$\pi_{0}$~\cite{black2024pi_0}       & 3B   & RSS 2025     & 3.870 & 96.8 & 98.8 & 95.8 & 85.2 & 94.2 & 79.6 & 66.1 & 69.2 \\
$\pi_{0.5}$~\cite{intelligence2025pi_} & 3B & CoRL 2025   & 4.313 & 98.8 & 98.2 & 98.0 & 92.4 & 96.8 & 72.7 & 68.4 & 69.7 \\
GR00T N1.5~\cite{bjorck2025gr00t}    & 3B   & 2025.03      & 4.010     & 96.2 & 94.0 & 96.0 & 90.0 & 94.0 & 52.4 & 43.7 & 61.9 \\
X-VLA~\cite{zheng2026x}              & 0.9B & ICLR 2026    & 4.430 & 98.2 & 98.6 & 97.8 & 97.6 & 98.1 & 80.4 & 75.7 & 95.8 \\
\bottomrule
\end{tabular}
\end{table*}

%% file: figures/11_appendix/ablation_arch.tex
\begin{figure*}[t]
\centering
\includegraphics[width=0.8\linewidth]{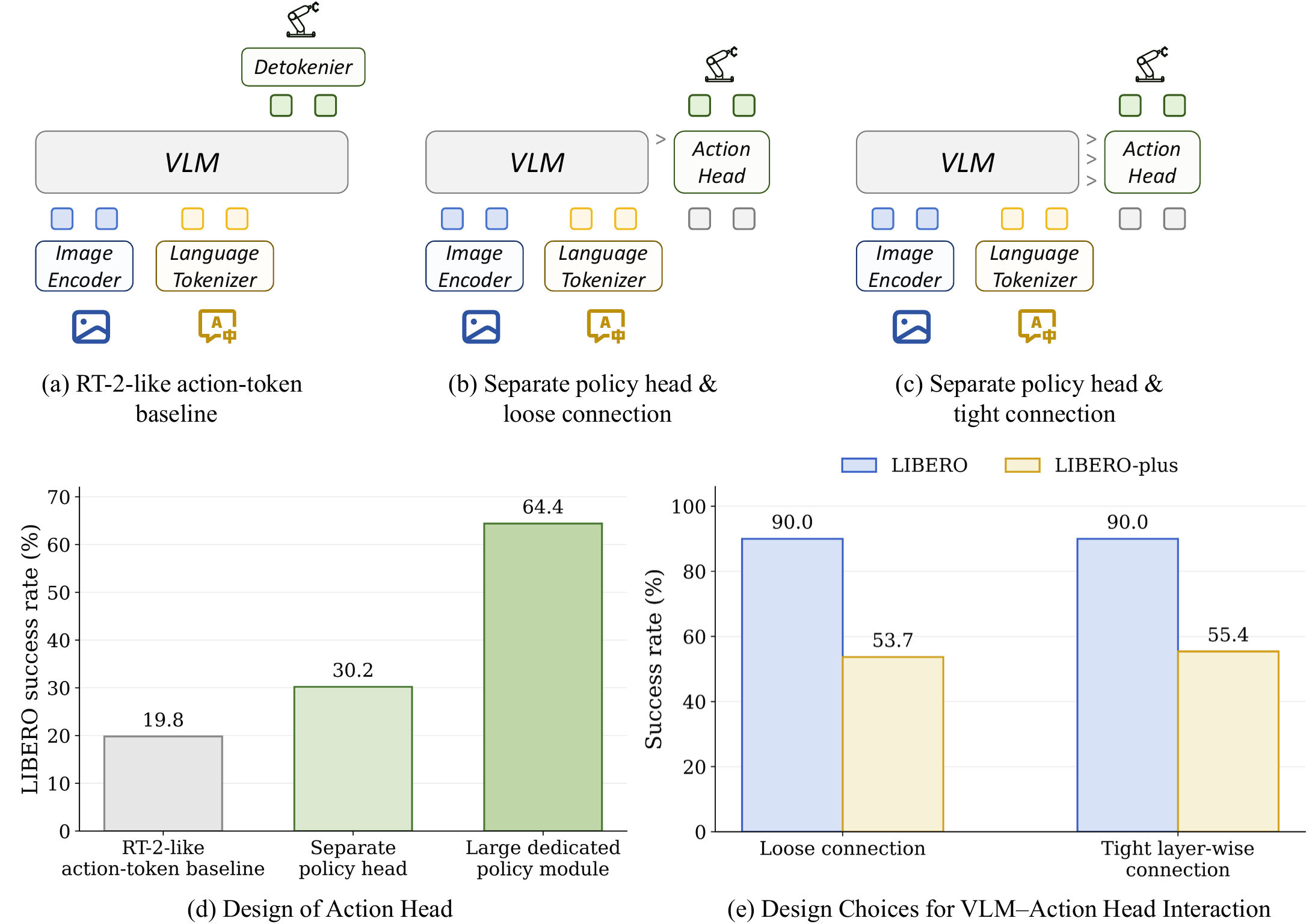}  
\vskip -0.05in
\caption{
\newer{Controlled evidence on action-module decoupling, policy capacity, and VLM--policy interaction. 
(a)--(c) illustrate direct action-token prediction, a separate action head with loose coupling, and a separate action head with tight layer-wise coupling, respectively. 
(d) shows that decoupling action generation from the VLM and increasing policy capacity substantially improve performance. 
(e) shows that tighter VLM--policy coupling provides only modest, setting-dependent gains once a sufficiently capable policy module is available.}
}
\label{fig:abla_arch}
\end{figure*}

%% file: tables/11-appendix/ablation_action.tex
\begin{table*}[t]
\centering
\caption{\newer{Controlled evidence on action representation and learning objectives. 
Here, per-dimension discrete action and bin classification denote the same general formulation: each continuous action dimension is independently quantized into scalar bins and predicted using a classification objective, although the exact binning and model configurations differ across the two studies. VQ-VAE classification instead represents an action chunk using learned discrete codebook entries. Direct regression predicts continuous action values without discretization, DDIM generates continuous action trajectories through iterative denoising, and flow matching learns a continuous vector field that transports noise samples toward action trajectories. In the final experimental block, MSE+BCE directly optimizes continuous motion dimensions with MSE and the binary gripper state with BCE, while flow matching generates the complete continuous action chunk.
CALVIN reports average consecutive task length; LIBERO and LIBERO-plus report success rates (\%).}}
\label{tab:abla_action}
\small
\vskip -0.05in
\begin{tabular}{llcc}
\toprule
\textbf{Controlled Factor} & \textbf{Variant} & \textbf{Benchmark} & \textbf{Result} \\
\midrule
\multirow{2}{*}{LLaVA action space~\cite{li2026matters}}
& Per-dimension discrete action & CALVIN ABCD$\rightarrow$D & 1.85 \\
& Continuous action & CALVIN ABCD$\rightarrow$D & \textbf{2.37} \\
\midrule
\multirow{5}{*}{Action objective~\cite{wu2026vlanext}}
& Bin classification & LIBERO / LIBERO-plus & 74.6 / 43.4 \\
& VQ-VAE classification & LIBERO / LIBERO-plus & 58.8 / 36.5 \\
& Direct regression & LIBERO / LIBERO-plus & \textbf{85.4 / 48.4} \\
& DDIM & LIBERO / LIBERO-plus & 80.0 / 48.3 \\
& Flow matching & LIBERO / LIBERO-plus & 80.0 / 45.0 \\
\midrule
\multirow{2}{*}{Objective under chunk execution~\cite{li2026matters}}
& MSE+BCE & CALVIN ABC$\rightarrow$D & 3.57 \\
& Flow matching & CALVIN ABC$\rightarrow$D & \textbf{3.68} \\
\bottomrule
\end{tabular}
\end{table*}

%% file: figures/11_appendix/ablation_action_horizon.tex
\begin{figure}[t]
\centering
\includegraphics[width=0.9\linewidth]{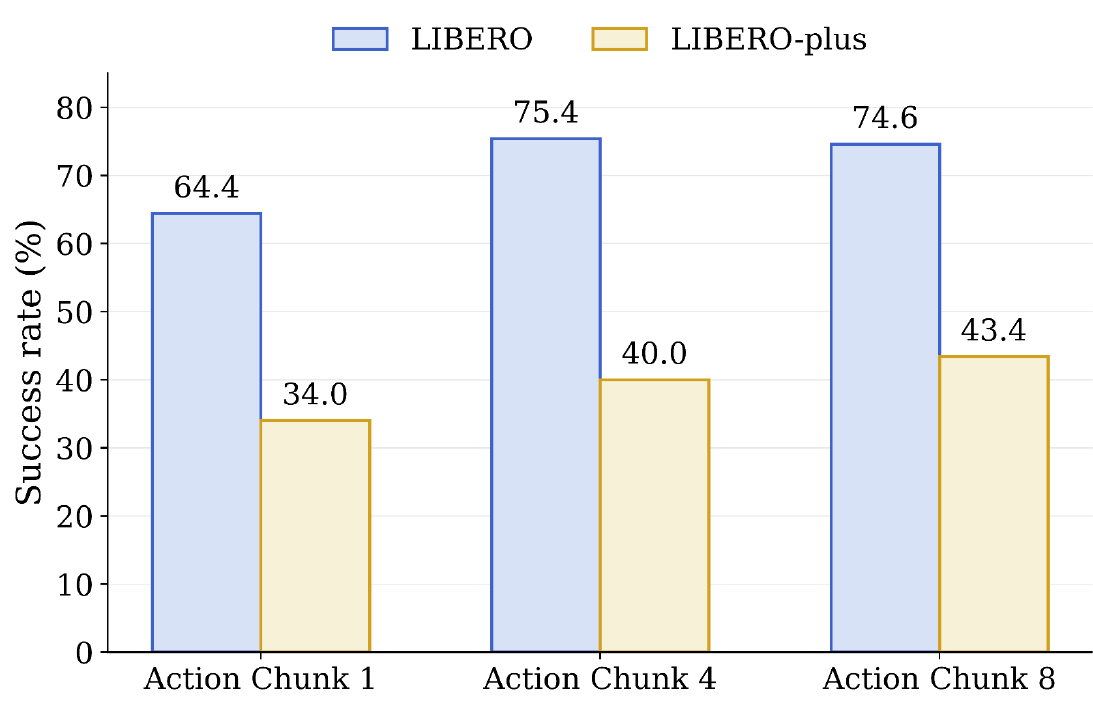}  
\vskip -0.05in
\caption{\newer{Effect of action-chunk horizon on policy performance. Action Chunk $H$ denotes jointly predicting a sequence of $H$ future actions. Relative to single-step prediction, both multi-step variants improve performance on LIBERO and LIBERO-plus. However, the preferred horizon depends on the evaluation setting: horizon 4 achieves the highest success rate on standard LIBERO, whereas horizon 8 performs best under the perturbations considered in LIBERO-plus.}}
\label{fig:abla_action_chunk}
\end{figure}

%% file: tables/11-appendix/ablation_vlm.tex
\begin{table*}[t]
\centering
\caption{\newer{Controlled evidence on VLM initialization, backbone scale, and modality-level adaptation. For rows containing two values, results are reported in the order CALVIN / SimplerEnv. CALVIN reports the average number of consecutively completed tasks, whereas SimplerEnv reports success rates (\%). Results should be compared only within each experimental block because the blocks use different backbones and adaptation settings~\cite{zhang2026vlm4vla}.}}
\label{tab:abla_vlm}
\vskip -0.05in
\small
\begin{tabularx}{\textwidth}{
    >{\raggedright\arraybackslash}p{0.20\textwidth}
    >{\raggedright\arraybackslash}p{0.28\textwidth}
    >{\centering\arraybackslash}p{0.25\textwidth}
    >{\centering\arraybackslash}X
}
\toprule
\textbf{Controlled Factor}
& \textbf{Variant}
& \textbf{Evaluation Benchmark}
& \textbf{Result} \\
\midrule

\multirow{2}{=}{Qwen2.5-VL-3B initialization~\cite{zhang2026vlm4vla}}
& Pretrained VLM initialization
& CALVIN / SimplerEnv
& \textbf{3.856 / 48.00} \\

& Random initialization
& CALVIN / SimplerEnv
& 1.381 / 15.75 \\
\midrule

\multirow{2}{=}{Qwen2.5-VL-7B initialization~\cite{zhang2026vlm4vla}}
& Pretrained VLM initialization
& CALVIN / SimplerEnv
& \textbf{4.057 / 46.75} \\

& Random initialization
& CALVIN / SimplerEnv
& 1.769 / 18.20 \\
\midrule

\multirow{2}{=}{PaliGemma initialization~\cite{zhang2026vlm4vla}}
& Pretrained VLM initialization
& CALVIN / SimplerEnv
& \textbf{3.506 / 55.25} \\

& Random initialization
& CALVIN / SimplerEnv
& 1.129 / 14.50 \\
\midrule

\multirow{4}{=}{Qwen3-VL backbone scale~\cite{zhang2026vlm4vla}}
& 2B
& CALVIN / SimplerEnv
& \textbf{4.142} / 49.0 \\

& 4B
& CALVIN / SimplerEnv
& 3.943 / 56.3 \\

& 8B
& CALVIN / SimplerEnv
& 4.035 / \textbf{58.3} \\

& 30B-A3B
& CALVIN / SimplerEnv
& 4.075 / 44.8 \\
\midrule

\multirow{3}{=}{Qwen2.5-VL-3B module adaptation~\cite{zhang2026vlm4vla}}
& Full fine-tuning
& CALVIN / SimplerEnv
& \textbf{3.856 / 48.00} \\

& Freeze vision encoder
& CALVIN / SimplerEnv
& 2.855 / 23.95 \\

& Freeze word embeddings
& CALVIN / SimplerEnv
& 3.849 / 46.88 \\
\midrule

\multirow{3}{=}{PaliGemma module adaptation~\cite{zhang2026vlm4vla}}
& Full fine-tuning
& CALVIN / SimplerEnv
& \textbf{3.506 / 55.25} \\

& Freeze vision encoder
& CALVIN / SimplerEnv
& 0.495 / 13.25 \\

& Freeze word embeddings
& CALVIN / SimplerEnv
& 3.485 / 52.25 \\
\bottomrule
\end{tabularx}
\end{table*}

%% file: tables/11-appendix/ablation_train.tex
\begin{table*}[t]
\centering
\caption{\newer{Controlled evidence on heterogeneous robot-data scaling and cross-embodiment transfer. The cumulative-mixture experiment evaluates LIBERO few-shot performance under a physically aligned end-effector-relative action representation, whereas the final experiment measures the effect of OXE pretraining on CALVIN few-shot adaptation. The results indicate that cross-embodiment pretraining can improve low-data transfer, but progressively enlarging a heterogeneous data mixture does not produce monotonic gains. LIBERO reports average success rates (\%), while the CALVIN improvement is reported as single-task success-rate points~\cite{wang2026rethinking,li2026matters}.}}
\label{tab:abla_data_scaling}
\vskip -0.05in
\small
\setlength{\tabcolsep}{4pt}
\renewcommand{\arraystretch}{1.08}
\begin{tabularx}{\textwidth}{
    >{\raggedright\arraybackslash}p{0.22\textwidth}
    >{\raggedright\arraybackslash}p{0.40\textwidth}
    >{\centering\arraybackslash}p{0.20\textwidth}
    >{\centering\arraybackslash}X
}
\toprule
\textbf{Controlled Factor}
& \textbf{Variant}
& \textbf{Evaluation Setting}
& \textbf{Result or Change} \\
\midrule

\multirow{4}{=}{Cumulative heterogeneous data mixture~\cite{wang2026rethinking}}
& Training from scratch
& LIBERO few-shot
& 66.9 \\

& OXE only
& LIBERO few-shot
& \textbf{77.3} \\

& OXE + additional real-robot EEF data
& LIBERO few-shot
& 73.8 \\

& OXE + real-robot and simulated EEF data
& LIBERO few-shot
& 72.1 \\
\midrule

Cross-embodiment initialization~\cite{li2026matters}
& OXE pretraining before target-domain few-shot adaptation
& CALVIN few-shot
& $+17.2$ \\
\bottomrule
\end{tabularx}
\end{table*}

%% file: tables/11-appendix/ablation_protocol.tex
\begin{table*}[t]
\centering
\caption{\newer{Reproducibility under a unified VLA evaluation harness~\cite{choi2026vla}. Each reproduced result is reported together with its difference from the corresponding published value in parentheses. Blank entries indicate settings that were not evaluated in the original study; therefore, no reproduced difference is reported.}}
\label{tab:abla_protocol}
\vskip -0.05in
\small
\begin{tabular}{lccc}
\toprule
\textbf{Model}
& \textbf{LIBERO}
& \textbf{CALVIN}
& \textbf{SimplerEnv} \\
\midrule

OpenVLA~\cite{kim2025openvla}
& 76.2 ($-0.3$)
& --
& -- \\

OpenVLA-OFT~\cite{kim2025fine}
& 96.7 ($-0.4$)
& --
& -- \\

GR00T N1.6~\cite{bjorck2025gr00t}
& 94.9 ($-2.1$)
& --
& 59.7 ($-8.0$) \\

CogACT~\cite{li2024cogact}
& 94.7 ($-0.2$)
& 4.02 ($-0.04$)
& 63.5 ($-6.0$) \\

X-VLA~\cite{zheng2026x}
& 97.4 ($-0.7$)
& 4.30 ($-0.13$)
& 94.8 ($-1.0$) \\
\bottomrule
\end{tabular}
\end{table*}

%% file: refs/1-intro.bib
@inproceedings{pinto2016supersizing,
  title={Supersizing self-supervision: Learning to grasp from 50k tries and 700 robot hours},
  author={Pinto, Lerrel and Gupta, Abhinav},
  booktitle={2016 IEEE international conference on robotics and automation (ICRA)},
  pages={3406--3413},
  year={2016},
  organization={IEEE}
}

@article{levine2016end,
  title={End-to-end training of deep visuomotor policies},
  author={Levine, Sergey and Finn, Chelsea and Darrell, Trevor and Abbeel, Pieter},
  journal={Journal of Machine Learning Research},
  volume={17},
  number={39},
  pages={1--40},
  year={2016}
}

@article{duan2017one,
  title={One-shot imitation learning},
  author={Duan, Yan and Andrychowicz, Marcin and Stadie, Bradly and Jonathan Ho, OpenAI and Schneider, Jonas and Sutskever, Ilya and Abbeel, Pieter and Zaremba, Wojciech},
  journal={Advances in neural information processing systems},
  volume={30},
  year={2017}
}

@inproceedings{rajeswaran2018learning,
  title={Learning Complex Dexterous Manipulation with Deep Reinforcement Learning and Demonstrations},
  author={Rajeswaran, Aravind and Kumar, Vikash and Gupta, Abhishek and Vezzani, Giulia and Schulman, John and Todorov, Emanuel and Levine, Sergey},
  booktitle={Robotics: Science and Systems},
  year={2018}
}

@article{thakar2023survey,
  title={A survey of wheeled mobile manipulation: A decision-making perspective},
  author={Thakar, Shantanu and Srinivasan, Srivatsan and Al-Hussaini, Sarah and Bhatt, Prahar M and Rajendran, Pradeep and Jung Yoon, Yeo and Dhanaraj, Neel and Malhan, Rishi K and Schmid, Matthias and Krovi, Venkat N and others},
  journal={Journal of Mechanisms and Robotics},
  volume={15},
  number={2},
  pages={020801},
  year={2023},
  publisher={American Society of Mechanical Engineers}
}

@article{gu2025humanoid,
  title={Humanoid Locomotion and Manipulation: Current Progress and Challenges in Control, Planning, and Learning},
  author={Gu, Zhaoyuan and Li, Junheng and Shen, Wenlan and Yu, Wenhao and Xie, Zhaoming and McCrory, Stephen and Cheng, Xianyi and Shamsah, Abdulaziz and Griffin, Robert and Liu, C Karen and others},
  journal={IEEEASME transactions on mechatronics},
  year={2025},
  publisher={IEEE}
}

@article{blanco2024t,
  title={T-dom: A taxonomy for robotic manipulation of deformable objects},
  author={Blanco-Mulero, David and Dong, Yifei and Borras, Julia and Pokorny, Florian T and Torras, Carme},
  journal={arXiv preprint arXiv:2412.20998},
  year={2024}
}

@article{wolf2025diffusion,
  title={Diffusion models for robotic manipulation: A survey},
  author={Wolf, Rosa Petra and Shi, Yitian and Liu, Sheng and Rayyes, Rania},
  journal={Frontiers in Robotics and AI},
  volume={12},
  pages={1606247},
  year={2025},
  publisher={Frontiers}
}

@article{zhong2025survey,
  title={A Survey on Vision-Language-Action Models: An Action Tokenization Perspective},
  author={Zhong, Yifan and Bai, Fengshuo and Cai, Shaofei and Huang, Xuchuan and Chen, Zhang and Zhang, Xiaowei and Wang, Yuanfei and Guo, Shaoyang and Guan, Tianrui and Lui, Ka Nam and others},
  journal={arXiv preprint arXiv:2507.01925},
  year={2025}
}

@article{xiang2025parallels,
  title={Parallels Between VLA Model Post-Training and Human Motor Learning: Progress, Challenges, and Trends},
  author={Xiang, Tian-Yu and Jin, Ao-Qun and Zhou, Xiao-Hu and Gui, Mei-Jiang and Xie, Xiao-Liang and Liu, Shi-Qi and Wang, Shuang-Yi and Duan, Sheng-Bin and Xie, Fu-Chao and Wang, Wen-Kai and others},
  journal={arXiv preprint arXiv:2506.20966},
  year={2025}
}

@article{zhang2025generative,
  title={Generative artificial intelligence in robotic manipulation: A survey},
  author={Zhang, Kun and Yun, Peng and Cen, Jun and Cai, Junhao and Zhu, Didi and Yuan, Hangjie and Zhao, Chao and Feng, Tao and Wang, Michael Yu and Chen, Qifeng and others},
  journal={arXiv preprint arXiv:2503.03464},
  year={2025}
}

@article{bai2025towards,
  title={Towards a unified understanding of robot manipulation: A comprehensive survey},
  author={Bai, Shuanghao and Song, Wenxuan and Chen, Jiayi and Ji, Yuheng and Zhong, Zhide and Yang, Jin and Zhao, Han and Zhou, Wanqi and Zhao, Wei and Li, Zhe and others},
  journal={arXiv preprint arXiv:2510.10903},
  year={2025}
}

@article{yang2024attribute,
  title={Attribute-based robotic grasping with data-efficient adaptation},
  author={Yang, Yang and Yu, Houjian and Lou, Xibai and Liu, Yuanhao and Choi, Changhyun},
  journal={IEEE Transactions on Robotics},
  volume={40},
  pages={1566--1579},
  year={2024},
  publisher={IEEE}
}

@article{ma2024survey,
  title={A survey on vision-language-action models for embodied ai},
  author={Ma, Yueen and Song, Zixing and Zhuang, Yuzheng and Hao, Jianye and King, Irwin},
  journal={arXiv preprint arXiv:2405.14093},
  year={2024}
}


%% file: refs/11-appendix.bib
@article{li2024cogact,
  title={Cogact: A foundational vision-language-action model for synergizing cognition and action in robotic manipulation},
  author={Li, Qixiu and Liang, Yaobo and Wang, Zeyu and Luo, Lin and Chen, Xi and Liao, Mozheng and Wei, Fangyun and Deng, Yu and Xu, Sicheng and Zhang, Yizhong and others},
  journal={arXiv preprint arXiv:2411.19650},
  year={2024}
}

@article{li2026matters,
  title={What matters in building vision--language--action models for generalist robots},
  author={Li, Xinghang and Li, Peiyan and Qian, Long and Liu, Minghuan and Wang, Dong and Liu, Jirong and Kang, Bingyi and Ma, Xiao and Wang, Xinlong and Guo, Di and others},
  journal={Nature Machine Intelligence},
  volume={8},
  number={2},
  pages={158--172},
  year={2026},
  publisher={Nature Publishing Group UK London}
}

@article{choi2026vla,
  title={vla-eval: A Unified Evaluation Harness for Vision-Language-Action Models},
  author={Choi, Suhwan and Lee, Yunsung and Park, Yubeen and Kim, Chris Dongjoo and Krishna, Ranjay and Fox, Dieter and Yu, Youngjae},
  journal={arXiv preprint arXiv:2603.13966},
  year={2026}
}

@inproceedings{wu2026vlanext,
  title={VLANeXt: Recipes for Building Strong VLA Models},
  author={Wu, Xiao-Ming and Fan, Bin and Liao, Kang and Jiang, Jian-Jian and Yang, Runze and Luo, Yihang and Wu, Zhonghua and Zheng, Wei-Shi and Loy, Chen Change},
  booktitle={Forty-third International Conference on Machine Learning},
  year={2026}
}

@article{wang2026rethinking,
  title={Rethinking visual-language-action model scaling: Alignment, mixture, and regularization},
  author={Wang, Ye and Zheng, Sipeng and Luo, Hao and Zhang, Wanpeng and Yuan, Haoqi and Xu, Chaoyi and Xu, Haiweng and Feng, Yicheng and Yu, Mingyang and Kang, Zhiyu and others},
  journal={arXiv preprint arXiv:2602.09722},
  year={2026}
}

@article{zhang2026vlm4vla,
  title={VLM4VLA: Revisiting Vision-Language-Models in Vision-Language-Action Models},
  author={Zhang, Jianke and Chen, Xiaoyu and Wang, Qiuyue and Li, Mingsheng and Guo, Yanjiang and Hu, Yucheng and Zhang, Jiajun and Bai, Shuai and Lin, Junyang and Chen, Jianyu},
  journal={arXiv preprint arXiv:2601.03309},
  year={2026}
}


%% file: refs/2-control_and_learning.bib
@article{ho2016generative,
  title={Generative adversarial imitation learning},
  author={Ho, Jonathan and Ermon, Stefano},
  journal={Advances in neural information processing systems},
  volume={29},
  year={2016}
}


%% file: refs/2-hardware.bib
@inproceedings{fu2025mobile,
  title={Mobile ALOHA: Learning Bimanual Mobile Manipulation using Low-Cost Whole-Body Teleoperation},
  author={Fu, Zipeng and Zhao, Tony Z and Finn, Chelsea},
  booktitle={Conference on Robot Learning},
  pages={4066--4083},
  year={2025},
  organization={PMLR}
}

@article{puhlmann2022rbo,
  title={RBO hand 3: A platform for soft dexterous manipulation},
  author={Puhlmann, Steffen and Harris, Jason and Brock, Oliver},
  journal={IEEE Transactions on Robotics},
  volume={38},
  number={6},
  pages={3434--3449},
  year={2022},
  publisher={IEEE}
}

@article{wang2025spirobs,
  title={SpiRobs: Logarithmic spiral-shaped robots for versatile grasping across scales},
  author={Wang, Zhanchi and Freris, Nikolaos M and Wei, Xi},
  journal={Device},
  volume={3},
  number={4},
  year={2025},
  publisher={Elsevier}
}


%% file: refs/2-robotics_models.bib
@inproceedings{he2016deep,
  title={Deep residual learning for image recognition},
  author={He, Kaiming and Zhang, Xiangyu and Ren, Shaoqing and Sun, Jian},
  booktitle={Proceedings of the IEEE conference on computer vision and pattern recognition},
  pages={770--778},
  year={2016}
}

@inproceedings{devlin2019bert,
  title={Bert: Pre-training of deep bidirectional transformers for language understanding},
  author={Devlin, Jacob and Chang, Ming-Wei and Lee, Kenton and Toutanova, Kristina},
  booktitle={Proceedings of the 2019 conference of the North American chapter of the association for computational linguistics: human language technologies, volume 1 (long and short papers)},
  pages={4171--4186},
  year={2019}
}

@article{touvron2023llama,
  title={Llama: Open and efficient foundation language models},
  author={Touvron, Hugo and Lavril, Thibaut and Izacard, Gautier and Martinet, Xavier and Lachaux, Marie-Anne and Lacroix, Timoth{\'e}e and Rozi{\`e}re, Baptiste and Goyal, Naman and Hambro, Eric and Azhar, Faisal and others},
  journal={arXiv preprint arXiv:2302.13971},
  year={2023}
}

@article{liu2023visual,
  title={Visual instruction tuning},
  author={Liu, Haotian and Li, Chunyuan and Wu, Qingyang and Lee, Yong Jae},
  journal={Advances in neural information processing systems},
  volume={36},
  pages={34892--34916},
  year={2023}
}

@inproceedings{driess2023palm,
  title={PaLM-E: An Embodied Multimodal Language Model},
  author={Driess, Danny and Xia, Fei and Sajjadi, Mehdi SM and Lynch, Corey and Chowdhery, Aakanksha and Ichter, Brian and Wahid, Ayzaan and Tompson, Jonathan and Vuong, Quan and Yu, Tianhe and others},
  booktitle={International Conference on Machine Learning},
  pages={8469--8488},
  year={2023},
  organization={PMLR}
}


%% file: refs/3-benchmarks.bib
@inproceedings{jiang2011efficient,
  title={Efficient grasping from rgbd images: Learning using a new rectangle representation},
  author={Jiang, Yun and Moseson, Stephen and Saxena, Ashutosh},
  booktitle={2011 IEEE International conference on robotics and automation},
  pages={3304--3311},
  year={2011},
  organization={IEEE}
}

@inproceedings{fang2020graspnet,
  title={Graspnet-1billion: A large-scale benchmark for general object grasping},
  author={Fang, Hao-Shu and Wang, Chenxi and Gou, Minghao and Lu, Cewu},
  booktitle={Proceedings of the IEEE/CVF conference on computer vision and pattern recognition},
  pages={11444--11453},
  year={2020}
}

@inproceedings{vuong2024language,
  title={Language-driven grasp detection},
  author={Vuong, An Dinh and Vu, Minh Nhat and Huang, Baoru and Nguyen, Nghia and Le, Hieu and Vo, Thieu and Nguyen, Anh},
  booktitle={Proceedings of the IEEE/CVF Conference on Computer Vision and Pattern Recognition},
  pages={17902--17912},
  year={2024}
}

@inproceedings{yu2020meta,
  title={Meta-world: A benchmark and evaluation for multi-task and meta reinforcement learning},
  author={Yu, Tianhe and Quillen, Deirdre and He, Zhanpeng and Julian, Ryan and Hausman, Karol and Finn, Chelsea and Levine, Sergey},
  booktitle={Conference on robot learning},
  pages={1094--1100},
  year={2020},
  organization={PMLR}
}

@article{james2020rlbench,
  title={Rlbench: The robot learning benchmark \& learning environment},
  author={James, Stephen and Ma, Zicong and Arrojo, David Rovick and Davison, Andrew J},
  journal={IEEE Robotics and Automation Letters},
  volume={5},
  number={2},
  pages={3019--3026},
  year={2020},
  publisher={IEEE}
}

@inproceedings{mandlekar2022matters,
  title={What Matters in Learning from Offline Human Demonstrations for Robot Manipulation},
  author={Mandlekar, Ajay and Xu, Danfei and Wong, Josiah and Nasiriany, Soroush and Wang, Chen and Kulkarni, Rohun and Fei-Fei, Li and Savarese, Silvio and Zhu, Yuke and Mart{\'\i}n-Mart{\'\i}n, Roberto},
  booktitle={Conference on Robot Learning},
  pages={1678--1690},
  year={2022},
  organization={PMLR}
}

@article{mees2022calvin,
  title={Calvin: A benchmark for language-conditioned policy learning for long-horizon robot manipulation tasks},
  author={Mees, Oier and Hermann, Lukas and Rosete-Beas, Erick and Burgard, Wolfram},
  journal={IEEE Robotics and Automation Letters},
  volume={7},
  number={3},
  pages={7327--7334},
  year={2022},
  publisher={IEEE}
}

@inproceedings{jiang2022vima,
  title={Vima: General robot manipulation with multimodal prompts},
  author={Jiang, Yunfan and Gupta, Agrim and Zhang, Zichen and Wang, Guanzhi and Dou, Yongqiang and Chen, Yanjun and Fei-Fei, Li and Anandkumar, Anima and Zhu, Yuke and Fan, Linxi},
  booktitle={International Conference on Machine Learning},
  year={2023}
}

@inproceedings{liu2023libero,
  title={Libero: Benchmarking knowledge transfer for lifelong robot learning},
  author={Liu, Bo and Zhu, Yifeng and Gao, Chongkai and Feng, Yihao and Liu, Qiang and Zhu, Yuke and Stone, Peter},
  booktitle={Advances in Neural Information Processing Systems},
  volume={36},
  pages={44776--44791},
  year={2023}
}

@inproceedings{li2025evaluating,
  title={Evaluating Real-World Robot Manipulation Policies in Simulation},
  author={Li, Xuanlin and Hsu, Kyle and Gu, Jiayuan and Mees, Oier and Pertsch, Karl and Walke, Homer Rich and Fu, Chuyuan and Lunawat, Ishikaa and Sieh, Isabel and Kirmani, Sean and others},
  booktitle={Conference on Robot Learning},
  pages={3705--3728},
  year={2025},
  organization={PMLR}
}

@inproceedings{zhang2025vlabench,
  title={Vlabench: A large-scale benchmark for language-conditioned robotics manipulation with long-horizon reasoning tasks},
  author={Zhang, Shiduo and Xu, Zhe and Liu, Peiju and Yu, Xiaopeng and Li, Yuan and Gao, Qinghui and Fei, Zhaoye and Yin, Zhangyue and Wu, Zuxuan and Jiang, Yu-Gang and others},
  booktitle={Proceedings of the IEEE/CVF International Conference on Computer Vision},
  pages={11142--11152},
  year={2025}
}

@article{chen2025robotwin,
  title={RoboTwin 2.0: A Scalable Data Generator and Benchmark with Strong Domain Randomization for Robust Bimanual Robotic Manipulation},
  author={Chen, Tianxing and Chen, Zanxin and Chen, Baijun and Cai, Zijian and Liu, Yibin and Liang, Qiwei and Li, Zixuan and Lin, Xianliang and Ge, Yiheng and Gu, Zhenyu and others},
  journal={arXiv preprint arXiv:2506.18088},
  year={2025}
}

@inproceedings{majumdar2023we,
  title={Where are we in the search for an artificial visual cortex for embodied intelligence?},
  author={Majumdar, Arjun and Yadav, Karmesh and Arnaud, Sergio and Ma, Jason and Chen, Claire and Silwal, Sneha and Jain, Aryan and Berges, Vincent-Pierre and Wu, Tingfan and Vakil, Jay and others},
  booktitle={Advances in Neural Information Processing Systems},
  volume={36},
  pages={655--677},
  year={2023}
}

@inproceedings{nasiriany2024robocasa,
  title={RoboCasa: Large-Scale Simulation of Everyday Tasks for Generalist Robots},
  author={Nasiriany, Soroush and Maddukuri, Abhiram and Zhang, Lance and Parikh, Adeet and Lo, Aaron and Joshi, Abhishek and Mandlekar, Ajay and Zhu, Yuke},
  booktitle={RSS 2024 Workshop: Data Generation for Robotics},
  year={2024}
}

@inproceedings{tao2025maniskill3,
  title={Maniskill3: Gpu parallelized robotics simulation and rendering for generalizable embodied ai},
  author={Tao, Stone and Xiang, Fanbo and Shukla, Arth and Qin, Yuzhe and Hinrichsen, Xander and Yuan, Xiaodi and Bao, Chen and Lin, Xinsong and Liu, Yulin and Chan, Tse-kai and others},
  booktitle={Robotics: Science and Systems},
  year={2025}
}

@inproceedings{brohan2023rt,
  title={Rt-1: Robotics transformer for real-world control at scale},
  author={Brohan, Anthony and Brown, Noah and Carbajal, Justice and Chebotar, Yevgen and Dabis, Joseph and Finn, Chelsea and Gopalakrishnan, Keerthana and Hausman, Karol and Herzog, Alex and Hsu, Jasmine and others},
  booktitle={Robotics: Science and Systems},
  year={2023}
}

@inproceedings{walke2023bridgedata,
  title={Bridgedata v2: A dataset for robot learning at scale},
  author={Walke, Homer Rich and Black, Kevin and Zhao, Tony Z and Vuong, Quan and Zheng, Chongyi and Hansen-Estruch, Philippe and He, Andre Wang and Myers, Vivek and Kim, Moo Jin and Du, Max and others},
  booktitle={Conference on Robot Learning},
  pages={1723--1736},
  year={2023},
  organization={PMLR}
}

@inproceedings{bharadhwaj2024roboagent,
  title={Roboagent: Generalization and efficiency in robot manipulation via semantic augmentations and action chunking},
  author={Bharadhwaj, Homanga and Vakil, Jay and Sharma, Mohit and Gupta, Abhinav and Tulsiani, Shubham and Kumar, Vikash},
  booktitle={2024 IEEE International Conference on Robotics and Automation (ICRA)},
  pages={4788--4795},
  year={2024},
  organization={IEEE}
}

@inproceedings{bu2025agibot,
  title={Agibot world colosseo: A large-scale manipulation platform for scalable and intelligent embodied systems},
  author={Bu, Qingwen and Cai, Jisong and Chen, Li and Cui, Xiuqi and Ding, Yan and Feng, Siyuan and Gao, Shenyuan and He, Xindong and Hu, Xuan and Huang, Xu and others},
  booktitle={2025 IEEE/RSJ International Conference on Intelligent Robots and Systems (IROS)},
  year={2025}
}


%% file: refs/4-manipulation_tasks.bib
@inproceedings{jones1990planning,
  title={Planning two-fingered grasps for pick-and-place operations on polyhedra},
  author={Jones, Joseph L and Lozano-P{\'e}rez, Tom{\'a}s},
  booktitle={Proceedings., IEEE International Conference on Robotics and Automation},
  pages={683--688},
  year={1990},
  organization={IEEE}
}

@article{miller2004graspit,
  title={Graspit! a versatile simulator for robotic grasping},
  author={Miller, Andrew T and Allen, Peter K},
  journal={IEEE Robotics \& Automation Magazine},
  volume={11},
  number={4},
  pages={110--122},
  year={2004},
  publisher={IEEE}
}

@inproceedings{stoytchev2005behavior,
  title={Behavior-grounded representation of tool affordances},
  author={Stoytchev, Alexander},
  booktitle={Proceedings of the 2005 ieee international conference on robotics and automation},
  pages={3060--3065},
  year={2005},
  organization={IEEE}
}

@inproceedings{redmon2015real,
  title={Real-time grasp detection using convolutional neural networks},
  author={Redmon, Joseph and Angelova, Anelia},
  booktitle={2015 IEEE international conference on robotics and automation (ICRA)},
  pages={1316--1322},
  year={2015},
  organization={IEEE}
}

@inproceedings{kumra2017robotic,
  title={Robotic grasp detection using deep convolutional neural networks},
  author={Kumra, Sulabh and Kanan, Christopher},
  booktitle={2017 IEEE/RSJ International Conference on Intelligent Robots and Systems (IROS)},
  pages={769--776},
  year={2017},
  organization={IEEE}
}

@inproceedings{morrison2018closing,
  title={Closing the loop for robotic grasping: a real-time, generative grasp synthesis approach},
  author={Morrison, Douglas and Corke, Peter and Leitner, J{\"u}rgen},
  booktitle={Robotics: Science and Systems 2018},
  year={2018},
  organization={The MIT Press}
}

@inproceedings{kumra2020grconvnet,
  title={Antipodal robotic grasping using generative residual convolutional neural network},
  author={Kumra, Sulabh and Joshi, Shirin and Sahin, Ferat},
  booktitle={2020 IEEE/RSJ International Conference on Intelligent Robots and Systems (IROS)},
  pages={9626--9633},
  year={2020},
  organization={IEEE}
}

@article{wang2022transformer,
  title={When transformer meets robotic grasping: Exploits context for efficient grasp detection},
  author={Wang, Shaochen and Zhou, Zhangli and Kan, Zhen},
  journal={IEEE robotics and automation letters},
  volume={7},
  number={3},
  pages={8170--8177},
  year={2022},
  publisher={IEEE}
}

@inproceedings{nguyen2024lightweight,
  title={Lightweight language-driven grasp detection using conditional consistency model},
  author={Nguyen, Nghia and Vu, Minh Nhat and Huang, Baoru and Vuong, An and Le, Ngan and Vo, Thieu and Nguyen, Anh},
  booktitle={2024 IEEE/RSJ International Conference on Intelligent Robots and Systems (IROS)},
  pages={13719--13725},
  year={2024},
  organization={IEEE}
}

@inproceedings{zhai2023monograspnet,
  title={MonoGraspNet: 6-DoF Grasping with a Single RGB Image},
  author={Zhai, Guangyao and Huang, Dianye and Wu, Shun-Cheng and Jung, HyunJun and Di, Yan and Manhardt, Fabian and Tombari, Federico and Navab, Nassir and Busam, Benjamin},
  booktitle={ICRA},
  year={2023}
}

@inproceedings{yan2018learning,
  title={Learning 6-dof grasping interaction via deep geometry-aware 3d representations},
  author={Yan, Xinchen and Hsu, Jasmined and Khansari, Mohammad and Bai, Yunfei and Pathak, Arkanath and Gupta, Abhinav and Davidson, James and Lee, Honglak},
  booktitle={2018 IEEE International Conference on Robotics and Automation (ICRA)},
  pages={3766--3773},
  year={2018},
  organization={IEEE}
}

@inproceedings{iwase2025zerograsp,
  title={ZeroGrasp: Zero-Shot Shape Reconstruction Enabled Robotic Grasping},
  author={Iwase, Shun and Irshad, Muhammad Zubair and Liu, Katherine and Guizilini, Vitor and Lee, Robert and Ikeda, Takuya and Amma, Ayako and Nishiwaki, Koichi and Kitani, Kris and Ambrus, Rares and others},
  booktitle={Proceedings of the Computer Vision and Pattern Recognition Conference},
  pages={17405--17415},
  year={2025}
}

@article{fan2025miscgrasp,
  title={MISCGrasp: Leveraging Multiple Integrated Scales and Contrastive Learning for Enhanced Volumetric Grasping},
  author={Fan, Qingyu and Cai, Yinghao and Li, Chao and Jiao, Chunting and Zheng, Xudong and Lu, Tao and Liang, Bin and Wang, Shuo},
  journal={2025 IEEE/RSJ International Conference on Intelligent Robots and Systems (IROS)},
  year={2025}
}

@inproceedings{mousavian20196dof,
  title={6-dof graspnet: Variational grasp generation for object manipulation},
  author={Mousavian, Arsalan and Eppner, Clemens and Fox, Dieter},
  booktitle={Proceedings of the IEEE/CVF international conference on computer vision},
  pages={2901--2910},
  year={2019}
}

@inproceedings{murali20206,
  title={6-dof grasping for target-driven object manipulation in clutter},
  author={Murali, Adithyavairavan and Mousavian, Arsalan and Eppner, Clemens and Paxton, Chris and Fox, Dieter},
  booktitle={2020 IEEE International Conference on Robotics and Automation (ICRA)},
  pages={6232--6238},
  year={2020},
  organization={IEEE}
}

@article{fang2023anygrasp,
  title={Anygrasp: Robust and efficient grasp perception in spatial and temporal domains},
  author={Fang, Hao-Shu and Wang, Chenxi and Fang, Hongjie and Gou, Minghao and Liu, Jirong and Yan, Hengxu and Liu, Wenhai and Xie, Yichen and Lu, Cewu},
  journal={IEEE Transactions on Robotics},
  volume={39},
  number={5},
  pages={3929--3945},
  year={2023},
  publisher={IEEE}
}

@article{xie2025rethinking,
  title={Rethinking 6-dof grasp detection: A flexible framework for high-quality grasping},
  author={Xie, Pengwei and Chen, Siang and Tang, Wei and Yang, Kaiqin and Wang, Guijin},
  journal={Pattern Recognition},
  pages={112088},
  year={2025},
  publisher={Elsevier}
}

@article{cheng2025pcf,
  title={PCF-Grasp: Converting Point Completion to Geometry Feature to Enhance 6-DoF Grasp},
  author={Cheng, Yaofeng and Zha, Fusheng and Guo, Wei and Wang, Pengfei and Zeng, Chao and Sun, Lining and Yang, Chenguang},
  journal={IEEE Transactions on Systems, Man, and Cybernetics: Systems},
  volume={56},
  number={1},
  pages={617--628},
  year={2025},
  publisher={IEEE}
}

@inproceedings{cai2025real,
  title={Real-to-Sim Grasp: Rethinking the Gap between Simulation and Real World in Grasp Detection},
  author={Cai, Jia-Feng and Chen, Zibo and Wu, Xiao-Ming and Jiang, Jian-Jian and Wei, Yi-Lin and Zheng, Wei-Shi},
  booktitle={Conference on Robot Learning},
  pages={1109--1124},
  year={2025},
  organization={PMLR}
}

@inproceedings{moghadam2023grasp,
  title={Grasp the graph (gtg): A super light graph-rl framework for robotic grasping},
  author={Moghadam, Ali Rashidi and Masouleh, Mehdi Tale and Kalhor, Ahmad},
  booktitle={2023 11th RSI International Conference on Robotics and Mechatronics (ICRoM)},
  pages={861--868},
  year={2023},
  organization={IEEE}
}

@inproceedings{hu2025orbitgrasp,
  title={OrbitGrasp: SE (3)-Equivariant Grasp Learning},
  author={Hu, Boce and Zhu, Xupeng and Wang, Dian and Dong, Zihao and Huang, Haojie and Wang, Chenghao and Walters, Robin and Platt, Robert},
  booktitle={Conference on Robot Learning},
  pages={2456--2474},
  year={2025},
  organization={PMLR}
}

@inproceedings{lim2025equigraspflow,
  title={EquiGraspFlow: SE (3)-Equivariant 6-DoF Grasp Pose Generative Flows},
  author={Lim, Byeongdo and Kim, Jongmin and Kim, Jihwan and Lee, Yonghyeon and Park, Frank C},
  booktitle={Conference on Robot Learning},
  pages={5067--5086},
  year={2025},
  organization={PMLR}
}

@article{chen2023efficient,
  title={Efficient Heatmap-Guided 6-Dof Grasp Detection in Cluttered Scenes},
  author={Chen, Siang and Tang, Wei and Xie, Pengwei and Yang, Wenming and Wang, Guijin},
  journal={IEEE Robotics and Automation Letters},
  volume={8},
  number={8},
  pages={4895--4902},
  year={2023},
  publisher={IEEE}
}

@inproceedings{sundermeyer2021contact,
  title={Contact-graspnet: Efficient 6-dof grasp generation in cluttered scenes},
  author={Sundermeyer, Martin and Mousavian, Arsalan and Triebel, Rudolph and Fox, Dieter},
  booktitle={2021 IEEE International Conference on Robotics and Automation (ICRA)},
  pages={13438--13444},
  year={2021},
  organization={IEEE}
}

@inproceedings{wu2024economic,
  title={An economic framework for 6-dof grasp detection},
  author={Wu, Xiao-Ming and Cai, Jia-Feng and Jiang, Jian-Jian and Zheng, Dian and Wei, Yi-Lin and Zheng, Wei-Shi},
  booktitle={European Conference on Computer Vision},
  pages={357--375},
  year={2024},
  organization={Springer}
}

@inproceedings{breyer2021volumetric,
  title={Volumetric grasping network: Real-time 6 dof grasp detection in clutter},
  author={Breyer, Michel and Chung, Jen Jen and Ott, Lionel and Siegwart, Roland and Nieto, Juan},
  booktitle={Conference on robot learning},
  pages={1602--1611},
  year={2021},
  organization={PMLR}
}

@inproceedings{song2025implicit,
  title={Implicit Grasp Diffusion: Bridging the Gap between Dense Prediction and Sampling-based Grasping},
  author={Song, Pinhao and Li, Pengteng and Detry, Renaud},
  booktitle={Conference on Robot Learning},
  pages={2948--2964},
  year={2025},
  organization={PMLR}
}

@inproceedings{jauhri2024learning,
  title={Learning Any-View 6DoF Robotic Grasping in Cluttered Scenes via Neural Surface Rendering},
  author={Jauhri, Snehal and Lunawat, Ishikaa and Chalvatzaki, Georgia},
  booktitle={RSS 2024 Workshop on Geometric and Algebraic Structure in Robot Learning},
  year={2024}
}

@inproceedings{rashid2023language,
  title={Language Embedded Radiance Fields for Zero-Shot Task-Oriented Grasping},
  author={Rashid, Adam and Sharma, Satvik and Kim, Chung Min and Kerr, Justin and Chen, Lawrence Yunliang and Kanazawa, Angjoo and Goldberg, Ken},
  booktitle={Conference on Robot Learning},
  pages={178--200},
  year={2023},
  organization={PMLR}
}

@inproceedings{dai2023graspnerf,
  title={GraspNeRF: Multiview-based 6-DoF Grasp Detection for Transparent and Specular Objects Using Generalizable NeRF},
  author={Dai, Qiyu and Zhu, Yan and Geng, Yiran and Ruan, Ciyu and Zhang, Jiazhao and Wang, He},
  booktitle={2023 IEEE International Conference on Robotics and Automation (ICRA)},
  pages={1757--1763},
  year={2023},
  organization={IEEE}
}

@inproceedings{ji2025graspsplats,
  title={GraspSplats: Efficient Manipulation with 3D Feature Splatting},
  author={Ji, Mazeyu and Qiu, Ri-Zhao and Zou, Xueyan and Wang, Xiaolong},
  booktitle={Conference on Robot Learning},
  pages={1443--1460},
  year={2025},
  organization={PMLR}
}

@article{zheng2024gaussiangrasper,
  title={Gaussiangrasper: 3d language gaussian splatting for open-vocabulary robotic grasping},
  author={Zheng, Yuhang and Chen, Xiangyu and Zheng, Yupeng and Gu, Songen and Yang, Runyi and Jin, Bu and Li, Pengfei and Zhong, Chengliang and Wang, Zengmao and Liu, Lina and others},
  journal={IEEE Robotics and Automation Letters},
  year={2024},
  publisher={IEEE}
}

@article{yu2024sparsegrasp,
  title={SparseGrasp: Robotic Grasping via 3D Semantic Gaussian Splatting from Sparse Multi-View RGB Images},
  author={Yu, Junqiu and Ren, Xinlin and Gu, Yongchong and Lin, Haitao and Wang, Tianyu and Zhu, Yi and Xu, Hang and Jiang, Yu-Gang and Xue, Xiangyang and Fu, Yanwei},
  journal={arXiv preprint arXiv:2412.02140},
  year={2024}
}

@inproceedings{xu2023joint,
  title={A Joint Modeling of Vision-Language-Action for Target-oriented Grasping in Clutter},
  author={Xu, Kechun and Zhao, Shuqi and Zhou, Zhongxiang and Li, Zizhang and Pi, Huaijin and Zhu, Yifeng and Wang, Yue and Xiong, Rong},
  booktitle={2023 IEEE International Conference on Robotics and Automation (ICRA)},
  pages={11597--11604},
  year={2023},
  organization={IEEE}
}

@article{tang2023graspgpt,
  title={Graspgpt: Leveraging semantic knowledge from a large language model for task-oriented grasping},
  author={Tang, Chao and Huang, Dehao and Ge, Wenqi and Liu, Weiyu and Zhang, Hong},
  journal={IEEE Robotics and Automation Letters},
  volume={8},
  number={11},
  pages={7551--7558},
  year={2023},
  publisher={IEEE}
}

@inproceedings{qian2025thinkgrasp,
  title={ThinkGrasp: A Vision-Language System for Strategic Part Grasping in Clutter},
  author={Qian, Yaoyao and Zhu, Xupeng and Biza, Ondrej and Jiang, Shuo and Zhao, Linfeng and Huang, Haojie and Qi, Yu and Platt, Robert},
  booktitle={Conference on Robot Learning},
  pages={3568--3586},
  year={2025},
  organization={PMLR}
}

@inproceedings{jin2025reasoning,
  title={Reasoning Grasping via Multimodal Large Language Model},
  author={Jin, Shiyu and XU, JINXUAN and Lei, Yutian and Zhang, Liangjun},
  booktitle={Conference on Robot Learning},
  pages={3809--3827},
  year={2025},
  publisher={PMLR}
}

@inproceedings{lu2023vl,
  title={Vl-grasp: a 6-dof interactive grasp policy for language-oriented objects in cluttered indoor scenes},
  author={Lu, Yuhao and Fan, Yixuan and Deng, Beixing and Liu, Fangfu and Li, Yali and Wang, Shengjin},
  booktitle={2023 IEEE/RSJ International Conference on Intelligent Robots and Systems (IROS)},
  pages={976--983},
  year={2023},
  organization={IEEE}
}

@inproceedings{tziafas2025towards,
  title={Towards Open-World Grasping with Large Vision-Language Models},
  author={Tziafas, Georgios and Kasaei, Hamidreza},
  booktitle={Conference on Robot Learning},
  pages={3304--3332},
  year={2025},
  organization={PMLR}
}

@inproceedings{xu2024rt,
  title={Rt-grasp: Reasoning tuning robotic grasping via multi-modal large language model},
  author={Xu, Jinxuan and Jin, Shiyu and Lei, Yutian and Zhang, Yuqian and Zhang, Liangjun},
  booktitle={2024 IEEE/RSJ International Conference on Intelligent Robots and Systems (IROS)},
  pages={7323--7330},
  year={2024},
  organization={IEEE}
}

@article{dong2025rtagrasp,
  title={Rtagrasp: Learning task-oriented grasping from human videos via retrieval, transfer, and alignment},
  author={Dong, Wenlong and Huang, Dehao and Liu, Jiangshan and Tang, Chao and Zhang, Hong},
  journal={2025 IEEE international conference on robotics and automation (ICRA)},
  year={2025}
}

@article{song2025learning,
  title={Learning 6-dof fine-grained grasp detection based on part affordance grounding},
  author={Song, Yaoxian and Sun, Penglei and Jin, Piaopiao and Ren, Yi and Zheng, Yu and Li, Zhixu and Chu, Xiaowen and Zhang, Yue and Li, Tiefeng and Gu, Jason},
  journal={IEEE Transactions on Automation Science and Engineering},
  year={2025},
  publisher={IEEE}
}

@inproceedings{deshpande2025graspmolmo,
  title={GraspMolmo: Generalizable Task-Oriented Grasping via Large-Scale Synthetic Data Generation},
  author={Deshpande, Abhay and Deng, Yuquan and Salvador, Jordi and Ray, Arijit and Han, Winson and Duan, Jiafei and Hendrix, Rose and Zhu, Yuke and Krishna, Ranjay},
  booktitle={Conference on Robot Learning},
  pages={2983--3007},
  year={2025},
  organization={PMLR}
}

@article{luo2025roboreflect,
  title={RoboReflect: A Robotic Reflective Reasoning Framework for Grasping Ambiguous-Condition Objects},
  author={Luo, Zhen and Yang, Yixuan and Zhang, Yanfu and Zheng, Feng},
  journal={arXiv preprint arXiv:2501.09307},
  year={2025}
}

@inproceedings{tang2025affordgrasp,
  title={Affordgrasp: In-context affordance reasoning for open-vocabulary task-oriented grasping in clutter},
  author={Tang, Yingbo and Zhang, Shuaike and Hao, Xiaoshuai and Wang, Pengwei and Wu, Jianlong and Wang, Zhongyuan and Zhang, Shanghang},
  booktitle={2025 IEEE/RSJ International Conference on Intelligent Robots and Systems (IROS)},
  pages={9433--9439},
  year={2025},
  organization={IEEE}
}

@article{weng2024dexdiffuser,
  title={Dexdiffuser: Generating dexterous grasps with diffusion models},
  author={Weng, Zehang and Lu, Haofei and Kragic, Danica and Lundell, Jens},
  journal={IEEE Robotics and Automation Letters},
  year={2024},
  publisher={IEEE}
}

@inproceedings{xu2024dexterous,
  title={Dexterous grasp transformer},
  author={Xu, Guo-Hao and Wei, Yi-Lin and Zheng, Dian and Wu, Xiao-Ming and Zheng, Wei-Shi},
  booktitle={Proceedings of the IEEE/CVF Conference on Computer Vision and Pattern Recognition},
  pages={17933--17942},
  year={2024}
}

@article{wei2024grasp,
  title={Grasp as you say: Language-guided dexterous grasp generation},
  author={Wei, Yi-Lin and Jiang, Jian-Jian and Xing, Chengyi and Tan, Xian-Tuo and Wu, Xiao-Ming and Li, Hao and Cutkosky, Mark and Zheng, Wei-Shi},
  journal={Advances in Neural Information Processing Systems},
  volume={37},
  pages={46881--46907},
  year={2024}
}

@inproceedings{li2023gendexgrasp,
  title={GenDexGrasp: Generalizable Dexterous Grasping},
  author={Li, Puhao and Liu, Tengyu and Li, Yuyang and Geng, Yiran and Zhu, Yixin and Yang, Yaodong and Huang, Siyuan},
  booktitle={2023 IEEE International Conference on Robotics and Automation (ICRA)},
  pages={8068--8074},
  year={2023},
  organization={IEEE}
}

@inproceedings{wei2024d,
  title={D (r, o) Grasp: A Unified Representation of Robot and Object Interaction for Cross-Embodiment Dexterous Grasping},
  author={Wei, Zhenyu and Xu, Zhixuan and Guo, Jingxiang and Hou, Yiwen and Gao, Chongkai and Cai, Zhehao and Luo, Jiayu and Shao, Lin},
  booktitle={CoRL Workshop on Learning Robot Fine and Dexterous Manipulation: Perception and Control},
  year={2024}
}

@article{mahler2019learning,
  title={Learning ambidextrous robot grasping policies},
  author={Mahler, Jeffrey and Matl, Matthew and Satish, Vishal and Danielczuk, Michael and DeRose, Bill and McKinley, Stephen and Goldberg, Ken},
  journal={Science Robotics},
  volume={4},
  number={26},
  pages={eaau4984},
  year={2019},
  publisher={American Association for the Advancement of Science}
}

@inproceedings{yamada2025combo,
  title={COMBO-Grasp: Learning Constraint-Based Manipulation for Bimanual Occluded Grasping},
  author={Yamada, Jun and Mitchell, Alexander Luis and Collins, Jack and Posner, Ingmar},
  booktitle={Conference on Robot Learning},
  pages={1102--1119},
  year={2025},
  organization={PMLR}
}

@article{deng2025fusegrasp,
  title={Fusegrasp: Radar-camera fusion for robotic grasping of transparent objects},
  author={Deng, Hongyu and Xue, Tianfan and Chen, He},
  journal={IEEE Transactions on Mobile Computing},
  year={2025},
  publisher={IEEE}
}

@inproceedings{duisterhof2024residual,
  title={Residual-nerf: Learning residual nerfs for transparent object manipulation},
  author={Duisterhof, Bardienus P and Mao, Yuemin and Teng, Si Heng and Ichnowski, Jeffrey},
  booktitle={2024 IEEE International Conference on Robotics and Automation (ICRA)},
  pages={13918--13924},
  year={2024},
  organization={IEEE}
}

@inproceedings{shi2024asgrasp,
  title={Asgrasp: Generalizable transparent object reconstruction and 6-dof grasp detection from rgb-d active stereo camera},
  author={Shi, Jun and Yong, A and Jin, Yixiang and Li, Dingzhe and Niu, Haoyu and Jin, Zhezhu and Wang, He},
  booktitle={2024 IEEE international conference on robotics and automation (ICRA)},
  pages={5441--5447},
  year={2024},
  organization={IEEE}
}

@inproceedings{kim2025t,
  title={T\textsuperscript{2}SQNet: A Recognition Model for Manipulating Partially Observed Transparent Tableware Objects},
  author={Kim, Young Hun and Kim, Seungyeon and Lee, Yonghyeon and Park, Frank C},
  booktitle={Conference on Robot Learning},
  pages={3622--3655},
  year={2025},
  organization={PMLR}
}

@article{welte2025interactive,
  title={Interactive imitation learning for dexterous robotic manipulation: challenges and perspectives—a survey},
  author={Welte, Edgar and Rayyes, Rania},
  journal={Frontiers in Robotics and AI},
  volume={12},
  pages={1682437},
  year={2025},
  publisher={Frontiers Media SA}
}

@inproceedings{bai2014dexterous,
  title={Dexterous manipulation using both palm and fingers},
  author={Bai, Yunfei and Liu, C Karen},
  booktitle={2014 IEEE International Conference on Robotics and Automation (ICRA)},
  pages={1560--1565},
  year={2014},
  organization={IEEE}
}

@article{luo2025precise,
  title={Precise and dexterous robotic manipulation via human-in-the-loop reinforcement learning},
  author={Luo, Jianlan and Xu, Charles and Wu, Jeffrey and Levine, Sergey},
  journal={Science Robotics},
  volume={10},
  number={105},
  pages={eads5033},
  year={2025},
  publisher={American Association for the Advancement of Science}
}

@article{kumar2016learning,
  title={Learning dexterous manipulation policies from experience and imitation},
  author={Kumar, Vikash and Gupta, Abhishek and Todorov, Emanuel and Levine, Sergey},
  journal={arXiv preprint arXiv:1611.05095},
  year={2016}
}

@inproceedings{nagabandi2020deep,
  title={Deep dynamics models for learning dexterous manipulation},
  author={Nagabandi, Anusha and Konolige, Kurt and Levine, Sergey and Kumar, Vikash},
  booktitle={Conference on robot learning},
  pages={1101--1112},
  year={2020},
  organization={PMLR}
}

@inproceedings{zhu2019dexterous,
  title={Dexterous manipulation with deep reinforcement learning: Efficient, general, and low-cost},
  author={Zhu, Henry and Gupta, Abhishek and Rajeswaran, Aravind and Levine, Sergey and Kumar, Vikash},
  booktitle={2019 International Conference on Robotics and Automation (ICRA)},
  pages={3651--3657},
  year={2019},
  organization={IEEE}
}

@inproceedings{arunachalam2023dexterous,
  title={Dexterous Imitation Made Easy: A Learning-Based Framework for Efficient Dexterous Manipulation},
  author={Arunachalam, Sridhar Pandian and Silwal, Sneha and Evans, Ben and Pinto, Lerrel},
  booktitle={2023 IEEE International Conference on Robotics and Automation (ICRA)},
  pages={5954--5961},
  year={2023},
  organization={IEEE}
}

@inproceedings{hu2023reboot,
  title={REBOOT: Reuse Data for Bootstrapping Efficient Real-World Dexterous Manipulation},
  author={Hu, Zheyuan and Rovinsky, Aaron and Luo, Jianlan and Kumar, Vikash and Gupta, Abhishek and Levine, Sergey},
  booktitle={7th Annual Conference on Robot Learning},
  year={2023}
}

@inproceedings{kannan2023deft,
  title={DEFT: Dexterous Fine-Tuning for Hand Policies},
  author={Kannan, Aditya and Shaw, Kenneth and Bahl, Shikhar and Mannam, Pragna and Pathak, Deepak},
  booktitle={Conference on Robot Learning},
  pages={928--942},
  year={2023},
  organization={PMLR}
}

@inproceedings{agarwal2023dexterous,
  title={Dexterous Functional Grasping},
  author={Agarwal, Ananye and Uppal, Shagun and Shaw, Kenneth and Pathak, Deepak},
  booktitle={Conference on Robot Learning},
  pages={3453--3467},
  year={2023},
  organization={PMLR}
}

@inproceedings{wang2024cyberdemo,
  title={Cyberdemo: Augmenting simulated human demonstration for real-world dexterous manipulation},
  author={Wang, Jun and Qin, Yuzhe and Kuang, Kaiming and Korkmaz, Yigit and Gurumoorthy, Akhilan and Su, Hao and Wang, Xiaolong},
  booktitle={Proceedings of the IEEE/CVF Conference on Computer Vision and Pattern Recognition},
  pages={17952--17963},
  year={2024}
}

@inproceedings{qin2022dexmv,
  title={Dexmv: Imitation learning for dexterous manipulation from human videos},
  author={Qin, Yuzhe and Wu, Yueh-Hua and Liu, Shaowei and Jiang, Hanwen and Yang, Ruihan and Fu, Yang and Wang, Xiaolong},
  booktitle={European Conference on Computer Vision},
  pages={570--587},
  year={2022},
  organization={Springer}
}

@inproceedings{chen2025vividex,
  title={Vividex: Learning vision-based dexterous manipulation from human videos},
  author={Chen, Zerui and Chen, Shizhe and Arlaud, Etienne and Laptev, Ivan and Schmid, Cordelia},
  booktitle={2025 IEEE international conference on robotics and automation (ICRA)},
  year={2025}
}

@inproceedings{liang2025dexhanddiff,
  title={Dexhanddiff: Interaction-aware diffusion planning for adaptive dexterous manipulation},
  author={Liang, Zhixuan and Mu, Yao and Wang, Yixiao and Chen, Tianxing and Shao, Wenqi and Zhan, Wei and Tomizuka, Masayoshi and Luo, Ping and Ding, Mingyu},
  booktitle={Proceedings of the Computer Vision and Pattern Recognition Conference},
  pages={1745--1755},
  year={2025}
}

@inproceedings{fu2025cordvip,
  title={Cordvip: Correspondence-based visuomotor policy for dexterous manipulation in real-world},
  author={Fu, Yankai and Feng, Qiuxuan and Chen, Ning and Zhou, Zichen and Liu, Mengzhen and Wu, Mingdong and Chen, Tianxing and Rong, Shanyu and Liu, Jiaming and Dong, Hao and others},
  booktitle={Robotics: Science and Systems},
  year={2025}
}

@article{li2025object,
  title={Object-Focus Actor for Data-efficient Robot Generalization Dexterous Manipulation},
  author={Li, Yihang and Zhang, Tianle and Wei, Xuelong and Li, Jiayi and Zhao, Lin and Huang, Dongchi and Fang, Zhirui and Zheng, Minhua and Dai, Wenjun and He, Xiaodong},
  journal={arXiv preprint arXiv:2505.15098},
  year={2025}
}

@article{barreiros2025careful,
  title={A Careful Examination of Large Behavior Models for Multitask Dexterous Manipulation},
  author={Barreiros, Jose and Beaulieu, Andrew and Bhat, Aditya and Cory, Rick and Cousineau, Eric and Dai, Hongkai and Fang, Ching-Hsin and Hashimoto, Kunimatsu and Irshad, Muhammad Zubair and Itkina, Masha and others},
  journal={arXiv preprint arXiv:2507.05331},
  year={2025}
}

@inproceedings{chen2025object,
  title={Object-Centric Dexterous Manipulation from Human Motion Data},
  author={Chen, Yuanpei and Wang, Chen and Yang, Yaodong and Liu, Karen},
  booktitle={Conference on Robot Learning},
  pages={3828--3846},
  year={2025},
  organization={PMLR}
}

@inproceedings{radosavovic2021state,
  title={State-only imitation learning for dexterous manipulation},
  author={Radosavovic, Ilija and Wang, Xiaolong and Pinto, Lerrel and Malik, Jitendra},
  booktitle={2021 IEEE/RSJ International Conference on Intelligent Robots and Systems (IROS)},
  pages={7865--7871},
  year={2021},
  organization={IEEE}
}

@inproceedings{dasari2023learning,
  title={Learning Dexterous Manipulation from Exemplar Object Trajectories and Pre-Grasps},
  author={Dasari, Sudeep and Gupta, Abhinav and Kumar, Vikash},
  booktitle={2023 IEEE International Conference on Robotics and Automation (ICRA)},
  pages={3889--3896},
  year={2023},
  organization={IEEE}
}

@article{mandi2025dexmachina,
  title={Dexmachina: Functional retargeting for bimanual dexterous manipulation},
  author={Mandi, Zhao and Hou, Yifan and Fox, Dieter and Narang, Yashraj and Mandlekar, Ajay and Song, Shuran},
  journal={arXiv preprint arXiv:2505.24853},
  year={2025}
}

@inproceedings{qin2023dexpoint,
  title={Dexpoint: Generalizable point cloud reinforcement learning for sim-to-real dexterous manipulation},
  author={Qin, Yuzhe and Huang, Binghao and Yin, Zhao-Heng and Su, Hao and Wang, Xiaolong},
  booktitle={Conference on Robot Learning},
  pages={594--605},
  year={2023},
  organization={PMLR}
}

@inproceedings{chen2023sequential,
  title={Sequential Dexterity: Chaining Dexterous Policies for Long-Horizon Manipulation},
  author={Chen, Yuanpei and Wang, Chen and Fei-Fei, Li and Liu, Karen},
  booktitle={Conference on Robot Learning},
  pages={3809--3829},
  year={2023},
  organization={PMLR}
}

@article{chen2022review,
  title={A review of soft manipulator research, applications, and opportunities},
  author={Chen, Xiaoqian and Zhang, Xiang and Huang, Yiyong and Cao, Lu and Liu, Jinguo},
  journal={Journal of Field Robotics},
  volume={39},
  number={3},
  pages={281--311},
  year={2022},
  publisher={Wiley Online Library}
}

@article{jones2006kinematics,
  title={Kinematics for multisection continuum robots},
  author={Jones, Bryan A and Walker, Ian D},
  journal={IEEE Transactions on Robotics},
  volume={22},
  number={1},
  pages={43--55},
  year={2006},
  publisher={IEEE}
}

@article{wang2020dynamic,
  title={Dynamic control of multisection three-dimensional continuum manipulators based on virtual discrete-jointed robot models},
  author={Wang, Chengshi and Frazelle, Chase G and Wagner, John R and Walker, Ian D},
  journal={IEEE/ASME Transactions on Mechatronics},
  volume={26},
  number={2},
  pages={777--788},
  year={2020},
  publisher={IEEE}
}

@inproceedings{lv2025multi,
  title={Multi-Segment Soft Robot Control Via Deep Koopman-Based Model Predictive Control},
  author={Lv, Lei and Liu, Lei and Bao, Lei and Sun, Fuchun and Dong, Jiahong and Zhang, Jianwei and Shan, Xuemei and Sun, Kai and Huang, Hao and Luo, Yu},
  booktitle={2025 IEEE International Conference on Robotics and Automation (ICRA)},
  pages={9266--9272},
  year={2025},
  organization={IEEE}
}

@inproceedings{kazemipour2022adaptive,
  title={Adaptive dynamic sliding mode control of soft continuum manipulators},
  author={Kazemipour, Amirhossein and Fischer, Oliver and Toshimitsu, Yasunori and Wong, Ki Wan and Katzschmann, Robert K},
  booktitle={2022 International Conference on Robotics and Automation (ICRA)},
  pages={3259--3265},
  year={2022},
  organization={IEEE}
}

@inproceedings{bruder2019modeling,
  title={Modeling and control of soft robots using the koopman operator and model predictive control},
  author={Bruder, Daniel and Gillespie, Brent and Remy, C David and Vasudevan, Ram},
  booktitle={Robotics: Science and Systems},
  year={2019}
}

@article{thuruthel2018stable,
  title={Stable open loop control of soft robotic manipulators},
  author={Thuruthel, Thomas George and Falotico, Egidio and Manti, Mariangela and Laschi, Cecilia},
  journal={IEEE Robotics and Automation Letters},
  volume={3},
  number={2},
  pages={1292--1298},
  year={2018},
  publisher={IEEE}
}

@article{thuruthel2017learning,
  title={Learning dynamic models for open loop predictive control of soft robotic manipulators},
  author={Thuruthel, Thomas George and Falotico, Egidio and Renda, Federico and Laschi, Cecilia},
  journal={Bioinspiration \& biomimetics},
  volume={12},
  number={6},
  pages={066003},
  year={2017},
  publisher={IOP Publishing}
}

@inproceedings{you2017model,
  title={Model-free control for soft manipulators based on reinforcement learning},
  author={You, Xuanke and Zhang, Yixiao and Chen, Xiaotong and Liu, Xinghua and Wang, Zhanchi and Jiang, Hao and Chen, Xiaoping},
  booktitle={2017 IEEE/RSJ international conference on intelligent robots and systems (IROS)},
  pages={2909--2915},
  year={2017},
  organization={IEEE}
}

@article{thuruthel2018model,
  title={Model-based reinforcement learning for closed-loop dynamic control of soft robotic manipulators},
  author={Thuruthel, Thomas George and Falotico, Egidio and Renda, Federico and Laschi, Cecilia},
  journal={IEEE Transactions on Robotics},
  volume={35},
  number={1},
  pages={124--134},
  year={2018},
  publisher={IEEE}
}

@article{centurelli2022closed,
  title={Closed-loop dynamic control of a soft manipulator using deep reinforcement learning},
  author={Centurelli, Andrea and Arleo, Luca and Rizzo, Alessandro and Tolu, Silvia and Laschi, Cecilia and Falotico, Egidio},
  journal={IEEE Robotics and Automation Letters},
  volume={7},
  number={2},
  pages={4741--4748},
  year={2022},
  publisher={IEEE}
}

@article{zhou2024cable,
  title={A Cable-Actuated Soft Manipulator for Dexterous Grasping Based on Deep Reinforcement Learning},
  author={Zhou, Kunyu and Mao, Baijin and Zhang, Yuzhu and Chen, Yaozhen and Xiang, Yuyaocen and Yu, Zhenping and Hao, Hongwei and Tang, Wei and Li, Yanwen and Liu, Houde and others},
  journal={Advanced Intelligent Systems},
  volume={6},
  number={10},
  pages={2400112},
  year={2024},
  publisher={Wiley Online Library}
}

@inproceedings{li2022towards,
  title={Towards adaptive continuous control of soft robotic manipulator using reinforcement learning},
  author={Li, Yingqi and Wang, Xiaomei and Kwok, Ka-Wai},
  booktitle={2022 IEEE/RSJ International Conference on Intelligent Robots and Systems (IROS)},
  pages={7074--7081},
  year={2022},
  organization={IEEE}
}

@article{meng2025sensor,
  title={Sensor-Space Based Robust Kinematic Control of Redundant Soft Manipulator by Learning},
  author={Meng, Yinan and Qian, Kun and Yang, Jiong and Su, Renbo and Li, Zhenhong and Wang, Charlie CL},
  journal={arXiv preprint arXiv:2507.16842},
  year={2025}
}

@article{nazeer2023soft,
  title={Soft dagger: Sample-efficient imitation learning for control of soft robots},
  author={Nazeer, Muhammad Sunny and Laschi, Cecilia and Falotico, Egidio},
  journal={Sensors},
  volume={23},
  number={19},
  pages={8278},
  year={2023},
  publisher={MDPI}
}

@inproceedings{yoo2025kinesoft,
  title={KineSoft: Learning Proprioceptive Manipulation Policies with Soft Robot Hands},
  author={Yoo, Uksang and Francis, Jonathan and Oh, Jean and Ichnowski, Jeffrey},
  booktitle={Conference on Robot Learning},
  pages={633--651},
  year={2025},
  organization={PMLR}
}

@article{liu2022touchless,
  title={Touchless interactive teaching of soft robots through flexible bimodal sensory interfaces},
  author={Liu, Wenbo and Duo, Youning and Liu, Jiaqi and Yuan, Feiyang and Li, Lei and Li, Luchen and Wang, Gang and Chen, Bohan and Wang, Siqi and Yang, Hui and others},
  journal={Nature communications},
  volume={13},
  number={1},
  pages={5030},
  year={2022},
  publisher={Nature Publishing Group UK London}
}

@article{sanchez2018robotic,
  title={Robotic manipulation and sensing of deformable objects in domestic and industrial applications: a survey},
  author={Sanchez, Jose and Corrales, Juan-Antonio and Bouzgarrou, Belhassen-Chedli and Mezouar, Youcef},
  journal={The International Journal of Robotics Research},
  volume={37},
  number={7},
  pages={688--716},
  year={2018},
  publisher={SAGE Publications Sage UK: London, England}
}

@article{gu2023survey,
  title={A survey on robotic manipulation of deformable objects: Recent advances, open challenges and new frontiers},
  author={Gu, Feida and Zhou, Yanmin and Wang, Zhipeng and Jiang, Shuo and He, Bin},
  journal={arXiv preprint arXiv:2312.10419},
  year={2023}
}

@article{saha2007manipulation,
  title={Manipulation planning for deformable linear objects},
  author={Saha, Mitul and Isto, Pekka},
  journal={IEEE Transactions on Robotics},
  volume={23},
  number={6},
  pages={1141--1150},
  year={2007},
  publisher={IEEE}
}

@article{luque2024model,
  title={Model predictive control for dynamic cloth manipulation: Parameter learning and experimental validation},
  author={Luque, Adri{\`a} and Parent, David and Colom{\'e}, Adri{\`a} and Ocampo-Martinez, Carlos and Torras, Carme},
  journal={IEEE Transactions on Control Systems Technology},
  volume={32},
  number={4},
  pages={1254--1270},
  year={2024},
  publisher={IEEE}
}

@inproceedings{matas2018sim,
  title={Sim-to-real reinforcement learning for deformable object manipulation},
  author={Matas, Jan and James, Stephen and Davison, Andrew J},
  booktitle={Conference on Robot Learning},
  pages={734--743},
  year={2018},
  organization={PMLR}
}

@article{hu20193,
  title={3-D deformable object manipulation using deep neural networks},
  author={Hu, Zhe and Han, Tao and Sun, Peigen and Pan, Jia and Manocha, Dinesh},
  journal={IEEE Robotics and Automation Letters},
  volume={4},
  number={4},
  pages={4255--4261},
  year={2019},
  publisher={IEEE}
}

@inproceedings{li2023dexdeform,
  title={DexDeform: Dexterous Deformable Object Manipulation with Human Demonstrations and Differentiable Physics},
  author={Li, Sizhe and Huang, Zhiao and Chen, Tao and Du, Tao and Su, Hao and Tenenbaum, Joshua B and Gan, Chuang},
  booktitle={The Eleventh International Conference on Learning Representations},
  year={2023}
}

@article{salhotra2022learning,
  title={Learning deformable object manipulation from expert demonstrations},
  author={Salhotra, Gautam and Liu, I-Chun Arthur and Dominguez-Kuhne, Marcus and Sukhatme, Gaurav S},
  journal={IEEE Robotics and Automation Letters},
  volume={7},
  number={4},
  pages={8775--8782},
  year={2022},
  publisher={IEEE}
}

@inproceedings{wu2023learning,
  title={Learning foresightful dense visual affordance for deformable object manipulation},
  author={Wu, Ruihai and Ning, Chuanruo and Dong, Hao},
  booktitle={Proceedings of the IEEE/CVF International Conference on Computer Vision},
  pages={10947--10956},
  year={2023}
}

@inproceedings{luan2025learning,
  title={Learning Efficient Robotic Garment Manipulation with Standardization},
  author={Luan, Feng and Meng, Shaoqiang and Wang, Zhipeng and Dong, Yanchao and Zhou, Yanmin and He, Bin and others},
  booktitle={Forty-second International Conference on Machine Learning},
  year={2025}
}

@article{duisterhof2024deformgs,
  title={Deformgs: Scene flow in highly deformable scenes for deformable object manipulation},
  author={Duisterhof, Bardienus P and Mandi, Zhao and Yao, Yunchao and Liu, Jia-Wei and Seidenschwarz, Jenny and Shou, Mike Zheng and Ramanan, Deva and Song, Shuran and Birchfield, Stan and Wen, Bowen and others},
  journal={Workshop on The Algorithmic Foundations OF Robotics},
  year={2024}
}

@article{weng2024interactive,
  title={Interactive perception for deformable object manipulation},
  author={Weng, Zehang and Zhou, Peng and Yin, Hang and Kravberg, Alexander and Varava, Anastasiia and Navarro-Alarcon, David and Kragic, Danica},
  journal={IEEE Robotics and Automation Letters},
  year={2024},
  publisher={IEEE}
}

@inproceedings{deng2024learning,
  title={Learning language-conditioned deformable object manipulation with graph dynamics},
  author={Deng, Yuhong and Mo, Kai and Xia, Chongkun and Wang, Xueqian},
  booktitle={2024 IEEE International Conference on Robotics and Automation (ICRA)},
  pages={7508--7514},
  year={2024},
  organization={IEEE}
}

@inproceedings{kuroki2024gendom,
  title={Gendom: Generalizable one-shot deformable object manipulation with parameter-aware policy},
  author={Kuroki, So and Guo, Jiaxian and Matsushima, Tatsuya and Okubo, Takuya and Kobayashi, Masato and Ikeda, Yuya and Takanami, Ryosuke and Yoo, Paul and Matsuo, Yutaka and Iwasawa, Yusuke},
  booktitle={2024 IEEE International Conference on Robotics and Automation (ICRA)},
  pages={14792--14799},
  year={2024},
  organization={IEEE}
}

@inproceedings{li2024deformnet,
  title={Deformnet: Latent space modeling and dynamics prediction for deformable object manipulation},
  author={Li, Chenchang and Ai, Zihao and Wu, Tong and Li, Xiaosa and Ding, Wenbo and Xu, Huazhe},
  booktitle={2024 IEEE International Conference on Robotics and Automation (ICRA)},
  pages={14770--14776},
  year={2024},
  organization={IEEE}
}

@inproceedings{thach2024defgoalnet,
  title={DefGoalNet: Contextual goal learning from demonstrations for deformable object manipulation},
  author={Thach, Bao and Watts, Tanner and Ho, Shing-Hei and Hermans, Tucker and Kuntz, Alan},
  booktitle={2024 IEEE International Conference on Robotics and Automation (ICRA)},
  pages={3145--3152},
  year={2024},
  organization={IEEE}
}

@inproceedings{bauer2024doughnet,
  title={Doughnet: A visual predictive model for topological manipulation of deformable objects},
  author={Bauer, Dominik and Xu, Zhenjia and Song, Shuran},
  booktitle={European Conference on Computer Vision},
  pages={92--108},
  year={2024},
  organization={Springer}
}

@article{scheikl2024movement,
  title={Movement primitive diffusion: Learning gentle robotic manipulation of deformable objects},
  author={Scheikl, Paul Maria and Schreiber, Nicolas and Haas, Christoph and Freymuth, Niklas and Neumann, Gerhard and Lioutikov, Rudolf and Mathis-Ullrich, Franziska},
  journal={IEEE Robotics and Automation Letters},
  volume={9},
  number={6},
  pages={5338--5345},
  year={2024},
  publisher={IEEE}
}

@article{caporali2024deformable,
  title={Deformable linear objects manipulation with online model parameters estimation},
  author={Caporali, Alessio and Kicki, Piotr and Galassi, Kevin and Zanella, Riccardo and Walas, Krzysztof and Palli, Gianluca},
  journal={IEEE Robotics and Automation Letters},
  volume={9},
  number={3},
  pages={2598--2605},
  year={2024},
  publisher={IEEE}
}

@inproceedings{shi2023robocook,
  title={RoboCook: Long-Horizon Elasto-Plastic Object Manipulation with Diverse Tools},
  author={Shi, Haochen and Xu, Huazhe and Clarke, Samuel and Li, Yunzhu and Wu, Jiajun},
  booktitle={Conference on Robot Learning},
  pages={642--660},
  year={2023},
  organization={PMLR}
}

@inproceedings{viswanath2023handloom,
  title={Handloom: Learned tracing of one-dimensional objects for inspection and manipulation},
  author={Viswanath, Vainavi and Shivakumar, Kaushik and Parulekar, Mallika and Ajmera, Jainil and Kerr, Justin and Ichnowski, Jeffrey and Cheng, Richard and Kollar, Thomas and Goldberg, Ken},
  booktitle={Conference on Robot Learning},
  pages={341--357},
  year={2023},
  organization={PMLR}
}

@article{holmberg2000development,
  title={Development and control of a holonomic mobile robot for mobile manipulation tasks},
  author={Holmberg, Robert and Khatib, Oussama},
  journal={The International Journal of Robotics Research},
  volume={19},
  number={11},
  pages={1066--1074},
  year={2000},
  publisher={SAGE Publications}
}

@article{hebert2015mobile,
  title={Mobile manipulation and mobility as manipulation—Design and algorithms of RoboSimian},
  author={Hebert, Paul and Bajracharya, Max and Ma, Jeremy and Hudson, Nicolas and Aydemir, Alper and Reid, Jason and Bergh, Charles and Borders, James and Frost, Matthew and Hagman, Michael and others},
  journal={Journal of Field Robotics},
  volume={32},
  number={2},
  pages={255--274},
  year={2015},
  publisher={Wiley Online Library}
}

@inproceedings{berenson2008optimization,
  title={An optimization approach to planning for mobile manipulation},
  author={Berenson, Dmitry and Kuffner, James and Choset, Howie},
  booktitle={2008 IEEE International Conference on Robotics and Automation},
  pages={1187--1192},
  year={2008},
  organization={IEEE}
}

@inproceedings{rana2023sayplan,
  title={SayPlan: Grounding Large Language Models using 3D Scene Graphs for Scalable Robot Task Planning},
  author={Rana, Krishan and Haviland, Jesse and Garg, Sourav and Abou-Chakra, Jad and Reid, Ian and Suenderhauf, Niko},
  booktitle={Conference on Robot Learning},
  pages={23--72},
  year={2023},
  organization={PMLR}
}

@inproceedings{hou2025tamma,
  title={TaMMa: Target-driven Multi-subscene Mobile Manipulation},
  author={Hou, Jiawei and Wang, Tianyu and Pan, Tongying and Wang, Shouyan and Xue, Xiangyang and Fu, Yanwei},
  booktitle={Conference on Robot Learning},
  pages={4119--4140},
  year={2025},
  organization={PMLR}
}

@inproceedings{xiao2025robi,
  title={Robi Butler: Multimodal Remote Interaction with a Household Robot Assistant},
  author={Xiao, Anxing and Janaka, Nuwan and Hu, Tianrun and Gupta, Anshul and Li, Kaixin and Yu, Cunjun and Hsu, David},
  booktitle={2025 IEEE International Conference on Robotics and Automation},
  year={2025}
}

@article{sundaresan2025homer,
  title={HoMeR: Learning In-the-Wild Mobile Manipulation via Hybrid Imitation and Whole-Body Control},
  author={Sundaresan, Priya and Malhotra, Rhea and Miao, Phillip and Yang, Jingyun and Wu, Jimmy and Hu, Hengyuan and Antonova, Rika and Engelmann, Francis and Sadigh, Dorsa and Bohg, Jeannette},
  journal={arXiv preprint arXiv:2506.01185},
  year={2025}
}

@article{zhicheng2025object,
  title={Object-Centric Mobile Manipulation through SAM2-Guided Perception and Imitation Learning},
  author={Zhicheng, Wang and Yagi, Satoshi and Yamamori, Satoshi and Morimoto, Jun},
  journal={arXiv preprint arXiv:2507.10899},
  year={2025}
}

@inproceedings{ciocarlie2012mobile,
  title={Mobile manipulation through an assistive home robot},
  author={Ciocarlie, Matei and Hsiao, Kaijen and Leeper, Adam and Gossow, David},
  booktitle={2012 IEEE/RSJ International Conference on Intelligent Robots and Systems},
  pages={5313--5320},
  year={2012},
  organization={IEEE}
}

@article{chitta2012mobile,
  title={Mobile manipulation in unstructured environments: Perception, planning, and execution},
  author={Chitta, Sachin and Jones, E Gil and Ciocarlie, Matei and Hsiao, Kaijen},
  journal={IEEE Robotics \& Automation Magazine},
  volume={19},
  number={2},
  pages={58--71},
  year={2012},
  publisher={IEEE}
}

@article{wang2020learning,
  title={Learning mobile manipulation through deep reinforcement learning},
  author={Wang, Cong and Zhang, Qifeng and Tian, Qiyan and Li, Shuo and Wang, Xiaohui and Lane, David and Petillot, Yvan and Wang, Sen},
  journal={Sensors},
  volume={20},
  number={3},
  pages={939},
  year={2020},
  publisher={MDPI}
}

@article{haviland2022holistic,
  title={A holistic approach to reactive mobile manipulation},
  author={Haviland, Jesse and S{\"u}nderhauf, Niko and Corke, Peter},
  journal={IEEE Robotics and Automation Letters},
  volume={7},
  number={2},
  pages={3122--3129},
  year={2022},
  publisher={IEEE}
}

@inproceedings{burgess2023architecture,
  title={An architecture for reactive mobile manipulation on-the-move},
  author={Burgess-Limerick, Ben and Lehnert, Chris and Leitner, Jurgen and Corke, Peter},
  booktitle={IEEE International Conference on Robotics and Automation 2023},
  pages={1623--1629},
  year={2023},
  organization={IEEE, Institute of Electrical and Electronics Engineers}
}

@article{xiong2024adaptive,
  title={Adaptive mobile manipulation for articulated objects in the open world},
  author={Xiong, Haoyu and Mendonca, Russell and Shaw, Kenneth and Pathak, Deepak},
  journal={arXiv preprint arXiv:2401.14403},
  year={2024}
}

@inproceedings{wu2020spatial,
  title={Spatial action maps for mobile manipulation},
  author={Wu, Jimmy and Sun, Xingyuan and Zeng, Andy and Song, Shuran and Lee, Johnny and Rusinkiewicz, Szymon and Funkhouser, Thomas},
  booktitle={Robotics: Science and Systems},
  year={2020}
}

@inproceedings{yang2024harmonic,
  title={Harmonic mobile manipulation},
  author={Yang, Ruihan and Kim, Yejin and Hendrix, Rose and Kembhavi, Aniruddha and Wang, Xiaolong and Ehsani, Kiana},
  booktitle={2024 IEEE/RSJ International Conference on Intelligent Robots and Systems (IROS)},
  pages={3658--3665},
  year={2024},
  organization={IEEE}
}

@article{pankert2020perceptive,
  title={Perceptive model predictive control for continuous mobile manipulation},
  author={Pankert, Johannes and Hutter, Marco},
  journal={IEEE Robotics and Automation Letters},
  volume={5},
  number={4},
  pages={6177--6184},
  year={2020},
  publisher={IEEE}
}

@article{honerkamp2021learning,
  title={Learning kinematic feasibility for mobile manipulation through deep reinforcement learning},
  author={Honerkamp, Daniel and Welschehold, Tim and Valada, Abhinav},
  journal={IEEE Robotics and Automation Letters},
  volume={6},
  number={4},
  pages={6289--6296},
  year={2021},
  publisher={IEEE}
}

@inproceedings{wu2025momanipvla,
  title={Momanipvla: Transferring vision-language-action models for general mobile manipulation},
  author={Wu, Zhenyu and Zhou, Yuheng and Xu, Xiuwei and Wang, Ziwei and Yan, Haibin},
  booktitle={Proceedings of the Computer Vision and Pattern Recognition Conference},
  pages={1714--1723},
  year={2025}
}

@inproceedings{huang2023skill,
  title={Skill transformer: A monolithic policy for mobile manipulation},
  author={Huang, Xiaoyu and Batra, Dhruv and Rai, Akshara and Szot, Andrew},
  booktitle={Proceedings of the IEEE/CVF International Conference on Computer Vision},
  pages={10852--10862},
  year={2023}
}

@inproceedings{jauhri2024active,
  title={Active-perceptive motion generation for mobile manipulation},
  author={Jauhri, Snehal and Lueth, Sophie and Chalvatzaki, Georgia},
  booktitle={2024 IEEE International Conference on Robotics and Automation (ICRA)},
  pages={1413--1419},
  year={2024},
  organization={IEEE}
}

@article{honerkamp2024language,
  title={Language-grounded dynamic scene graphs for interactive object search with mobile manipulation},
  author={Honerkamp, Daniel and B{\"u}chner, Martin and Despinoy, Fabien and Welschehold, Tim and Valada, Abhinav},
  journal={IEEE Robotics and Automation Letters},
  year={2024},
  publisher={IEEE}
}

@inproceedings{hu2023causal,
  title={Causal policy gradient for whole-body mobile manipulation},
  author={Hu, Jiaheng and Stone, Peter and Mart{\'\i}n-Mart{\'\i}n, Roberto},
  booktitle={Robotics: Science and Systems},
  year={2023}
}

@inproceedings{yang2023moma,
  title={Moma-force: Visual-force imitation for real-world mobile manipulation},
  author={Yang, Taozheng and Jing, Ya and Wu, Hongtao and Xu, Jiafeng and Sima, Kuankuan and Chen, Guangzeng and Sima, Qie and Kong, Tao},
  booktitle={2023 IEEE/RSJ International Conference on Intelligent Robots and Systems (IROS)},
  pages={6847--6852},
  year={2023},
  organization={IEEE}
}

@inproceedings{wu2025tidybot++,
  title={TidyBot++: An Open-Source Holonomic Mobile Manipulator for Robot Learning},
  author={Wu, Jimmy and Chong, William and Holmberg, Robert and Prasad, Aaditya and Gao, Yihuai and Khatib, Oussama and Song, Shuran and Rusinkiewicz, Szymon and Bohg, Jeannette},
  booktitle={Conference on Robot Learning},
  pages={3729--3741},
  year={2025},
  organization={PMLR}
}

@article{jeon2023learning,
  title={Learning whole-body manipulation for quadrupedal robot},
  author={Jeon, Seunghun and Jung, Moonkyu and Choi, Suyoung and Kim, Beomjoon and Hwangbo, Jemin},
  journal={IEEE Robotics and Automation Letters},
  volume={9},
  number={1},
  pages={699--706},
  year={2023},
  publisher={IEEE}
}

@inproceedings{zhi2025learning,
  title={Learning Unified Force and Position Control for Legged Loco-Manipulation},
  author={Zhi, Peiyuan and Li, Peiyang and Yin, Jianqin and Jia, Baoxiong and Huang, Siyuan},
  booktitle={Conference on Robot Learning},
  year={2025}
}

@inproceedings{niu2025human2locoman,
  title={Human2LocoMan: Learning Versatile Quadrupedal Manipulation with Human Pretraining},
  author={Niu, Yaru and Zhang, Yunzhe and Yu, Mingyang and Lin, Changyi and Li, Chenhao and Wang, Yikai and Yang, Yuxiang and Yu, Wenhao and Zhang, Tingnan and Chen, Bingqing and others},
  booktitle={Robotics: Science and Systems},
  year={2025}
}

@inproceedings{hou2025efficient,
  title={Efficient learning of a unified policy for whole-body manipulation and locomotion skills},
  author={Hou, Dianyong and Zhu, Chengrui and Zhang, Zhen and Li, Zhibin and Guo, Chuang and Liu, Yong},
  booktitle={2025 IEEE/RSJ International Conference on Intelligent Robots and Systems (IROS)},
  pages={5455--5461},
  year={2025},
  organization={IEEE}
}

@article{arcari2023bayesian,
  title={Bayesian multi-task learning mpc for robotic mobile manipulation},
  author={Arcari, Elena and Minniti, Maria Vittoria and Scampicchio, Anna and Carron, Andrea and Farshidian, Farbod and Hutter, Marco and Zeilinger, Melanie N},
  journal={IEEE Robotics and Automation Letters},
  volume={8},
  number={6},
  pages={3222--3229},
  year={2023},
  publisher={IEEE}
}

@inproceedings{fu2023deep,
  title={Deep whole-body control: learning a unified policy for manipulation and locomotion},
  author={Fu, Zipeng and Cheng, Xuxin and Pathak, Deepak},
  booktitle={Conference on Robot Learning},
  pages={138--149},
  year={2023},
  organization={PMLR}
}

@article{yokoyama2023asc,
  title={Asc: Adaptive skill coordination for robotic mobile manipulation},
  author={Yokoyama, Naoki and Clegg, Alex and Truong, Joanne and Undersander, Eric and Yang, Tsung-Yen and Arnaud, Sergio and Ha, Sehoon and Batra, Dhruv and Rai, Akshara},
  journal={IEEE Robotics and Automation Letters},
  volume={9},
  number={1},
  pages={779--786},
  year={2023},
  publisher={IEEE}
}

@article{pan2025roboduet,
  title={RoboDuet: Learning a Cooperative Policy for Whole-body Legged Loco-Manipulation},
  author={Pan, Guoping and Ben, Qingwei and Yuan, Zhecheng and Jiang, Guangqi and Ji, Yandong and Li, Shoujie and Pang, Jiangmiao and Liu, Houde and Xu, Huazhe},
  journal={IEEE Robotics and Automation Letters},
  year={2025},
  publisher={IEEE}
}

@inproceedings{mendonca2025continuously,
  title={Continuously Improving Mobile Manipulation with Autonomous Real-World RL},
  author={Mendonca, Russell and Panov, Emmanuel and Bucher, Bernadette and Wang, Jiuguang and Pathak, Deepak},
  booktitle={Conference on Robot Learning},
  pages={5204--5219},
  year={2025},
  organization={PMLR}
}

@inproceedings{zhang2024gamma,
  title={Gamma: Graspability-aware mobile manipulation policy learning based on online grasping pose fusion},
  author={Zhang, Jiazhao and Gireesh, Nandiraju and Wang, Jilong and Fang, Xiaomeng and Xu, Chaoyi and Chen, Weiguang and Dai, Liu and Wang, He},
  booktitle={2024 IEEE International Conference on Robotics and Automation (ICRA)},
  pages={1399--1405},
  year={2024},
  organization={IEEE}
}

@inproceedings{he2024learning,
  title={Learning visual quadrupedal loco-manipulation from demonstrations},
  author={He, Zhengmao and Lei, Kun and Ze, Yanjie and Sreenath, Koushil and Li, Zhongyu and Xu, Huazhe},
  booktitle={2024 IEEE/RSJ International Conference on Intelligent Robots and Systems (IROS)},
  pages={9102--9109},
  year={2024},
  organization={IEEE}
}

@inproceedings{cheng2023legs,
  title={Legs as Manipulator: Pushing Quadrupedal Agility Beyond Locomotion},
  author={Cheng, Xuxin and Kumar, Ashish and Pathak, Deepak},
  booktitle={2023 IEEE International Conference on Robotics and Automation (ICRA)},
  pages={5106--5112},
  year={2023},
  organization={IEEE}
}

@article{wang2025quadwbg,
  title={Quadwbg: Generalizable quadrupedal whole-body grasping},
  author={Wang, Jilong and Rajabov, Javokhirbek and Xu, Chaoyi and Zheng, Yiming and Wang, He},
  journal={2025 IEEE International Conference on Robotics and Automation (ICRA)},
  year={2025}
}

@article{jiang2024learning,
  title={Learning Whole-Body Loco-Manipulation for Omni-Directional Task Space Pose Tracking with a Wheeled-Quadrupedal-Manipulator},
  author={Jiang, Kaiwen and Fu, Zhen and Guo, Junde and Zhang, Wei and Chen, Hua},
  journal={IEEE Robotics and Automation Letters},
  year={2024},
  publisher={IEEE}
}

@inproceedings{huang2024hilma,
  title={Hilma-res: A general hierarchical framework via residual rl for combining quadrupedal locomotion and manipulation},
  author={Huang, Xiaoyu and Liao, Qiayuan and Ni, Yiming and Li, Zhongyu and Smith, Laura and Levine, Sergey and Peng, Xue Bin and Sreenath, Koushil},
  booktitle={2024 IEEE/RSJ International Conference on Intelligent Robots and Systems (IROS)},
  pages={9050--9057},
  year={2024},
  organization={IEEE}
}

@inproceedings{wolfslag2020optimisation,
  title={Optimisation of body-ground contact for augmenting the whole-body loco-manipulation of quadruped robots},
  author={Wolfslag, Wouter J and McGreavy, Christopher and Xin, Guiyang and Tiseo, Carlo and Vijayakumar, Sethu and Li, Zhibin},
  booktitle={2020 IEEE/RSJ International Conference on Intelligent Robots and Systems (IROS)},
  pages={3694--3701},
  year={2020},
  organization={IEEE}
}

@inproceedings{arm2024pedipulate,
  title={Pedipulate: Enabling manipulation skills using a quadruped robot’s leg},
  author={Arm, Philip and Mittal, Mayank and Kolvenbach, Hendrik and Hutter, Marco},
  booktitle={2024 IEEE International Conference on Robotics and Automation (ICRA)},
  pages={5717--5723},
  year={2024},
  organization={IEEE}
}

@article{ferrolho2023roloma,
  title={Roloma: Robust loco-manipulation for quadruped robots with arms},
  author={Ferrolho, Henrique and Ivan, Vladimir and Merkt, Wolfgang and Havoutis, Ioannis and Vijayakumar, Sethu},
  journal={Autonomous Robots},
  volume={47},
  number={8},
  pages={1463--1481},
  year={2023},
  publisher={Springer}
}

@inproceedings{lin2024locoman,
  title={Locoman: Advancing versatile quadrupedal dexterity with lightweight loco-manipulators},
  author={Lin, Changyi and Liu, Xingyu and Yang, Yuxiang and Niu, Yaru and Yu, Wenhao and Zhang, Tingnan and Tan, Jie and Boots, Byron and Zhao, Ding},
  booktitle={2024 IEEE/RSJ International Conference on Intelligent Robots and Systems (IROS)},
  pages={6877--6884},
  year={2024},
  organization={IEEE}
}

@inproceedings{song2024germ,
  title={Germ: A generalist robotic model with mixture-of-experts for quadruped robot},
  author={Song, Wenxuan and Zhao, Han and Ding, Pengxiang and Cui, Can and Lyu, Shangke and Fan, Yaning and Wang, Donglin},
  booktitle={2024 IEEE/RSJ International Conference on Intelligent Robots and Systems (IROS)},
  pages={11879--11886},
  year={2024},
  organization={IEEE}
}

@inproceedings{ding2024quar,
  title={Quar-vla: Vision-language-action model for quadruped robots},
  author={Ding, Pengxiang and Zhao, Han and Zhang, Wenjie and Song, Wenxuan and Zhang, Min and Huang, Siteng and Yang, Ningxi and Wang, Donglin},
  booktitle={European Conference on Computer Vision},
  pages={352--367},
  year={2024},
  organization={Springer}
}

@inproceedings{zhao2025more,
  title={More: Unlocking scalability in reinforcement learning for quadruped vision-language-action models},
  author={Zhao, Han and Song, Wenxuan and Wang, Donglin and Tong, Xinyang and Ding, Pengxiang and Cheng, Xuelian and Ge, Zongyuan},
  booktitle={2025 IEEE international conference on robotics and automation (ICRA)},
  year={2025}
}

@inproceedings{liu2025visual,
  title={Visual Whole-Body Control for Legged Loco-Manipulation},
  author={Liu, Minghuan and Chen, Zixuan and Cheng, Xuxin and Ji, Yandong and Qiu, Ri-Zhao and Yang, Ruihan and Wang, Xiaolong},
  booktitle={Conference on Robot Learning},
  pages={234--257},
  year={2025},
  organization={PMLR}
}

@article{harada2006dynamics,
  title={Dynamics and balance of a humanoid robot during manipulation tasks},
  author={Harada, Kensuke and Kajita, Shuuji and Kaneko, Kenji and Hirukawa, Hirohisa},
  journal={IEEE Transactions on Robotics},
  volume={22},
  number={3},
  pages={568--575},
  year={2006},
  publisher={IEEE}
}

@article{bouyarmane2012humanoid,
  title={Humanoid robot locomotion and manipulation step planning},
  author={Bouyarmane, Karim and Kheddar, Abderrahmane},
  journal={Advanced Robotics},
  volume={26},
  number={10},
  pages={1099--1126},
  year={2012},
  publisher={Taylor \& Francis}
}

@inproceedings{seo2023deep,
  title={Deep imitation learning for humanoid loco-manipulation through human teleoperation},
  author={Seo, Mingyo and Han, Steve and Sim, Kyutae and Bang, Seung Hyeon and Gonzalez, Carlos and Sentis, Luis and Zhu, Yuke},
  booktitle={2023 IEEE-RAS 22nd International Conference on Humanoid Robots (Humanoids)},
  pages={1--8},
  year={2023},
  organization={IEEE}
}

@inproceedings{he2025omnih2o,
  title={OmniH2O: Universal and Dexterous Human-to-Humanoid Whole-Body Teleoperation and Learning},
  author={He, Tairan and Luo, Zhengyi and He, Xialin and Xiao, Wenli and Zhang, Chong and Zhang, Weinan and Kitani, Kris M and Liu, Changliu and Shi, Guanya},
  booktitle={Conference on Robot Learning},
  pages={1516--1540},
  year={2025},
  organization={PMLR}
}

@inproceedings{li2025okami,
  title={OKAMI: Teaching Humanoid Robots Manipulation Skills through Single Video Imitation},
  author={Li, Jinhan and Zhu, Yifeng and Xie, Yuqi and Jiang, Zhenyu and Seo, Mingyo and Pavlakos, Georgios and Zhu, Yuke},
  booktitle={Conference on Robot Learning},
  pages={299--317},
  year={2025},
  organization={PMLR}
}

@article{ze2025generalizable,
  title={Generalizable humanoid manipulation with 3d diffusion policies},
  author={Ze, Yanjie and Chen, Zixuan and Wang, Wenhao and Chen, Tianyi and He, Xialin and Yuan, Ying and Peng, Xue Bin and Wu, Jiajun},
  journal={2025 IEEE/RSJ International Conference on Intelligent Robots and Systems (IROS)},
  year={2025}
}

@inproceedings{fu2025humanplus,
  title={HumanPlus: Humanoid Shadowing and Imitation from Humans},
  author={Fu, Zipeng and Zhao, Qingqing and Wu, Qi and Wetzstein, Gordon and Finn, Chelsea},
  booktitle={Conference on Robot Learning},
  pages={2828--2844},
  year={2025},
  organization={PMLR}
}

@article{ding2025humanoid,
  title={Humanoid-vla: Towards universal humanoid control with visual integration},
  author={Ding, Pengxiang and Ma, Jianfei and Tong, Xinyang and Zou, Binghong and Luo, Xinxin and Fan, Yiguo and Wang, Ting and Lu, Hongchao and Mo, Panzhong and Liu, Jinxin and others},
  journal={arXiv preprint arXiv:2502.14795},
  year={2025}
}

@inproceedings{qiu2025humanoid,
  title={Humanoid Policy Human Policy},
  author={Qiu, Ri-Zhao and Yang, Shiqi and Cheng, Xuxin and Chawla, Chaitanya and Li, Jialong and He, Tairan and Yan, Ge and Yoon, David J and Hoque, Ryan and Paulsen, Lars and others},
  booktitle={Conference on Robot Learning},
  pages={2888--2906},
  year={2025},
  organization={PMLR}
}

@article{murooka2025tact,
  title={TACT: Humanoid Whole-body Contact Manipulation through Deep Imitation Learning with Tactile Modality},
  author={Murooka, Masaki and Hoshi, Takahiro and Fukumitsu, Kensuke and Masuda, Shimpei and Hamze, Marwan and Sasaki, Tomoya and Morisawa, Mitsuharu and Yoshida, Eiichi},
  journal={IEEE Robotics and Automation Letters},
  year={2025},
  publisher={IEEE}
}

@article{schakkal2025hierarchical,
  title={Hierarchical Vision-Language Planning for Multi-Step Humanoid Manipulation},
  author={Schakkal, Andr{\'e} and Zandonati, Ben and Yang, Zhutian and Azizan, Navid},
  journal={arXiv preprint arXiv:2506.22827},
  year={2025}
}

@article{zhang2025flam,
  title={Flam: Foundation model-based body stabilization for humanoid locomotion and manipulation},
  author={Zhang, Xianqi and Wei, Hongliang and Wang, Wenrui and Wang, Xingtao and Fan, Xiaopeng and Zhao, Debin},
  journal={arXiv preprint arXiv:2503.22249},
  year={2025}
}

@article{bjorck2025gr00t,
  title={Gr00t n1: An open foundation model for generalist humanoid robots},
  author={Bjorck, Johan and Casta{\~n}eda, Fernando and Cherniadev, Nikita and Da, Xingye and Ding, Runyu and Fan, Linxi and Fang, Yu and Fox, Dieter and Hu, Fengyuan and Huang, Spencer and others},
  journal={arXiv preprint arXiv:2503.14734},
  year={2025}
}

@article{xu2024humanvla,
  title={Humanvla: Towards vision-language directed object rearrangement by physical humanoid},
  author={Xu, Xinyu and Zhang, Yizheng and Li, Yong-Lu and Han, Lei and Lu, Cewu},
  journal={Advances in Neural Information Processing Systems},
  volume={37},
  pages={18633--18659},
  year={2024}
}

@article{gao2024coohoi,
  title={Coohoi: Learning cooperative human-object interaction with manipulated object dynamics},
  author={Gao, Jiawei and Wang, Ziqin and Xiao, Zeqi and Wang, Jingbo and Wang, Tai and Cao, Jinkun and Hu, Xiaolin and Liu, Si and Dai, Jifeng and Pang, Jiangmiao},
  journal={Advances in Neural Information Processing Systems},
  volume={37},
  pages={79741--79763},
  year={2024}
}

@article{xie2023hierarchical,
  title={Hierarchical planning and control for box loco-manipulation},
  author={Xie, Zhaoming and Tseng, Jonathan and Starke, Sebastian and van de Panne, Michiel and Liu, C Karen},
  journal={Proceedings of the ACM on Computer Graphics and Interactive Techniques},
  volume={6},
  number={3},
  pages={1--18},
  year={2023},
  publisher={ACM New York, NY, USA}
}

@inproceedings{lin2025sim,
  title={Sim-to-Real Reinforcement Learning for Vision-Based Dexterous Manipulation on Humanoids},
  author={Lin, Toru and Sachdev, Kartik and Fan, Linxi and Malik, Jitendra and Zhu, Yuke},
  booktitle={Conference on Robot Learning},
  pages={4926--4940},
  year={2025},
  organization={PMLR}
}

@article{gao2025towards,
  title={Towards Human-level Intelligence via Human-like Whole-Body Manipulation},
  author={Gao, Guang and Wang, Jianan and Zuo, Jinbo and Jiang, Junnan and Zhang, Jingfan and Zeng, Xianwen and Zhu, Yuejiang and Ma, Lianyang and Chen, Ke and Sheng, Minhua and others},
  journal={arXiv preprint arXiv:2507.17141},
  year={2025}
}


%% file: refs/5-3d_representation.bib
@inproceedings{simeonov2022neural,
  title={Neural descriptor fields: Se (3)-equivariant object representations for manipulation},
  author={Simeonov, Anthony and Du, Yilun and Tagliasacchi, Andrea and Tenenbaum, Joshua B and Rodriguez, Alberto and Agrawal, Pulkit and Sitzmann, Vincent},
  booktitle={2022 International Conference on Robotics and Automation (ICRA)},
  pages={6394--6400},
  year={2022},
  organization={IEEE}
}

@inproceedings{simeonov2023se,
  title={Se (3)-equivariant relational rearrangement with neural descriptor fields},
  author={Simeonov, Anthony and Du, Yilun and Lin, Yen-Chen and Garcia, Alberto Rodriguez and Kaelbling, Leslie Pack and Lozano-P{\'e}rez, Tom{\'a}s and Agrawal, Pulkit},
  booktitle={Conference on Robot Learning},
  pages={835--846},
  year={2023},
  organization={PMLR}
}

@inproceedings{shen2023distilled,
  title={Distilled Feature Fields Enable Few-Shot Language-Guided Manipulation},
  author={Shen, William and Yang, Ge and Yu, Alan and Wong, Jansen and Kaelbling, Leslie Pack and Isola, Phillip},
  booktitle={Conference on Robot Learning},
  pages={405--424},
  year={2023},
  organization={PMLR}
}

@inproceedings{wang2025d,
  title={D$^3$Fields: Dynamic 3D Descriptor Fields for Zero-Shot Generalizable Rearrangement},
  author={Wang, Yixuan and Zhang, Mingtong and Li, Zhuoran and Kelestemur, Tarik and Driggs-Campbell, Katherine Rose and Wu, Jiajun and Fei-Fei, Li and Li, Yunzhu},
  booktitle={Conference on Robot Learning},
  pages={272--298},
  year={2025},
  organization={PMLR}
}

@inproceedings{jiang2025roboexp,
  title={RoboEXP: Action-Conditioned Scene Graph via Interactive Exploration for Robotic Manipulation},
  author={Jiang, Hanxiao and Huang, Binghao and Wu, Ruihai and Li, Zhuoran and Garg, Shubham and Nayyeri, Hooshang and Wang, Shenlong and Li, Yunzhu},
  booktitle={Conference on Robot Learning},
  pages={3027--3052},
  year={2025},
  organization={PMLR}
}

@inproceedings{abou2025physically,
  title={Physically Embodied Gaussian Splatting: A Visually Learnt and Physically Grounded 3D Representation for Robotics},
  author={Abou-Chakra, Jad and Rana, Krishan and Dayoub, Feras and Suenderhauf, Niko},
  booktitle={Conference on Robot Learning},
  pages={513--530},
  year={2025},
  organization={PMLR}
}

@inproceedings{huang2025imagination,
  title={IMAGINATION POLICY: Using Generative Point Cloud Models for Learning Manipulation Policies},
  author={Huang, Haojie and Schmeckpeper, Karl and Wang, Dian and Biza, Ondrej and Qian, Yaoyao and Liu, Haotian and Jia, Mingxi and Platt, Robert and Walters, Robin},
  booktitle={Conference on Robot Learning},
  pages={5150--5165},
  year={2025},
  organization={PMLR}
}

@inproceedings{shorinwa2025splat,
  title={Splat-MOVER: Multi-Stage, Open-Vocabulary Robotic Manipulation via Editable Gaussian Splatting},
  author={Shorinwa, Olaolu and Tucker, Johnathan and Smith, Aliyah and Swann, Aiden and Chen, Timothy and Firoozi, Roya and Kennedy, Monroe David and Schwager, Mac},
  booktitle={Conference on Robot Learning},
  pages={4748--4770},
  year={2025},
  organization={PMLR}
}

@article{sheng2024msgfield,
  title={MSGField: A Unified Scene Representation Integrating Motion, Semantics, and Geometry for Robotic Manipulation},
  author={Sheng, Yu and Lin, Runfeng and Wang, Lidian and Qiu, Quecheng and Zhang, YanYong and Zhang, Yu and Hua, Bei and Ji, Jianmin},
  journal={arXiv preprint arXiv:2410.15730},
  year={2024}
}

@inproceedings{yang2025novel,
  title={Novel demonstration generation with gaussian splatting enables robust one-shot manipulation},
  author={Yang, Sizhe and Yu, Wenye and Zeng, Jia and Lv, Jun and Ren, Kerui and Lu, Cewu and Lin, Dahua and Pang, Jiangmiao},
  booktitle={Robotics: Science and Systems},
  year={2025}
}

@inproceedings{li2024object,
  title={Object-aware gaussian splatting for robotic manipulation},
  author={Li, Yulong and Pathak, Deepak},
  booktitle={ICRA 2024 Workshop on 3D Visual Representations for Robot Manipulation},
  year={2024}
}


%% file: refs/5-affordance.bib
@incollection{gibson2014theory,
  title={The theory of affordances:(1979)},
  author={Gibson, James J},
  booktitle={The people, place, and space reader},
  pages={56--60},
  year={2014},
  publisher={Routledge}
}

@inproceedings{jiang2022ditto,
  title={Ditto: Building digital twins of articulated objects from interaction},
  author={Jiang, Zhenyu and Hsu, Cheng-Chun and Zhu, Yuke},
  booktitle={Proceedings of the IEEE/CVF Conference on Computer Vision and Pattern Recognition},
  pages={5616--5626},
  year={2022}
}

@inproceedings{geng2024sage,
  title={Sage: Bridging semantic and actionable parts for generalizable manipulation of articulated objects},
  author={Geng, Haoran and Wei, Songlin and Deng, Congyue and Shen, Bokui and Wang, He and Guibas, Leonidas},
  booktitle={Robotics: Science and Systems},
  year={2024}
}

@inproceedings{geng2023gapartnet,
  title={Gapartnet: Cross-category domain-generalizable object perception and manipulation via generalizable and actionable parts},
  author={Geng, Haoran and Xu, Helin and Zhao, Chengyang and Xu, Chao and Yi, Li and Huang, Siyuan and Wang, He},
  booktitle={Proceedings of the IEEE/CVF Conference on Computer Vision and Pattern Recognition},
  pages={7081--7091},
  year={2023}
}

@inproceedings{yin2025partinstruct,
  title={PartInstruct: Part-level Instruction Following for Fine-grained Robot Manipulation},
  author={Yin, Yifan and Han, Zhengtao and Aarya, Shivam and Wang, Jianxin and Xu, Shuhang and Peng, Jiawei and Wang, Angtian and Yuille, Alan and Shu, Tianmin},
  booktitle={Robotics: Science and Systems},
  year={2025}
}

@inproceedings{nguyen2024language,
  title={Language-conditioned affordance-pose detection in 3d point clouds},
  author={Nguyen, Toan and Vu, Minh Nhat and Huang, Baoru and Van Vo, Tuan and Truong, Vy and Le, Ngan and Vo, Thieu and Le, Bac and Nguyen, Anh},
  booktitle={2024 IEEE International Conference on Robotics and Automation (ICRA)},
  pages={3071--3078},
  year={2024},
  organization={IEEE}
}

@inproceedings{liu2023composable,
  title={Composable Part-Based Manipulation},
  author={Liu, Weiyu and Mao, Jiayuan and Hsu, Joy and Hermans, Tucker and Garg, Animesh and Wu, Jiajun},
  booktitle={Conference on Robot Learning},
  pages={1300--1315},
  year={2023},
  organization={PMLR}
}

@inproceedings{chen2024spatialvlm,
  title={Spatialvlm: Endowing vision-language models with spatial reasoning capabilities},
  author={Chen, Boyuan and Xu, Zhuo and Kirmani, Sean and Ichter, Brian and Sadigh, Dorsa and Guibas, Leonidas and Xia, Fei},
  booktitle={Proceedings of the IEEE/CVF Conference on Computer Vision and Pattern Recognition},
  pages={14455--14465},
  year={2024}
}

@inproceedings{yuan2025robopoint,
  title={RoboPoint: A Vision-Language Model for Spatial Affordance Prediction in Robotics},
  author={Yuan, Wentao and Duan, Jiafei and Blukis, Valts and Pumacay, Wilbert and Krishna, Ranjay and Murali, Adithyavairavan and Mousavian, Arsalan and Fox, Dieter},
  booktitle={Conference on Robot Learning},
  pages={4005--4020},
  year={2025},
  organization={PMLR}
}

@inproceedings{zeng2021transporter,
  title={Transporter networks: Rearranging the visual world for robotic manipulation},
  author={Zeng, Andy and Florence, Pete and Tompson, Jonathan and Welker, Stefan and Chien, Jonathan and Attarian, Maria and Armstrong, Travis and Krasin, Ivan and Duong, Dan and Sindhwani, Vikas and others},
  booktitle={Conference on Robot Learning},
  pages={726--747},
  year={2021},
  organization={PMLR}
}

@inproceedings{shridhar2022cliport,
  title={Cliport: What and where pathways for robotic manipulation},
  author={Shridhar, Mohit and Manuelli, Lucas and Fox, Dieter},
  booktitle={Conference on robot learning},
  pages={894--906},
  year={2022},
  organization={PMLR}
}

@inproceedings{kuang2025ram,
  title={RAM: Retrieval-Based Affordance Transfer for Generalizable Zero-Shot Robotic Manipulation},
  author={Kuang, Yuxuan and Ye, Junjie and Geng, Haoran and Mao, Jiageng and Deng, Congyue and Guibas, Leonidas and Wang, He and Wang, Yue},
  booktitle={Conference on Robot Learning},
  pages={547--565},
  year={2025},
  organization={PMLR}
}

@inproceedings{liu2024moka,
  title={Moka: Open-world robotic manipulation through mark-based visual prompting},
  author={Liu, Fangchen and Fang, Kuan and Abbeel, Pieter and Levine, Sergey},
  booktitle={Robotics: Science and Systems},
  year={2024}
}

@inproceedings{borja2022affordance,
  title={Affordance learning from play for sample-efficient policy learning},
  author={Borja-Diaz, Jessica and Mees, Oier and Kalweit, Gabriel and Hermann, Lukas and Boedecker, Joschka and Burgard, Wolfram},
  booktitle={2022 International Conference on Robotics and Automation (ICRA)},
  pages={6372--6378},
  year={2022},
  organization={IEEE}
}

@inproceedings{tang2025uad,
  title={UAD: Unsupervised Affordance Distillation for Generalization in Robotic Manipulation},
  author={Tang, Yihe and Huang, Wenlong and Wang, Yingke and Li, Chengshu and Yuan, Roy and Zhang, Ruohan and Wu, Jiajun and Fei-Fei, Li},
  booktitle={2025 IEEE International Conference on Robotics and Automation (ICRA)},
  year={2025}
}

@inproceedings{lopes2007affordance,
  title={Affordance-based imitation learning in robots},
  author={Lopes, Manuel and Melo, Francisco S and Montesano, Luis},
  booktitle={2007 IEEE/RSJ international conference on intelligent robots and systems},
  pages={1015--1021},
  year={2007},
  organization={IEEE}
}

@inproceedings{huang2024copa,
  title={Copa: General robotic manipulation through spatial constraints of parts with foundation models},
  author={Huang, Haoxu and Lin, Fanqi and Hu, Yingdong and Wang, Shengjie and Gao, Yang},
  booktitle={2024 IEEE/RSJ International Conference on Intelligent Robots and Systems (IROS)},
  pages={9488--9495},
  year={2024},
  organization={IEEE}
}


%% file: refs/5-high-level_planner.bib
@inproceedings{huang2019continuous,
  title={Continuous relaxation of symbolic planner for one-shot imitation learning},
  author={Huang, De-An and Xu, Danfei and Zhu, Yuke and Garg, Animesh and Savarese, Silvio and Fei-Fei, Li and Niebles, Juan Carlos},
  booktitle={2019 IEEE/RSJ International Conference on Intelligent Robots and Systems (IROS)},
  pages={2635--2642},
  year={2019},
  organization={IEEE}
}

@inproceedings{brohan2023can,
  title={Do as i can, not as i say: Grounding language in robotic affordances},
  author={Brohan, Anthony and Chebotar, Yevgen and Finn, Chelsea and Hausman, Karol and Herzog, Alexander and Ho, Daniel and Ibarz, Julian and Irpan, Alex and Jang, Eric and Julian, Ryan and others},
  booktitle={Conference on robot learning},
  pages={287--318},
  year={2023},
  organization={PMLR}
}

@article{huang2023grounded,
  title={Grounded decoding: Guiding text generation with grounded models for embodied agents},
  author={Huang, Wenlong and Xia, Fei and Shah, Dhruv and Driess, Danny and Zeng, Andy and Lu, Yao and Florence, Pete and Mordatch, Igor and Levine, Sergey and Hausman, Karol and others},
  journal={Advances in Neural Information Processing Systems},
  volume={36},
  pages={59636--59661},
  year={2023}
}

@article{huang2022inner,
  title={Inner monologue: Embodied reasoning through planning with language models},
  author={Huang, Wenlong and Xia, Fei and Xiao, Ted and Chan, Harris and Liang, Jacky and Florence, Pete and Zeng, Andy and Tompson, Jonathan and Mordatch, Igor and Chebotar, Yevgen and others},
  journal={arXiv preprint arXiv:2207.05608},
  year={2022}
}

@inproceedings{song2023llm,
  title={Llm-planner: Few-shot grounded planning for embodied agents with large language models},
  author={Song, Chan Hee and Wu, Jiaman and Washington, Clayton and Sadler, Brian M and Chao, Wei-Lun and Su, Yu},
  booktitle={Proceedings of the IEEE/CVF international conference on computer vision},
  pages={2998--3009},
  year={2023}
}

@article{liu2023llm+,
  title={Llm+ p: Empowering large language models with optimal planning proficiency},
  author={Liu, Bo and Jiang, Yuqian and Zhang, Xiaohan and Liu, Qiang and Zhang, Shiqi and Biswas, Joydeep and Stone, Peter},
  journal={arXiv preprint arXiv:2304.11477},
  year={2023}
}

@inproceedings{liu2023reflect,
  title={REFLECT: Summarizing Robot Experiences for Failure Explanation and Correction},
  author={Liu, Zeyi and Bahety, Arpit and Song, Shuran},
  booktitle={Conference on Robot Learning},
  pages={3468--3484},
  year={2023},
  organization={PMLR}
}

@article{singh2024malmm,
  title={Malmm: Multi-agent large language models for zero-shot robotics manipulation},
  author={Singh, Harsh and Das, Rocktim Jyoti and Han, Mingfei and Nakov, Preslav and Laptev, Ivan},
  journal={arXiv preprint arXiv:2411.17636},
  year={2024}
}

@inproceedings{wang2024polaris,
  title={Polaris: Open-ended interactive robotic manipulation via syn2real visual grounding and large language models},
  author={Wang, Tianyu and Lin, Haitao and Yu, Junqiu and Fu, Yanwei},
  booktitle={2024 IEEE/RSJ International Conference on Intelligent Robots and Systems (IROS)},
  pages={9676--9683},
  year={2024},
  publisher={IEEE}
}

@inproceedings{zhao2023chat,
  title={Chat with the environment: Interactive multimodal perception using large language models},
  author={Zhao, Xufeng and Li, Mengdi and Weber, Cornelius and Hafez, Muhammad Burhan and Wermter, Stefan},
  booktitle={2023 IEEE/RSJ International Conference on Intelligent Robots and Systems (IROS)},
  pages={3590--3596},
  year={2023},
  organization={IEEE}
}

@inproceedings{mandi2024roco,
  title={Roco: Dialectic multi-robot collaboration with large language models},
  author={Mandi, Zhao and Jain, Shreeya and Song, Shuran},
  booktitle={2024 IEEE International Conference on Robotics and Automation (ICRA)},
  pages={286--299},
  year={2024},
  organization={IEEE}
}

@inproceedings{li2023blip,
  title={Blip-2: Bootstrapping language-image pre-training with frozen image encoders and large language models},
  author={Li, Junnan and Li, Dongxu and Savarese, Silvio and Hoi, Steven},
  booktitle={International conference on machine learning},
  pages={19730--19742},
  year={2023},
  organization={PMLR}
}

@inproceedings{du2024video,
  title={Video Language Planning},
  author={Du, Yilun and Yang, Sherry and Florence, Pete and Xia, Fei and Wahid, Ayzaan and Sermanet, Pierre and Yu, Tianhe and Abbeel, Pieter and Tenenbaum, Joshua B and Kaelbling, Leslie Pack and others},
  booktitle={The Twelfth International Conference on Learning Representations},
  year={2024}
}

@inproceedings{gao2024physically,
  title={Physically grounded vision-language models for robotic manipulation},
  author={Gao, Jensen and Sarkar, Bidipta and Xia, Fei and Xiao, Ted and Wu, Jiajun and Ichter, Brian and Majumdar, Anirudha and Sadigh, Dorsa},
  booktitle={2024 IEEE International Conference on Robotics and Automation (ICRA)},
  pages={12462--12469},
  year={2024},
  organization={IEEE}
}

@article{mu2023embodiedgpt,
  title={Embodiedgpt: Vision-language pre-training via embodied chain of thought},
  author={Mu, Yao and Zhang, Qinglong and Hu, Mengkang and Wang, Wenhai and Ding, Mingyu and Jin, Jun and Wang, Bin and Dai, Jifeng and Qiao, Yu and Luo, Ping},
  journal={Advances in Neural Information Processing Systems},
  volume={36},
  pages={25081--25094},
  year={2023}
}

@inproceedings{zhang2025learning2,
  title={Learning manipulation skills through robot chain-of-thought with sparse failure guidance},
  author={Zhang, Kaifeng and Yin, Zhao-Heng and Ye, Weirui and Gao, Yang},
  booktitle={2025 IEEE/RSJ International Conference on Intelligent Robots and Systems (IROS)},
  pages={8012--8018},
  year={2025},
  organization={IEEE}
}

@article{qi2025sofar,
  title={Sofar: Language-grounded orientation bridges spatial reasoning and object manipulation},
  author={Qi, Zekun and Zhang, Wenyao and Ding, Yufei and Dong, Runpei and Yu, Xinqiang and Li, Jingwen and Xu, Lingyun and Li, Baoyu and He, Xialin and Fan, Guofan and others},
  journal={Advances in neural information processing systems},
  volume={38},
  pages={76367--76412},
  year={2025}
}

@inproceedings{zeng2023socratic,
  title={Socratic Models: Composing Zero-Shot Multimodal Reasoning with Language},
  author={Zeng, Andy and Attarian, Maria and Choromanski, Krzysztof Marcin and Wong, Adrian and Welker, Stefan and Tombari, Federico and Purohit, Aveek and Ryoo, Michael S and Sindhwani, Vikas and Lee, Johnny and others},
  booktitle={The Eleventh International Conference on Learning Representations},
  year={2023}
}

@inproceedings{duan2025aha,
  title={Aha: A vision-language-model for detecting and reasoning over failures in robotic manipulation},
  author={Duan, Jiafei and Pumacay, Wilbert and Kumar, Nishanth and Wang, Yi Ru and Tian, Shulin and Yuan, Wentao and Krishna, Ranjay and Fox, Dieter and Mandlekar, Ajay and Guo, Yijie},
  booktitle={The Thirteenth International Conference on Learning Representations},
  year={2025}
}

@inproceedings{ji2025robobrain,
  title={Robobrain: A unified brain model for robotic manipulation from abstract to concrete},
  author={Ji, Yuheng and Tan, Huajie and Shi, Jiayu and Hao, Xiaoshuai and Zhang, Yuan and Zhang, Hengyuan and Wang, Pengwei and Zhao, Mengdi and Mu, Yao and An, Pengju and others},
  booktitle={Proceedings of the Computer Vision and Pattern Recognition Conference},
  pages={1724--1734},
  year={2025}
}

@article{team2025gemini,
  title={Gemini robotics: Bringing ai into the physical world},
  author={Team, Gemini Robotics and Abeyruwan, Saminda and Ainslie, Joshua and Alayrac, Jean-Baptiste and Arenas, Montserrat Gonzalez and Armstrong, Travis and Balakrishna, Ashwin and Baruch, Robert and Bauza, Maria and Blokzijl, Michiel and others},
  journal={arXiv preprint arXiv:2503.20020},
  year={2025}
}

@article{dang2025rynnec,
  title={RynnEC: Bringing MLLMs into Embodied World},
  author={Dang, Ronghao and Yuan, Yuqian and Mao, Yunxuan and Li, Kehan and Liu, Jiangpin and Wang, Zhikai and Li, Xin and Wang, Fan and Zhao, Deli},
  journal={arXiv preprint arXiv:2508.14160},
  year={2025}
}

@inproceedings{zhou2025code,
  title={Code-as-monitor: Constraint-aware visual programming for reactive and proactive robotic failure detection},
  author={Zhou, Enshen and Su, Qi and Chi, Cheng and Zhang, Zhizheng and Wang, Zhongyuan and Huang, Tiejun and Sheng, Lu and Wang, He},
  booktitle={2025 IEEE/CVF Conference on Computer Vision and Pattern Recognition (CVPR)},
  pages={6919--6929},
  year={2025},
  organization={IEEE}
}

@inproceedings{venkatesh2021translating,
  title={Translating natural language instructions to computer programs for robot manipulation},
  author={Venkatesh, Sagar Gubbi and Upadrashta, Raviteja and Amrutur, Bharadwaj},
  booktitle={2021 IEEE/RSJ International Conference on Intelligent Robots and Systems (IROS)},
  pages={1919--1926},
  year={2021},
  organization={IEEE}
}

@inproceedings{liang2023code,
  title={Code as Policies: Language Model Programs for Embodied Control},
  author={Liang, Jacky and Huang, Wenlong and Xia, Fei and Xu, Peng and Hausman, Karol and Ichter, Brian and Florence, Pete and Zeng, Andy},
  booktitle={2023 IEEE International Conference on Robotics and Automation (ICRA)},
  pages={9493--9500},
  year={2023},
  organization={IEEE}
}

@inproceedings{singh2023progprompt,
  title={ProgPrompt: Generating Situated Robot Task Plans using Large Language Models},
  author={Singh, Ishika and Blukis, Valts and Mousavian, Arsalan and Goyal, Ankit and Xu, Danfei and Tremblay, Jonathan and Fox, Dieter and Thomason, Jesse and Garg, Animesh},
  booktitle={2023 IEEE International Conference on Robotics and Automation (ICRA)},
  pages={11523--11530},
  year={2023},
  organization={IEEE}
}

@article{huang2023instruct2act,
  title={Instruct2act: Mapping multi-modality instructions to robotic actions with large language model},
  author={Huang, Siyuan and Jiang, Zhengkai and Dong, Hao and Qiao, Yu and Gao, Peng and Li, Hongsheng},
  journal={arXiv preprint arXiv:2305.11176},
  year={2023}
}

@inproceedings{wang2023demo2code,
  title={Demo2code: From summarizing demonstrations to synthesizing code via extended chain-of-thought},
  author={Wang, Yuki and Gonzalez-Pumariega, Gonzalo and Sharma, Yash and Choudhury, Sanjiban},
  booktitle={Advances in Neural Information Processing Systems},
  volume={36},
  pages={14848--14956},
  year={2023}
}

@inproceedings{murray2024teaching,
  title={Teaching robots with show and tell: Using foundation models to synthesize robot policies from language and visual demonstration},
  author={Murray, Michael and Gupta, Abhishek and Cakmak, Maya},
  booktitle={8th Annual Conference on Robot Learning},
  year={2024}
}

@inproceedings{yoneda2024statler,
  title={Statler: State-maintaining language models for embodied reasoning},
  author={Yoneda, Takuma and Fang, Jiading and Li, Peng and Zhang, Huanyu and Jiang, Tianchong and Lin, Shengjie and Picker, Ben and Yunis, David and Mei, Hongyuan and Walter, Matthew R},
  booktitle={2024 IEEE International Conference on Robotics and Automation (ICRA)},
  pages={15083--15091},
  year={2024},
  organization={IEEE}
}

@article{liu2025hycodepolicy,
  title={HyCodePolicy: Hybrid Language Controllers for Multimodal Monitoring and Decision in Embodied Agents},
  author={Liu, Yibin and Liang, Zhixuan and Chen, Zanxin and Chen, Tianxing and Hu, Mengkang and Dong, Wanxi and Xu, Congsheng and Han, Zhaoming and Qin, Yusen and Mu, Yao},
  journal={arXiv preprint arXiv:2508.02629},
  year={2025}
}

@inproceedings{huang2023voxposer,
  title={VoxPoser: Composable 3D Value Maps for Robotic Manipulation with Language Models},
  author={Huang, Wenlong and Wang, Chen and Zhang, Ruohan and Li, Yunzhu and Wu, Jiajun and Fei-Fei, Li},
  booktitle={Conference on Robot Learning},
  pages={540--562},
  year={2023},
  organization={PMLR}
}

@inproceedings{li2024manipllm,
  title={Manipllm: Embodied multimodal large language model for object-centric robotic manipulation},
  author={Li, Xiaoqi and Zhang, Mingxu and Geng, Yiran and Geng, Haoran and Long, Yuxing and Shen, Yan and Zhang, Renrui and Liu, Jiaming and Dong, Hao},
  booktitle={Proceedings of the IEEE/CVF Conference on Computer Vision and Pattern Recognition},
  pages={18061--18070},
  year={2024}
}

@inproceedings{huang2025rekep,
  title={ReKep: Spatio-Temporal Reasoning of Relational Keypoint Constraints for Robotic Manipulation},
  author={Huang, Wenlong and Wang, Chen and Li, Yunzhu and Zhang, Ruohan and Fei-Fei, Li},
  booktitle={Conference on Robot Learning},
  pages={4573--4602},
  year={2025},
  organization={PMLR}
}

@article{tang2025geomanip,
  title={Geomanip: Geometric constraints as general interfaces for robot manipulation},
  author={Tang, Weiliang and Pan, Jia-Hui and Liu, Yun-Hui and Tomizuka, Masayoshi and Li, Li Erran and Fu, Chi-Wing and Ding, Mingyu},
  journal={arXiv preprint arXiv:2501.09783},
  year={2025}
}

@inproceedings{hu2024look,
  title={Look Before You Leap: Unveiling the Power of GPT-4V in Robotic Vision-Language Planning},
  author={Hu, Yingdong and Lin, Fanqi and Zhang, Tong and Yi, Li and Gao, Yang},
  booktitle={First Workshop on Vision-Language Models for Navigation and Manipulation at ICRA 2024},
  year={2024}
}


%% file: refs/6-il.bib
@inproceedings{huang2019non,
  title={Non-parametric imitation learning of robot motor skills},
  author={Huang, Yanlong and Rozo, Leonel and Silv{\'e}rio, Joao and Caldwell, Darwin G},
  booktitle={2019 International Conference on Robotics and Automation (ICRA)},
  pages={5266--5272},
  year={2019},
  organization={IEEE}
}

@article{sun2025exploring,
  title={Exploring Pose-Guided Imitation Learning for Robotic Precise Insertion},
  author={Sun, Han and Wang, Yizhao and Zhou, Zhenning and Wang, Shuai and Yang, Haibo and Sun, Jingyuan and Cao, Qixin},
  journal={arXiv preprint arXiv:2505.09424},
  year={2025}
}

@inproceedings{dalal2023imitating,
  title={Imitating Task and Motion Planning with Visuomotor Transformers},
  author={Dalal, Murtaza and Mandlekar, Ajay and Garrett, Caelan Reed and Handa, Ankur and Salakhutdinov, Ruslan and Fox, Dieter},
  booktitle={Conference on Robot Learning},
  pages={2565--2593},
  year={2023},
  organization={PMLR}
}

@inproceedings{mcdonald2022guided,
  title={Guided imitation of task and motion planning},
  author={McDonald, Michael James and Hadfield-Menell, Dylan},
  booktitle={Conference on Robot Learning},
  pages={630--640},
  year={2022},
  organization={PMLR}
}

@inproceedings{katz2016imitation,
  title={Imitation learning as cause-effect reasoning},
  author={Katz, Garrett and Huang, Di-Wei and Gentili, Rodolphe and Reggia, James},
  booktitle={International Conference on Artificial General Intelligence},
  pages={64--73},
  year={2016},
  organization={Springer}
}
